\documentclass[11pt]{article}

\usepackage[final]{acl}

\usepackage{times}
\usepackage{latexsym}

\usepackage[T1]{fontenc}
\usepackage[utf8]{inputenc}

\usepackage{microtype}

\usepackage{inconsolata}

\usepackage{graphicx}

\usepackage{xspace}

\newcommand{\eg}{\emph{e.g.,}\xspace}
\newcommand{\ie}{\emph{i.e.,}\xspace}

\newcommand{\cf}{\emph{cf.}\xspace}
\usepackage{amsmath}
\usepackage{hyperref}
\usepackage{url}
\usepackage{booktabs}
\usepackage{graphicx} 
\usepackage[table,x11names]{xcolor}
\usepackage{multirow}
\usepackage{pifont}
\usepackage{wrapfig} 
\usepackage{amssymb}
\usepackage{amsthm}
\usepackage{caption}
\usepackage{subcaption}
\usepackage{colortbl}
\usepackage{float}
\usepackage{siunitx}
\usepackage{fancyhdr}
\usepackage{enumitem}

\definecolor{mycolor}{rgb}{0, 0, 0} 
\newcommand{\coloredtext}[1]{\textcolor{mycolor}{#1}}

\newtheorem{theorem}{Theorem}       % 定义“theorem”环境（全局编号）
\newtheorem{assumption}{Assumption}

\usepackage{titlesec}

\titlespacing*{\paragraph}
  {0pt}            % 左缩进
  {1ex plus .2ex minus .2ex}  % 段前竖直间距（缩小）
  {1em}            % 标题后到正文的水平间距

\makeatletter
\newcommand{\appendixtableofcontents}{%
  \section*{Contents}% 标题你可以改
    \begingroup
    \small
    \setlength{\parskip}{2ex plus 0.2ex minus 0.2ex}%
    \setlength{\itemsep}{0.4ex}%
    \@starttoc{app}%
  \endgroup
}

\newcommand{\appsection}[1]{%
  \section{#1}%
  \addcontentsline{app}{section}{\protect\numberline{\thesection}#1}%
}

\newcommand{\appsubsection}[1]{%
  \subsection{#1}%
  \addcontentsline{app}{subsection}{\protect\numberline{\thesubsection}#1}%
}
\makeatother

\title{Rethinking Entropy Interventions in RLVR: \\
An Entropy Change Perspective}

\author{\textbf{Zhezheng Hao}$^{1}$\thanks{Equal Contribution.}\footnotemark[3]\quad
  \textbf{Hong Wang}$^{2}$\footnotemark[1]\quad
  \textbf{Haoyang Liu}$^{3}$\quad
  \textbf{Jian Luo}$^{3}$\\[3pt]
  \textbf{Jiarui Yu}$^{2}$\quad
  \textbf{Hande Dong}$^{2}$\thanks{Corresponding Authors}\quad
  \textbf{Qiang Lin}$^{2}$\quad
  \textbf{Can Wang}$^{1,4}$\thanks{State Key Laboratory of Blockchain and Data Security, Zhejiang University}\quad
  \textbf{Jiawei Chen}$^{1,4}$\footnotemark[2]\footnotemark[3]\\[3pt]
  $^{1}$ Zhejiang University \quad
  $^{2}$ Tencent \quad
  $^{3}$ Independent Researcher \\[3pt]
  $^{4}$ Hangzhou High-Tech Zone (Binjiang) Institute of Blockchain and Data Security \\[3pt]
    \small Emails: \url{haozhezheng@outlook.com}, 
  \small\url{donghd66@gmail.com}, 
  \small\url{sleepyhunt@zju.edu.cn}
}

\begin{document}

\maketitle

\begin{abstract}
Reinforcement Learning with Verifiable Rewards (RLVR) serves as a cornerstone technique for enhancing the reasoning capabilities of Large Language Models (LLMs). However, its training is often plagued by \emph{entropy collapse}, a rapid decline in policy entropy that limits exploration and undermines training effectiveness. While recent works attempt to mitigate this issue via several heuristic entropy interventions, the underlying mechanisms remain poorly understood. In this work, we conduct comprehensive theoretical and empirical analyses of entropy dynamics in RLVR, offering two main insights: (1) We derive a tight analytical approximation for token-level entropy change at each update step, revealing four governing factors and providing a unified theoretical framework to explain how existing methods influence entropy; (2) We reveal a fundamental limitation of recent approaches: they rely on heuristic adjustments to one or two of these factors, leaving other relevant factors unconsidered, thus inherently limiting their effectiveness. Motivated by these findings, we propose STEER, a principled entropy-modulation method that adaptively reweights tokens based on theoretically-estimated entropy variations. Extensive experiments across six mathematical reasoning and three coding benchmarks demonstrate that STEER effectively mitigates entropy collapse and consistently outperforms state-of-the-art baselines.\footnote{Code is available at \url{https://github.com/zz-haooo/STEER}.}

\end{abstract}

\section{Introduction}

% Large Language Models (LLMs) have recently demonstrated remarkable reasoning capabilities in complex tasks such as mathematics and coding~\citep{gpt52, gemini3, anthropic2025claude45}.
% A key technique driving these advances is Reinforcement Learning with Verifiable Rewards (RLVR), which adopts policy-gradient algorithms such as PPO~\citep{schulman2017proximal}, GRPO~\citep{shao2024deepseekmath}, and others~\citep{hu2025reinforce++,hao2025policy, ahmadian2024back}.
% This verifiable reward mechanism provides effective learning signals that enable RL training to scale~\citep{jaech2024openai, shao2024deepseekmath, liu2025deepseek}.
% Large Language Models (LLMs) have recently demonstrated remarkable reasoning capabilities in complex tasks such as mathematics and coding~\citep{gpt52, gemini3, anthropic2025claude45}. A key technique driving these advances is {Reinforcement Learning with Verifiable Rewards} (RLVR)~\citep{shao2024deepseekmath, team2025kimi}.  RLVR employs policy-gradient algorithms (\eg  PPO~\citep{schulman2017proximal} and GRPO~\citep{shao2024deepseekmath}) with verifiable reward signals, providing an effective supervision mechanism for post-training.
% This paradigm has been instrumental in unlocking the post-training scaling of reasoning performance~\citep{jaech2024openai, shao2024deepseekmath, liu2025deepseek},  thereby substantially advancing the reasoning capabilities of LLMs and laying the foundation for recent progress in advanced reasoning models.
Large Language Models (LLMs) have recently demonstrated remarkable reasoning capabilities in complex tasks such as mathematics and coding~\citep{gpt52, gemini3, anthropic2025claude45, zhang2025survey}. A key technique driving these advances is {Reinforcement Learning with Verifiable Rewards} (RLVR)~\citep{shao2024deepseekmath,team2025kimi}.  RLVR employs policy-gradient algorithms (\eg  PPO~\citep{schulman2017proximal} and GRPO~\citep{shao2024deepseekmath}) guided by automatically verifiable reward signals.
%providing an effective supervision mechanism for post-training.
This paradigm has been instrumental in unlocking the post-training scaling of reasoning performance~\citep{jaech2024openai, shao2024deepseekmath, liu2025deepseek},  serving as a fundamental technique for the latest generation of advanced reasoning models.

Despite its success, recent studies have identified a major pitfall of RLVR: \emph{entropy collapse},  a rapid decline in policy entropy during training~\citep{wu2025invisible, song2025outcome, li2025cure, cui2025entropy}. This collapse degrades training in two critical ways: (1) It severely restricts exploration, causing the model to generate homogeneous reasoning trajectories (rollouts), thereby failing to discover informative and potentially correct solutions. (2) It disrupts optimization.  Particularly for advanced group-based algorithms like GRPO~\citep{shao2024deepseekmath}, the computed advantages become less discriminative, causing training to stagnate. This raises an important research question: \emph{How can policy entropy be effectively modulated to preserve exploration?}

% Several recent methods for addressing entropy collapse can be broadly classified into three categories: (1) \emph{Clip‑Higher}, which increases the upper bound of the importance‑sampling ratio~\citep{yu2025dapo, yang2025dcpo}; (2) \emph{Positive‑Reweighting}, which reduces the weight of positive samples with high generation probability~\citep{zhu2025surprising, he2025rewarding}; (3) \emph{Entropy‑Aware Advantage}, which assigns larger advantage to tokens with high entropy~\citep{cheng2025reasoning, tan2025gtpo, wang2025beyond, wang2025stabilizing, deng2025decomposing}. While these strategies can alleviate entropy collapse to some extent, they remain largely heuristic, and the mechanisms through which they affect entropy are still poorly understood. Consequently, their effectiveness is limited, and entropy remains only loosely controlled. These limitations call for a principled understanding of entropy dynamics in RLVR---one that not only explains the behavior of existing methods, but also guides the design of more effective entropy-control strategies.

Several recent methods for tackling entropy collapse can be broadly classified into three categories: (1) \emph{Clip‑Higher}, which increases the upper bound of the importance‑sampling ratio~\citep{yu2025dapo, yang2025dcpo}; (2) \emph{Positive‑Reweighting}, which down-weights positive samples with high generation probability~\citep{zhu2025surprising, he2025rewarding}; (3) \emph{Entropy‑Aware Advantage}, which assigns larger advantages to tokens with high entropy~\citep{cheng2025reasoning, tan2025gtpo, wang2025beyond, wang2025stabilizing, deng2025decomposing}. While these strategies empirically mitigate the collapse to some extent, they remain fundamentally heuristic, leaving the underlying mechanisms of their impact on entropy largely opaque. Consequently, their effectiveness is bounded, and entropy remains only loosely controlled. These limitations call for {a principled understanding of entropy dynamics in RLVR} --- one that not only demystifies existing heuristics, but also guides the design of more effective entropy-modulation strategies.

Towards this end, this work conducts comprehensive theoretical and empirical analyses of entropy dynamics in RLVR. Moving beyond conventional coarse-grained global expectations, we scrutinize fine-grained token-level entropy variations at each training step. While ~\citet{cui2025entropy} offers a preliminary exploration of entropy, their analyses rely on unrealistic assumptions that lead to highly imprecise estimates (\cf Section 3, Remark 1) and cannot explain the mechanisms underlying existing entropy-intervention methods. In contrast, our analysis yields two main insights: (1) We derive a precise analytical approximation for token-level entropy change, revealing that it is governed by four key factors --- clipping strategy, advantage, token probability, and conditional entropy. We further clarify how existing methods influence these factors, thereby explaining their empirical success. (2) We identify key limitations of prior methods: their heuristic modulations are applied only to a small subset of tokens, neglecting others that may also experience severe entropy collapse; and they consider only partial relevant factors, which reduces their effectiveness and can even accelerate entropy collapse in some cases.

Motivated by these insights, we introduce STEER (Stabilizing Token-level Entropy-changE via Reweighting), a simple yet principled entropy-modulation method grounded in our theoretical analysis. STEER directly translates the theoretically-estimated entropy variations into adaptive token-level weights. By down-weighting tokens with excessively large entropy change, it enables fine-grained modulation of step-wise entropy dynamics, and effectively steers the policy toward sustained exploration. Extensive experiments across six mathematical reasoning and three coding benchmarks demonstrate that STEER consistently outperforms state-of-the-art baselines by a substantial margin. Importantly, STEER also generalizes well across model scales (1.5B/7B/14B), model families (Qwen/Llama/Mistral), and RL algorithms (GRPO/RLOO/OPO), consistently maintaining stable entropy dynamics and achieving stronger performance.

% In summary, our contributions are:
% \begin{itemize}[itemsep=3pt, topsep=2pt, parsep=0pt, leftmargin=1em, labelsep=0.4em]
% \item We conduct comprehensive theoretical analyses of token-level entropy dynamics,  revealing its key governing factors and explaining the mechanisms and limitations of recent entropy-intervention strategies.

% \item We propose STEER, a theoretically-grounded entropy-control method that adaptively down-weights tokens with large entropy changes.

% \item We conduct comprehensive experiments on six mathematical reasoning benchmarks and three coding benchmarks, demonstrating the effectiveness of STEER over strong baselines and its generalization across model scales, model families, and RL algorithms.
% \end{itemize}

In summary, our main contributions are:
\begin{itemize}[itemsep=3pt, topsep=2pt, parsep=0pt, leftmargin=1em, labelsep=0.4em]
\item We derive a tight analytical approximation for token-level entropy dynamics, revealing four governing factors and thereby demystifying the mechanisms of existing heuristic entropy-intervention strategies.

\item We propose STEER, a theoretically grounded entropy-modulation method that adaptively down-weights tokens susceptible to drastic entropy decay. 

\item We comprehensively evaluate STEER across six mathematical reasoning and three coding benchmarks, demonstrating its superiority over strong baselines and generalization across varying models and RL algorithms.
\end{itemize}

\section{Preliminaries}

\subsection{RLVR Algorithms}
% Given a prompt $q$ sampled from data $\mathcal{D}$, let $\pi_{\theta}$ denote the policy model parameterized with $\theta$ and $o$ denote the response sampled from $\pi_{\text{old}}(\cdot | q)$.
% Given a query $q$ sampled from a dataset $\mathcal{D}$, let $\pi_\theta$ be the policy model and $o$ be a sampled response.
% PPO~\citep{schulman2017proximal} optimizes the policy by maximizing the expected advantage and stabilizes the training process through the clipped surrogate.
% GRPO~\citep{shao2024deepseekmath} removes the value model and instead samples a group of rollouts $\{o_i\}_{i=1}^{G}$ for query $q$, optimizing LLM policies using relative advantage scores computed within groups of samples:

% In Reinforcement Learning with Verifiable Rewards (RLVR), the goal is to optimize an LLM policy to generate correct responses for given queries. 
Traditional algorithms like PPO~\citep{schulman2017proximal} drive policy optimization by maximizing an expected advantage, heavily relying on a separate value network for baseline estimation. In contrast, GRPO~\citep{shao2024deepseekmath} mitigates memory overhead by eliminating the value model, instead computing relative advantages within a sampled group.

% Formally, let $q$ be a query sampled from a dataset $\mathcal{D}$, and $\pi_\theta$ be the policy model. For each query $q$, a group of responses (rollouts) $\{o_i\}_{i=1}^{G}$ is sampled from a reference policy $\pi_{old}$. Each response $o_i$ is formulated as a token sequence, where $o_{i,t}$ denotes its $t$-th token. GRPO computes the relative advantage as:
Formally, let $q$ be a query sampled from a dataset $\mathcal{D}$, and $\pi_\theta$ be the policy model. GRPO samples a group of responses (rollouts) $\{o_i\}_{i=1}^{G}$ from a reference policy $\pi_{old}$. For the $t$-th token $o_{i,t}$ within response $o_i$, the advantage is computed as:
\begin{equation}
A_{i,t}=\frac{R_i-\text{mean}(\{R_i\}_{i=1}^G)}{\text{std}(\{R_i\}_{i=1}^G)},
\end{equation}
where $R_i \in \{1,-1\}$ indicates the correctness of $o_i$. This advantage $A_{i,t}$ typically remains constant across all tokens within the response. 
% where $R_i \in \{1,-1\}$ indicates whether the $i$-th response is correct, and $A_{i,t}$ denotes the advantage assigned to $t$-th token of the $i$-th response --- a value that is typically shared across all tokens within the response.    
% where $R_i \in \{1,-1\}$ indicates whether the $i$-th response is correct,
% and $A_{i,t}$ denotes the advantage assigned to $t$-th token in the $i$-th response --- a value that is typically shared across all tokens within the response. 
% and the advantage $A_{i,t}$ is shared across all tokens in the response.
Following the token-level formulation~\citep{yu2025dapo}, GRPO maximizes the following objective:
\begin{equation}\label{grpo_Eq}
\begin{aligned}
\mathcal{J}(\theta)
= &\mathbb{E}_{\substack{
    q \sim \mathcal{D}, \\
    \{o_i\}_{i=1}^G \sim \pi_{\text{old}}(\cdot \mid q)
}} \Bigg[
    \frac{1}{L} 
    \sum_{i=1}^{G} \sum_{t=1}^{|o_i|} \min \Big(
        r_{i,t} A_{i,t},\,
\\
&  \operatorname{clip}\big( r_{i,t}, 1-\varepsilon, 1+\varepsilon \big) A_{i,t}
    \Big)
\Bigg].
\end{aligned}
\end{equation}
where $r_{i,t} = \frac{\pi_\theta(o_{i,t}|q,o_{i,<t})}{\pi_{\text{old}}(o_{i,t}|q,o_{i,<t})}$ denotes the importance sampling ratio and $L=\sum_{i=1}^{G} |o_i|$ denotes the sum of response lengths within a group. Here
we omit the KL divergence term following~\citep{shao2024deepseekmath}, which has been shown to yield better performance and broader exploration~\citep{chu2025gpg, hu2025open}.

\subsection{Policy Entropy of LLMs}
Shannon entropy quantifies the uncertainty of a policy model's action selection given a state~\citep{haarnoja2018soft}.
For LLMs, entropy can be quantified at each token generation step.
% For each sampled response $o_i$ and step $t$, the next-token entropy under policy $\pi_\theta$ is:
For each sampled response $o_i$ and its $t$-th token generation, the token-level entropy under policy $\pi_\theta$ is:
\begin{equation}\label{token_entropy}
\mathcal{H}(q, o_{i,<t})
=
- \mathbb{E}_{\substack{
    a \sim \pi_\theta(\cdot \mid q, o_{i,<t})
}}
\left[
    \log \pi_\theta(a \mid q, o_{i,<t})
\right].
\end{equation}
% Policy entropy measures a policy model’s overall uncertainty on a dataset by averaging token entropy over all tokens.
% For policy model $\pi_{\theta}$ and the dataset $\mathcal{D}$, the policy entropy is defined as
The (global) policy entropy measures the model’s generation uncertainty over a query dataset, which can be estimated by averaging token-level entropy over sampled responses:
\begin{equation}
\mathcal{H}(\pi_\theta)
=
\mathbb{E}_{\substack{
    q \sim \mathcal{D}, \\
    \{o_i\} \sim \pi_{\theta}(\cdot \mid q)
}}
    \frac{1}{L}
    \sum_{i=1}^G \sum_{t=1}^{|o_i|} \mathcal{H}(q, o_{i,<t}).
\end{equation}
% Although $\mathcal{H}(\pi_\theta)$ is computed on a given  query distribution $\mathcal{D}$, it can reflect the diversity of model's responses 
In practice, we may estimate $\mathcal{H}(\pi_\theta)$ using the training dataset and empirically observe that this estimation generalizes well to unseen data (\eg test sets, \cf Appendix~\ref{Empirical Properties of Policy Entropy}).

Maintaining adequate policy entropy is essential for balancing exploration and exploitation during RL training~\citep{haarnoja2018soft, ziebart2008maximum}.  However, RLVR training often suffers from \emph{entropy collapse}, in which policy entropy drops rapidly~\citep{song2025outcome, li2025cure, cui2025entropy}. This notorious phenomenon is detrimental in two respects: (1) It severely weakens exploration, limiting the model’s ability to discover informative and potentially correct solutions. (2)  It also undermines training effectiveness. Typically, for group-based methods such as GRPO~\citep{shao2024deepseekmath}, increasingly homogeneous rollouts yield relative advantage scores $A_{i,t}$ that are less discriminative, causing training to stagnate.

\section{Analyses on Entropy Dynamics}
% 对于一个模型来说，熵的大小侧面反应了模型当前的多样性。
% 训练的过程，整体熵的变化，反应了当前模型探索与利用的平衡。整体宏观的熵变由逐token微观的熵变组合而成。单步训练对单个token造成熵变是熵动态过程的最小组成单元。
% 本文将从该微观角度出发，通过理论推导，定量的分析影响单个token熵变的直接因素。然后再从该微观视角出发，分析现有算法和参数对宏观熵的影响。
% 下面我们先分析影响单个token熵变的直接因素
% We argue that the overall policy entropy during training is determined by the accumulation of per-token entropy changes.
% Therefore, analyzing entropy change at token level helps reveal the overall entropy dynamics.
% From this token-level perspective, we derive a quantitative analysis of entropy change and use it to examine how existing entropy methods
% shape the policy entropy during training.
In this section, we present a comprehensive theoretical and empirical analysis of entropy change. Our investigation is conducted at a fine-grained token level, which enables precise monitoring of entropy dynamics.
Without loss of generality, this section primarily focuses on the representative GRPO algorithm for clarity of exposition. Nevertheless, \textbf{these theoretical results generalize seamlessly to other advanced RLVR algorithms} (\eg PPO~\citep{schulman2017proximal}, RLOO~\citep{ahmadian2024back} and OPO~\citep{hao2025policy}; \cf Appendix~\ref{Theoretical analyses extended to other RL algorithms}).

\coloredtext{To quantitatively characterize the per-token entropy change after one update, we rewrite the GRPO policy gradient (Eq.~\eqref{grpo_Eq}) as}
\begin{equation}
\begin{aligned}
    &\nabla_\theta J(\theta)
= \mathbb{E}_{\substack{
    q \sim \mathcal{D}, \\
    \{o_i\}_{i=1}^G \sim \pi_{\text{old}}(\cdot \mid q)
}} \Bigg[ \frac{1}{L} \sum_{i=1}^G\sum_{t=1}^{|o_i|} \\
&\mathbb{I}_{\text{clip}} \,r_{i,t} A_{i,t}
\nabla_\theta\log\pi_\theta(o_{i,t}\mid q,o_{i,<t}) \Bigg].
\end{aligned}
\end{equation}
% where the clipping indicator $\mathbb{I}_{\text{clip}}$ satisfies
\coloredtext{where the clipping indicator $\mathbb{I}_{\text{clip}}$ is derived from the ratio clipping operation and is defined as:}
\begin{equation}
\mathbb{I}_{\text{clip}} =
\begin{cases}
0, & A_{i,t} > 0 \;\;\text{and}\;\; r_{i,t} > 1+\varepsilon, \\
0, & A_{i,t} < 0 \;\;\text{and}\;\; r_{i,t} < 1-\varepsilon, \\
1, & \text{otherwise}.
\end{cases}
\end{equation}
\vspace{-0.2em}
% During the RLVR training, token logits are shaped by entangled parameters in the model, which makes entropy change difficult to quantify.
% To make the analysis tractable, we work under a logit-independent model, where a gradient step on the sampled token primarily changes its own logit and does not affect the logits of the other tokens.
% For token position $(i,t)$, we write the context (state)
% $s \triangleq (q,o_{i,<t})$ and token (action) $a \triangleq o_{i,t}$ for short and adopt this notation throughout the paper.
% When no ambiguity arises, we further abbreviate
% $\pi_\theta(a\mid s)$ as $\pi_\theta$ and $A(s,a)$ as $A$.
% Under this simplification, we derive the following theorem on token-level entropy change (proof shown in Appendix~\ref{proof}).
\noindent For notational convenience, we denote the state (\ie context) by $s \triangleq (q, o_{i,<t})$ and abbreviate $\pi_\theta(a \mid s)$ as $\pi_\theta$ and $A(s,a)$ as $A$.
Then we derive the following theorem on token-level entropy change.

\begin{theorem}\label{theorem_entropy_change} 
{\textit{(Token-level Entropy Change).}}
For a logit-independent policy model $\pi_\theta$, trained using GRPO with a learning rate $\eta$, the change of the token-level entropy on state $s$ between two consecutive steps can be approximated as:
\vspace{-0.2em}
\begin{equation}\label{Omega}
\begin{aligned}
\Omega(s)
&\triangleq
-\,\frac{\eta}{L}\,
\mathbb{E}_{\pi_\theta(\cdot \mid s)}
\Big[
\frac{\mathbb{I}_{\textup{clip}} \, A}{\pi_{\text{old}}}
\,\pi_\theta \bigl(1-\pi_\theta\bigr) \\
&\qquad\qquad\qquad
\bigl(
\log \pi_\theta
+
\mathcal{H}(s)
\bigr)
\Big],
\end{aligned}
\end{equation}
\vspace{-0.2em}
and the approximation error is bounded by $O(\eta^{2})$.
% \begin{equation}\label{Omega}
% \begin{aligned}
% \Delta \mathcal{H}(s)
% &\approx
% -\,\frac{\eta}{L}\,
% \mathbb{E}_{\pi_\theta(\cdot \mid s)}
% \Big[
% \frac{\mathbb{I}_{\textup{clip}} \, A}{\pi_{\text{old}}}
% \,\pi_\theta \bigl(1-\pi_\theta\bigr) \\
% &\qquad\qquad\qquad
% \bigl(
% \log \pi_\theta
% +
% \mathcal{H}(\pi_\theta\mid s)
% \bigr)
% \Big].
% \end{aligned}
% \end{equation}
% where the approximation error is bounded by $O(\eta^{2})$, with $\eta$ typically small ($< 10^{-4}$) in practice. 
\end{theorem}

The proof is presented in Appendix~\ref{proof}. The theorem is not specific to GRPO, and also holds for other algorithms (\cf Appendix~\ref{Theoretical analyses extended to other RL algorithms}). We highlight three important remarks:

\noindent\textbf{Remark 1 (Accurate Estimator): } \label{Remark 1} Notably, in practice, since the learning rate $\eta$ is typically small ($< 10^{-4}$), {Theorem~\ref{theorem_entropy_change} delivers an accurate closed-form estimator of token-level entropy dynamics.} In contrast, while prior work~\citep{cui2025entropy} also explores entropy change, it assumes a uniform entropy distribution across different queries within the same batch.
This assumption is rarely satisfied in practice, which leads to an inaccurate approximation of the ground-truth entropy change.

To quantify, we compare our entropy change estimation with their estimation (denoted as \textit{Cov}) during a standard GRPO training process.
Table~\ref{MSE_and_PCC} reports the Mean Squared Error (\textit{MSE}), Pearson Correlation Coefficient (\textit{PCC}), and Spearman's Rank Correlation Coefficient (\textit{SRCC}) between ground-truth entropy change and estimation.
Across all three metrics, our method delivers orders-of-magnitude lower MSE and substantially higher \textit{PCC} and \textit{SRCC} than \textit{Cov}. Remarkably, the MSE of our estimator is on the order of 1e-4, demonstrating its exceptional precision.
More results are shown in Appendix~\ref{Entropy Change Estimation Comparison}.

\begin{figure}[t]
% \vspace{-1em}
    \centering
\includegraphics[width=0.85\linewidth, height=4cm]{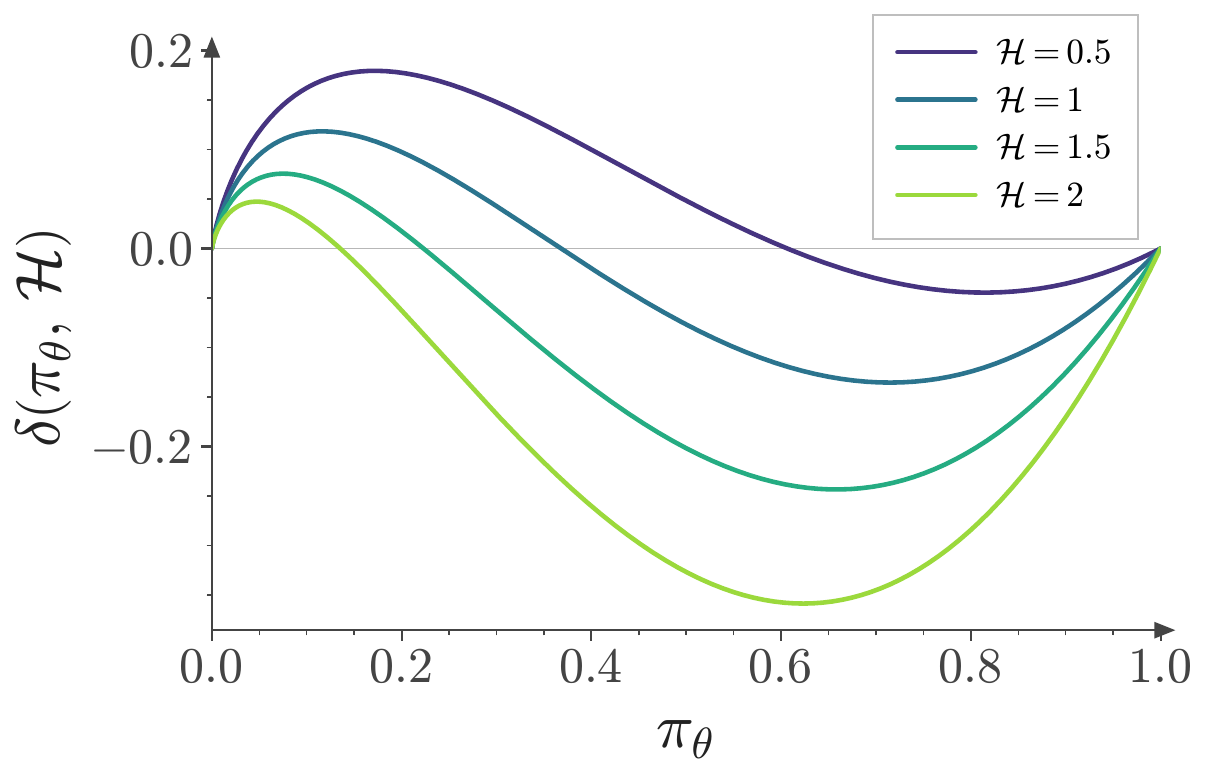}
\vspace{-0.8em}
    \caption{$\delta$ as a function of $\pi_\theta$ and $\mathcal{H}(s)$.}
    \label{delta_plot}
    \vspace{-0.5em}
\end{figure}

\noindent\textbf{Remark 2 (Four Governing Factors): }
Theorem~\ref{theorem_entropy_change} demonstrates that token-level entropy change is jointly determined by multiple factors:
\ding{182}  the clip indicator $\mathbb{I}_{\text{clip}}$, which prevents the entropy change of the tokens with overly large or small importance sampling ratio;
\ding{183} the advantage $A$ and the old-policy term $\pi_{old}$, which act as weighting factors for entropy change;
 \ding{184} the token generation probability $\pi_{\theta}$ and 
\ding{185} the token entropy $\mathcal{H}(s)$ of current state.
The contribution of $\pi_{\theta}$ and $\mathcal{H}(s)$ to entropy change can be expressed with the following function:
\begin{equation}\label{delta_def}
    \delta(\pi_{\theta}, \mathcal{H}) \triangleq - \pi_\theta (1 - \pi_\theta) \bigl[\log(\pi_\theta) + \mathcal{H}(s)\bigr].
\end{equation}
% Figure~\ref{delta_plot} illustrates the impact of the current policy $\pi_{\theta}$ and entropy $\mathcal{H}(s)$ on the function $\delta(\pi_{\theta}, \mathcal{H}(s))$.
The impact of $\pi_{\theta}$ and $\mathcal{H}(s)$ on the function $\delta(\pi_{\theta}, \mathcal{H}(s))$ is illustrated in Figure~\ref{delta_plot}.
% several factors.
% To analyze the property of token-level entropy change, we rewrite $\Omega(s)$ in Eq. (\ref{Omega}) as
% \begin{equation}
% \Omega(s) = 
% -\,\frac{\eta}{L}\,
% \mathbb{E}_{a \sim \pi_\theta^k(\cdot \mid s)}
% \big[
% \frac{\mathbb{I}_{\textup{clip}} \, A}{\pi_{\text{old}}} \,
% \delta(a|s)
% \big],
% \end{equation}
% where $\delta(a|s)$ is determined by token distribution
% \begin{equation}
%     \delta(a|s) \triangleq - \pi_\theta (1 - \pi_\theta) \bigl[\log(\pi_\theta) + \mathcal{H}(\pi_\theta|s)\bigr].
% \end{equation}
% Then, we visualize $\delta(a|s)$ as a function of $\pi_\theta$ and $\mathcal{H}(\pi_\theta|s)$ in Figure~\ref{delta_plot}.
% Based on this, we can now analyze entropy change by examining how token-level entropy changes with different advantage $A$, generation probability $\pi_\theta^k$ and current entropy $\mathcal{H}(\pi_\theta^k\mid s)$.

% \begin{figure}[htbp]
% % \vspace{-1em}
%     \centering
% \includegraphics[width=0.85\linewidth, height=3.8cm]{figs/function_plot_english.pdf}
% \vspace{-0.5em}
%     \caption{$\delta$ as a function of $\pi_\theta$ and $\mathcal{H}(s)$.}
%     \label{delta_plot}
%     \vspace{-1em}
% \end{figure}

\begin{table}[t]
    \centering
    \setlength{\tabcolsep}{5pt}
    \renewcommand{\arraystretch}{1.2}
    \small
    \definecolor{ourrow}{RGB}{235,243,252}
    \begin{tabular}{llrrr}
        \toprule
        \textbf{Model} & \textbf{Method}
            & \textbf{MSE}\,$\downarrow$
            & \textbf{PCC}\,$\uparrow$
            & \textbf{SRCC}\,$\uparrow$ \\
        \midrule
        \multirow{2}{*}{Math-1.5B}
            & Cov           & 5.37                        & $-6\mathrm{e}{-5}$ & $+0.04$ \\
            & \cellcolor{ourrow}\textbf{Ours} & \cellcolor{ourrow}$\mathbf{5e{-4}}$  & \cellcolor{ourrow}$\mathbf{+0.42}$   & \cellcolor{ourrow}$\mathbf{+0.65}$ \\
        \midrule
        \multirow{2}{*}{Qwen-7B}
            & Cov           & 0.53                        & $+0.05$            & $+0.08$ \\
            & \cellcolor{ourrow}\textbf{Ours} & \cellcolor{ourrow}$\mathbf{8e{-4}}$  & \cellcolor{ourrow}$\mathbf{+0.39}$   & \cellcolor{ourrow}$\mathbf{+0.72}$ \\
        \midrule
        \multirow{2}{*}{Math-7B}
            & Cov           & 0.29                        & $+0.03$            & $+0.06$ \\
            & \cellcolor{ourrow}\textbf{Ours} & \cellcolor{ourrow}$\mathbf{4e{-4}}$  & \cellcolor{ourrow}$\mathbf{+0.42}$   & \cellcolor{ourrow}$\mathbf{+0.61}$ \\
        \bottomrule
    \end{tabular}
    \vspace{-0.8em}
    \caption{Accuracy of entropy change estimation: the baseline \textit{Cov} vs. our estimator.}
    \label{MSE_and_PCC}
\vspace{-1.5em}
\end{table}

% \subsection{Analysis of Advantage–Probability Effects}\label{Advantage–Probability Effects}

\begin{figure}[t]
\vspace{-1em}
    \centering
    \includegraphics[width=\linewidth]{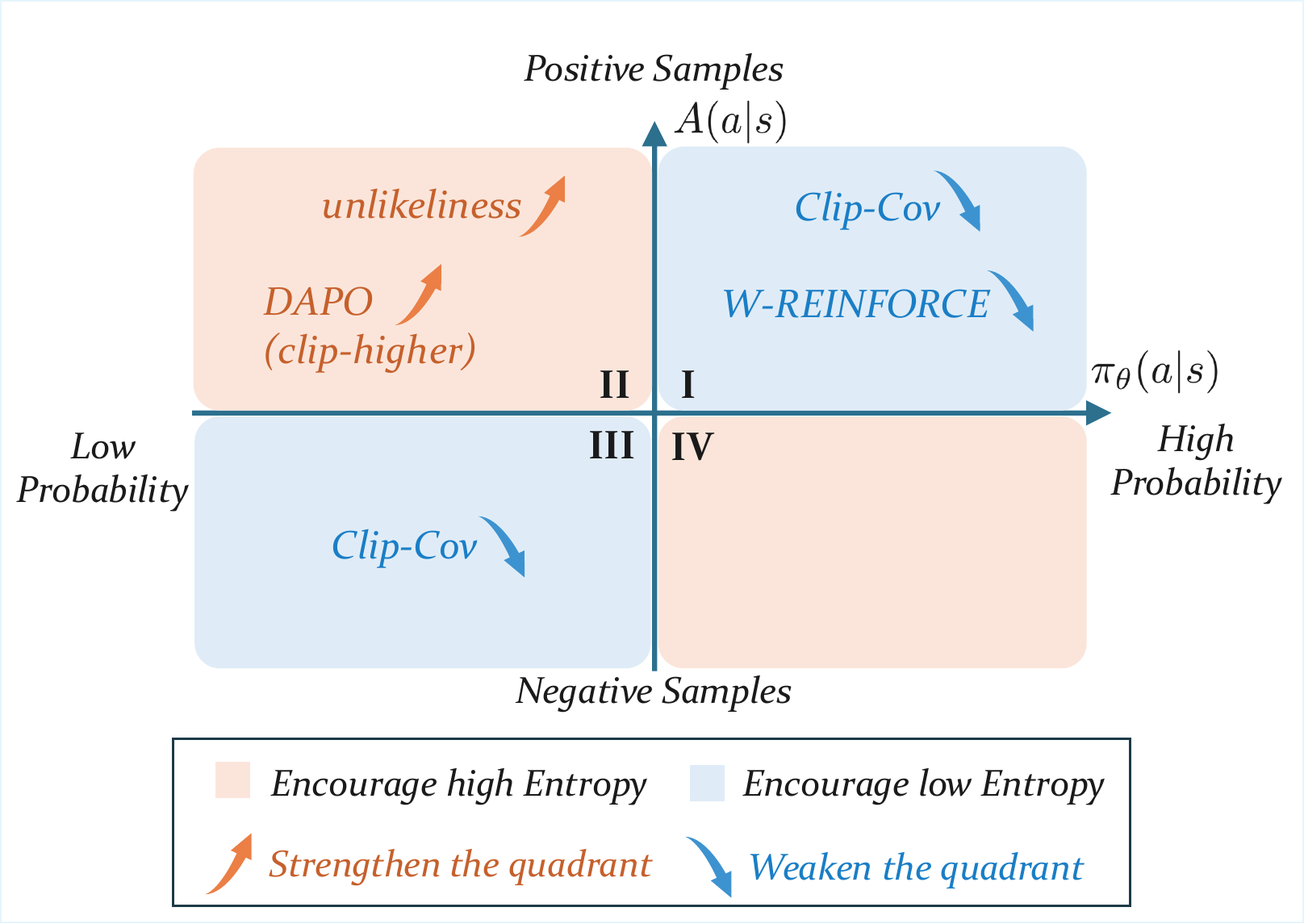}
    \vspace{-1.5em}
    \caption{Four-quadrant view of token-level entropy change direction with varying advantage and probability. Each token resides in one quadrant, and the macroscopic policy entropy represents the cumulative effect of all quadrants.}
    \label{Advantage_and_Probability}
    \vspace{-1.5em}
\end{figure}

\noindent\textbf{Remark 3 (Entropy Change Direction): }
We further investigate the direction of entropy change (increase or decrease), which is controlled by the joint effect of the advantage and token probability. Given that the function $\delta(\pi_{\theta}, \mathcal{H})$ takes negative values for high-probability tokens and positive values for low policy probability,  we conceptualize this joint effect in terms of four qualitative quadrants in the $(A, \pi_\theta)$ space shown in Figure~\ref{Advantage_and_Probability}:

\noindent\hypertarget{Quadrant I}{\textbf{Quadrant I:}} \textbf{Exploitation (entropy decrease).} For high-probability  tokens in correct outputs ($A>0, \delta<0$), rewarding a specialized behavior concentrates probability mass, thus \emph{decreasing} entropy.

\noindent\hypertarget{Quadrant II}{\textbf{Quadrant II:}} \textbf{Exploration (entropy increase).} For low-probability  tokens in correct outputs ($A>0, \delta>0$), rewarding a rare-but-correct behavior diversifies the policy, thereby \emph{increasing} entropy.
    
\noindent\hypertarget{Quadrant III}{\textbf{Quadrant III:}} \textbf{Suppression (entropy decrease).} For low-probability tokens in incorrect outputs ($A<0, \delta>0$), penalizing rare behaviors further concentrates the probability mass, thereby \emph{decreasing} entropy.

\noindent\hypertarget{Quadrant IV}{\textbf{Quadrant IV:}} \textbf{Error-Correction (entropy increase).} For high-probability tokens in incorrect outputs ($A<0, \delta<0$), penalizing overconfident errors flattens the distribution to encourage seeking alternatives and tends to \emph{increase} entropy.

% We next ask whether these theoretical findings of quadrant-level tendencies can be used to actively steer entropy in practice.
% Guided by the above quadrant-level tendencies, we design a simple intervention on each of the quadrants and the results strongly support our analysis.
% Comprehensive empirical studies are provided in the Appendix~\ref{appendix_strength_weaken}
We further perform empirical analyses by intervening on the samples in each quadrant.
Our empirical observations are closely aligned with our theoretical findings. Readers may refer to Appendix~\ref{appendix_strength_weaken} for additional details and results.

\begin{figure*}[htbp]
% \vspace{-1em}
    \centering
    \begin{subfigure}[t]{0.22\textwidth}
        \centering
    \includegraphics[width=\textwidth, height=2.2cm]{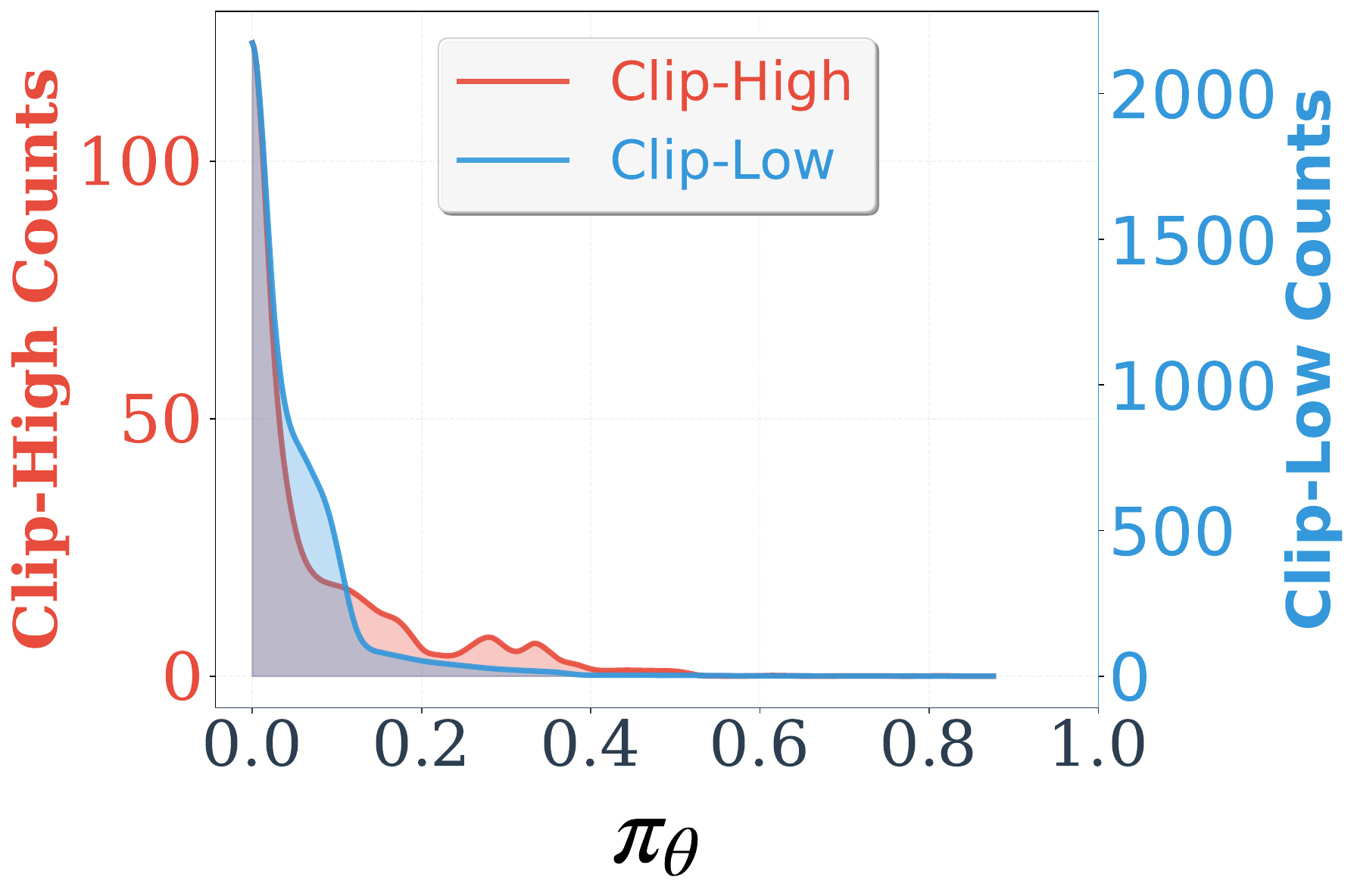}
    \caption{Clipping statistics}
    \label{Clip Counts}
    \end{subfigure}
    % (a) clip-high
    \begin{subfigure}[t]{0.25\textwidth}
        \centering
        \includegraphics[width=\linewidth,height=2.3cm]{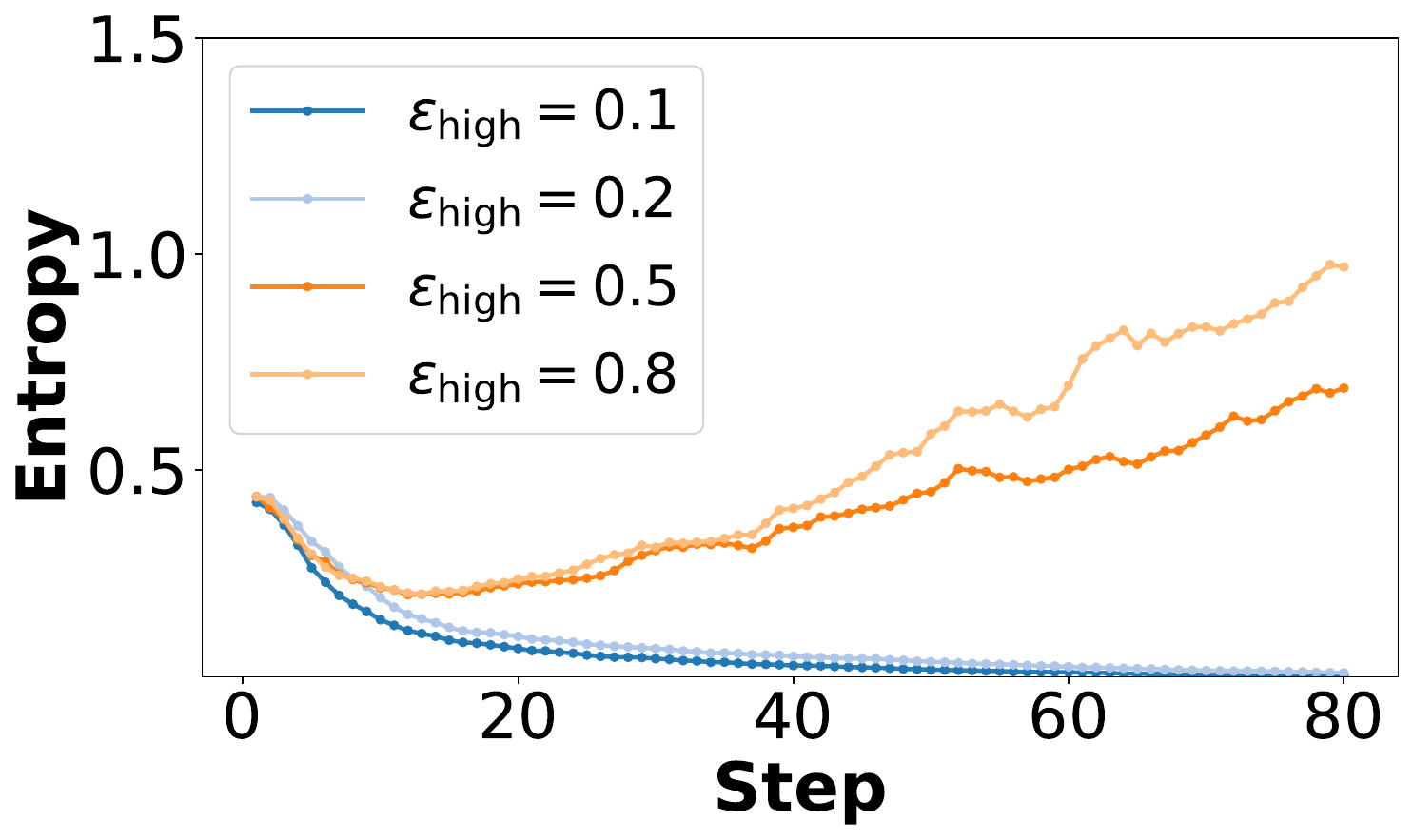}
        \caption{Entropy effect of $\varepsilon_{\text{high}}$}
        \label{clip-high}
    \end{subfigure}
    \hfill
    % (b) clip-low
    \begin{subfigure}[t]{0.25\textwidth}
        \centering
        \includegraphics[width=\linewidth,height=2.3cm]{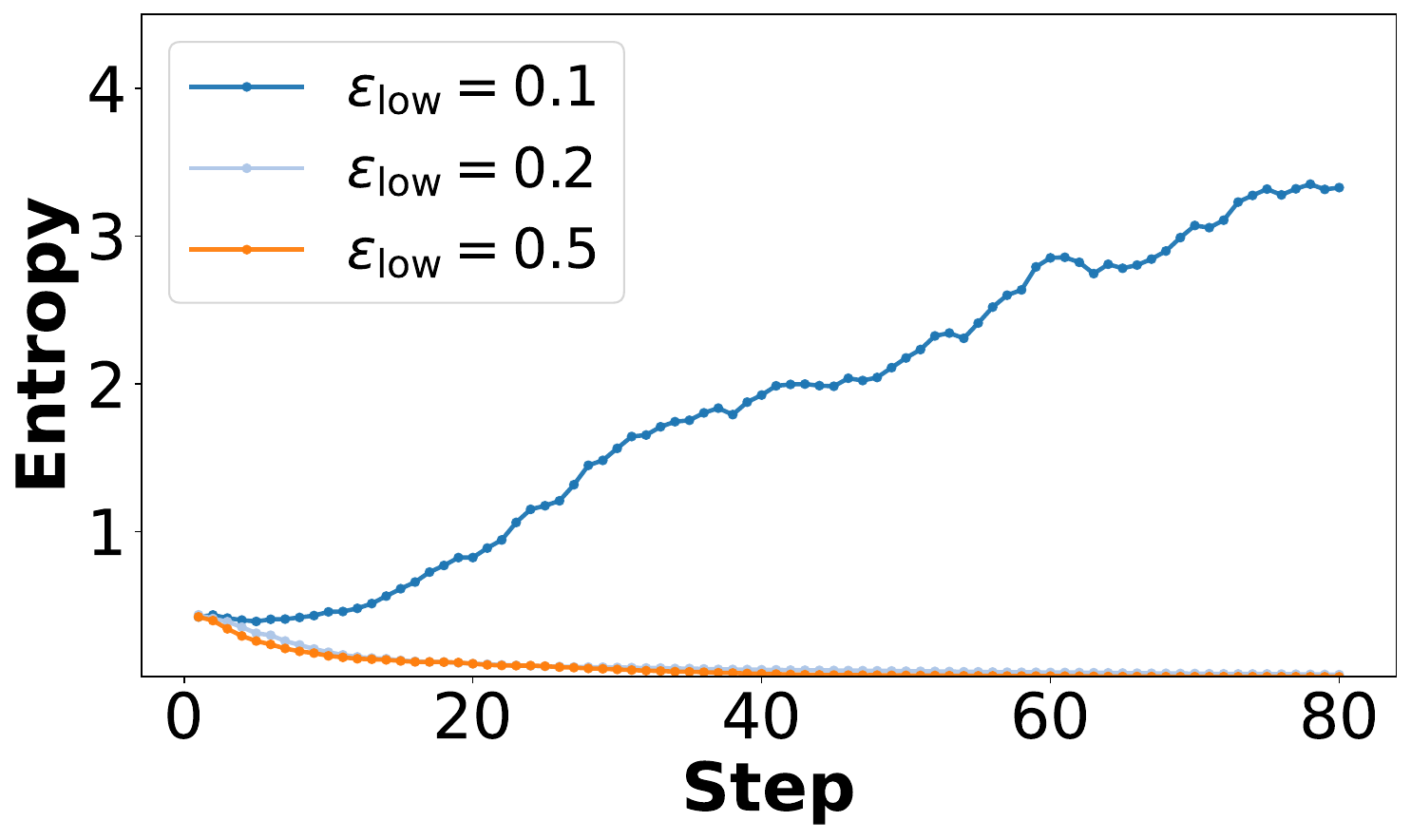}
        \caption{Entropy effect of $\varepsilon_{\text{low}}$}
        \label{clip-low}
    \end{subfigure}
    \hfill
    % (c) PSR-NSR
    \begin{subfigure}[t]{0.25\textwidth}
        \centering
        \includegraphics[width=\linewidth,height=2.3cm]{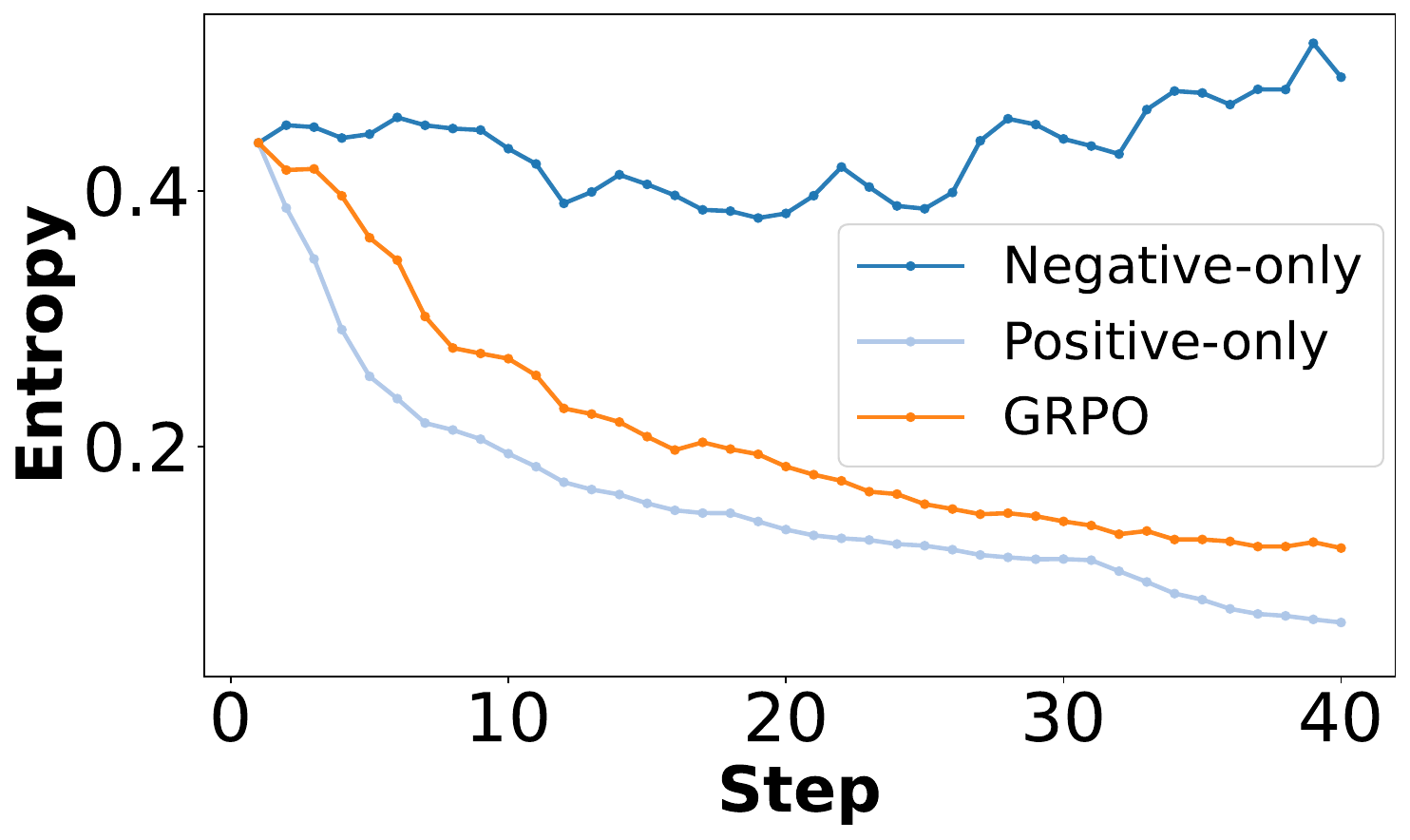}
        \caption{Positive/Negative-only}
        \label{PSR-NSR}
    \end{subfigure}
    \vspace{-0.5em}
    \caption{Empirical validation of entropy-change mechanisms of Clip-Higher and Positive-Reweighting.}
    \label{entropy-psr-nsr-all}
    \vspace{-1.2em}
\end{figure*}

% 在标准的强化学习流程中，四种情景并存，策略熵在熵增与熵减的合力下动态演化。熵崩溃现象可被视为由“利用”情景主导的熵减效应，压倒了由“探索”和“纠错”情景贡献的熵增效应。为了更好的说明，下面通过正负样本权重以及clip操作对熵的影响来阐释。 % 若训练数据仅包含正样本，则策略更新将不包含负样本驱动的“纠错”机制。这意味着模型无法有效修正其高置信度的错误预测，增加了陷入局部最优（Mode Collapse）的风险。因为缺乏了最主要的熵增来源之一（惩罚高概率错误），由“利用”驱动的熵减效应可能在训练中占据主导，导致策略过早收敛。 % 这也符合XX论文的实验现象 % 若训练数据仅包含负样本，策略更新将由“纠错”和“抑制”两种情景主导。其中，由“纠错”情景（惩罚高概率错误）带来的熵增效应是主要驱动力。因此，仅使用负样本成为一种强效的熵维持（Entropy Maintenance）机制，能有效防止策略熵的过快衰减。然而，由于缺乏正向奖励信号的引导，策略可能难以高效地收敛至最优解，仅能学到“何事不可为”，而对“何事可为”的认知不足。

% In standard RLVR, the four parts (rewarding/punishing × high/low probability) co-exist, and policy entropy evolves from the superposition of entropy-increasing and entropy-decreasing updates.
% Entropy collapse can be viewed as exploitation-driven entropy decreases overwhelming the exploration-driven entropy increases and error-correction.
% Building on the above insights, the entropy effects of positive/negative rebalance and ratio clipping can be incorporated into our analysis.
In the RLVR process, these four quadrant-level dynamics coexist as competing forces that shape the policy.
The global policy entropy evolves from the combined effect of these competing updates.
Consequently, entropy collapse can be understood as a state where the exploitation-driven, entropy-decreasing updates (Quadrants \hyperlink{Quadrant I}{I} and \hyperlink{Quadrant III}{III}) consistently overwhelm the exploration-driven, entropy-increasing updates (Quadrants \hyperlink{Quadrant II}{II} and \hyperlink{Quadrant IV}{IV}).
% This quadrant-based advantage–probability effect not only explains entropy collapse but also offers a unified basis for analyzing other interventions.

\section{Analyses on Existing Entropy Intervention Methods}
\vspace{-0.1em}
Building on the above theoretical findings, we conduct a comprehensive analysis of existing entropy-intervention techniques to reveal their underlying mechanisms and limitations.

\paragraph{Entropy Effect of Clip-Higher.}
Recent work ~\citep{yu2025dapo} decouples the lower and higher clipping bounds with $\varepsilon_{\text{high}}$ and $\varepsilon_{\text{low}}$, and shows that increasing $\varepsilon_{\text{high}}$ can mitigate entropy collapse.

\noindent\emph{Mechanism: }
The underlying mechanism can be explained as follows: from the importance ratio $r = \frac{\pi_\theta}{\pi_{\text{old}}}$, the ratio is more likely to attain a large value when $\pi_{old}$ is relatively small.
Consequently, clipping is predominantly triggered for low-probability tokens, a phenomenon confirmed by our empirical observations in Figure~\ref{Clip Counts}, which counts the clipping events in the first \(10\) steps of GRPO.
In this case, the clipping via $\varepsilon_{\text{high}}$ acts as a filter that removes a considerable number of low-probability positive samples, which typically fall within Quadrant~\hyperlink{Quadrant II}{II}. Increasing $\varepsilon_{\text{high}}$ therefore reduces the number of filtered instances, allowing more samples to contribute to entropy increase, thereby mitigating collapse.
A similar reasoning applies to the role of $\varepsilon_{\text{low}}$:  increasing $\varepsilon_{\text{low}}$ filters fewer instances in Quadrant~\hyperlink{Quadrant III}{III} which in turn exacerbates entropy collapse.

\noindent\emph{Empirical Evidence: } We further validate these observations through empirical experiments.
Figure~\ref{clip-high} and Figure~\ref{clip-low} illustrate the impact of varying $\varepsilon_{\text{high}}$ and $\varepsilon_{\text{low}}$. We observe that increasing $\varepsilon_{\text{high}}$ can mitigate or  even reverse entropy collapse, while increasing $\varepsilon_{\text{low}}$ intensifies collapse.

\noindent\emph{Limitation:} Briefly, the effect of tuning clipping thresholds on entropy dynamics can be interpreted as reweighting tokens in Quadrant~\hyperlink{Quadrant II}{II} and Quadrant~\hyperlink{Quadrant III}{III} in Figure~\ref{Advantage_and_Probability}.
However, such heuristic methods remain coarse-grained: DAPO, for example, seeks to affect entropy by controlling the updates of some tokens in Quadrant~\hyperlink{Quadrant II}{II}, without explicitly controlling tokens in other quadrants.

\paragraph{Entropy Effect of Re-weighting Positives.}
Recent studies have shown that re-weighting positive samples can mitigate entropy collapse.
For example, unlikeliness~\citep{he2025rewarding} up-weighting tokens in Quadrant~\hyperlink{Quadrant II}{II} and down-weighting tokens in Quadrant~\hyperlink{Quadrant I}{I}, while W-REINFORCE~\citep{zhu2025surprising} down-weighting all positives in training.

\noindent\emph{Mechanism: }
The effect of unlikeliness is clear from Figure~\ref{Advantage_and_Probability}: it strengthens the entropy-increasing quadrant and weakens entropy-decreasing quadrant, thereby increasing policy entropy.
As for W-REINFORCE, the key insight is that token-level updates are dominated by high-probability tokens since they are more likely to be sampled.
Therefore, down-weighting all positives primarily weakens the entropy-decreasing contribution from Quadrant~\hyperlink{Quadrant I}{I}, thereby increasing policy entropy.

\noindent\emph{Empirical Evidence:}
To validate this mechanism, we conduct positive-only and negative-only experiments in GRPO setting, where the policy is trained using only positive or only negative samples, respectively.
The results, shown in Figure~\ref{PSR-NSR}, are consistent with the mechanism: training only on positive samples rapidly collapses policy entropy, while training only on negative samples sustains consistently high entropy.

\noindent\emph{Limitation: } While these positive re-weighting methods improve global entropy, they only reweight a subset of tokens, leaving the remaining tokens still vulnerable to entropy collapse.

\paragraph{Entropy Effect of Entropy-aware Advantage.}
Several studies have proposed incorporating entropy-related terms into the advantage function to mitigate entropy collapse, such as Entro. Adv.~\citep{cheng2025reasoning} and GTPO~\citep{tan2025gtpo}.
Such methods assign larger advantages to tokens with higher entropy. However, our findings indicate that these methods are not universally effective;
in fact, they can sometimes exacerbate entropy collapse instead of mitigating it.

\begin{figure}[t]
    \centering
    \begin{subfigure}[t]{0.48\linewidth}
        \centering
\includegraphics[width=\linewidth]{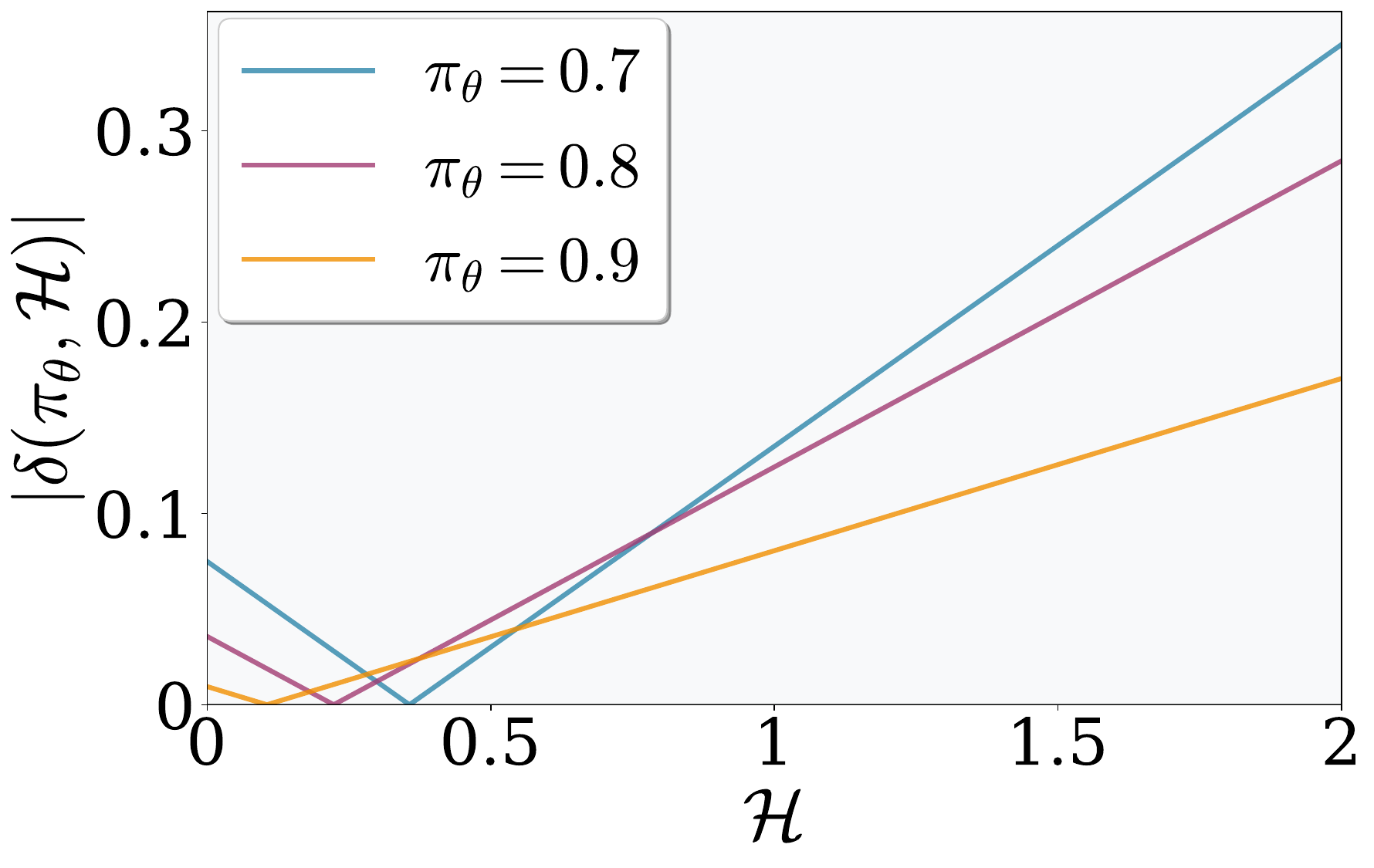}
        \caption{Theoretical correlation}
        \label{Theoretical correlation.}
    \end{subfigure}
    \begin{subfigure}[t]{0.48\linewidth}
        \centering
\includegraphics[width=\linewidth]{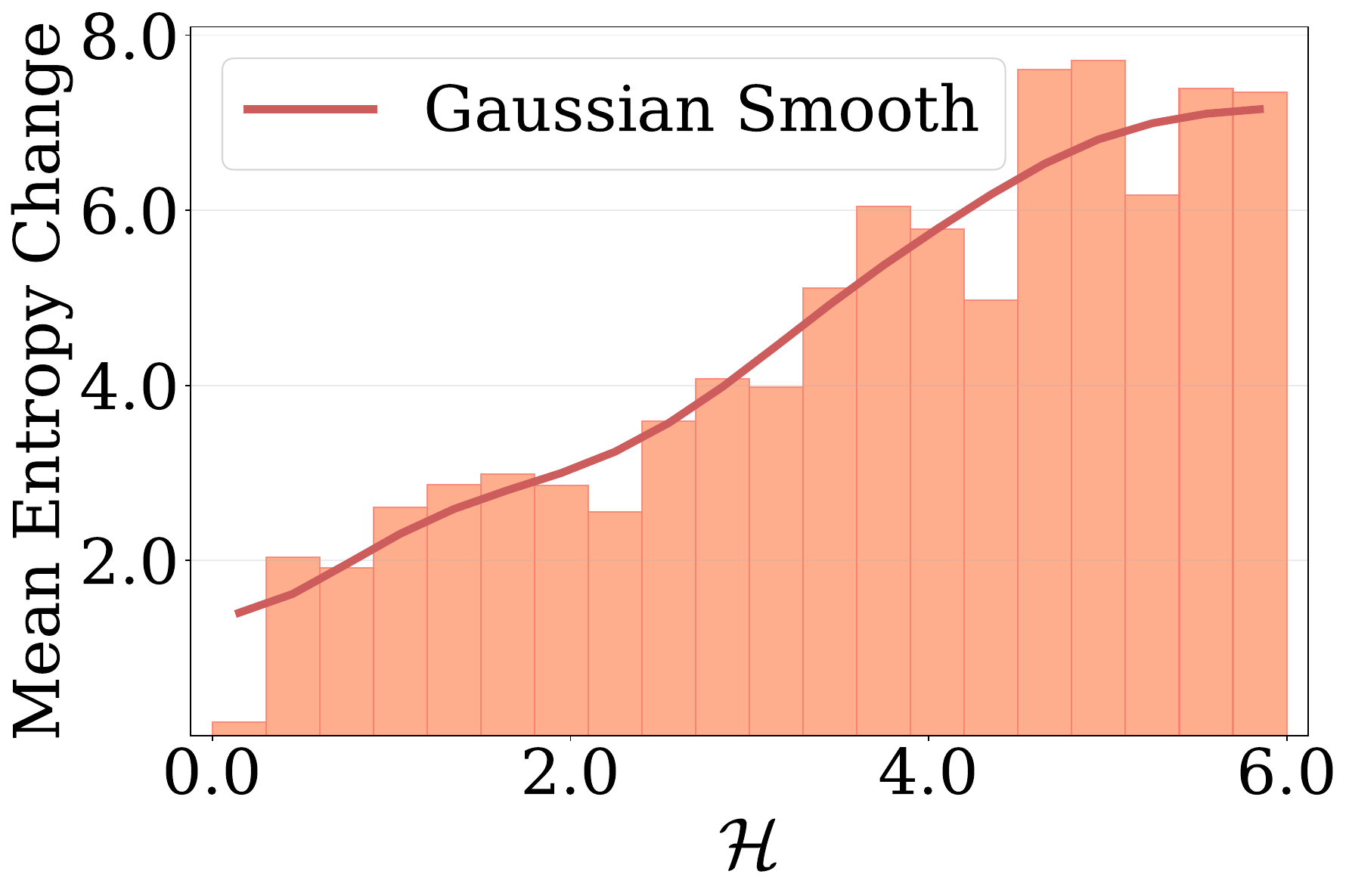}
        \caption{Empirical behavior}
        \label{Empirical behavior.}
    \end{subfigure}
    \vspace{-0.8em}
    \caption{Entropy change as a function of entropy.}
    \label{entropy_entropy_change}
    \vspace{-1em}
\end{figure}

\noindent\emph{Mechanism: } 
Based on Eq.~\eqref{delta_def}, we plot $\delta(\pi_{\theta}, \mathcal{H})$ as a function of $\mathcal{H}$ in Figure~\ref{Theoretical correlation.}.
We can infer that high-entropy tokens tend to induce larger entropy changes.
This mechanism is confirmed by empirical behavior during training in  Figure~\ref{Empirical behavior.}, which tracks the mean entropy change over the first $50$ training steps: tokens with higher entropy exhibit larger entropy changes.
Consequently, assigning greater advantages to high-entropy tokens can amplify their entropy change.
When the policy enters an entropy-decreasing phase, this amplification exacerbates the collapse rather than mitigating it.
% If these tokens tend to decrease entropy, their amplified contribution intensifies the collapse instead of mitigating it.

\begin{figure}[t]
    \centering
    \begin{subfigure}[t]{0.48\linewidth}
        \centering
\includegraphics[width=\linewidth]{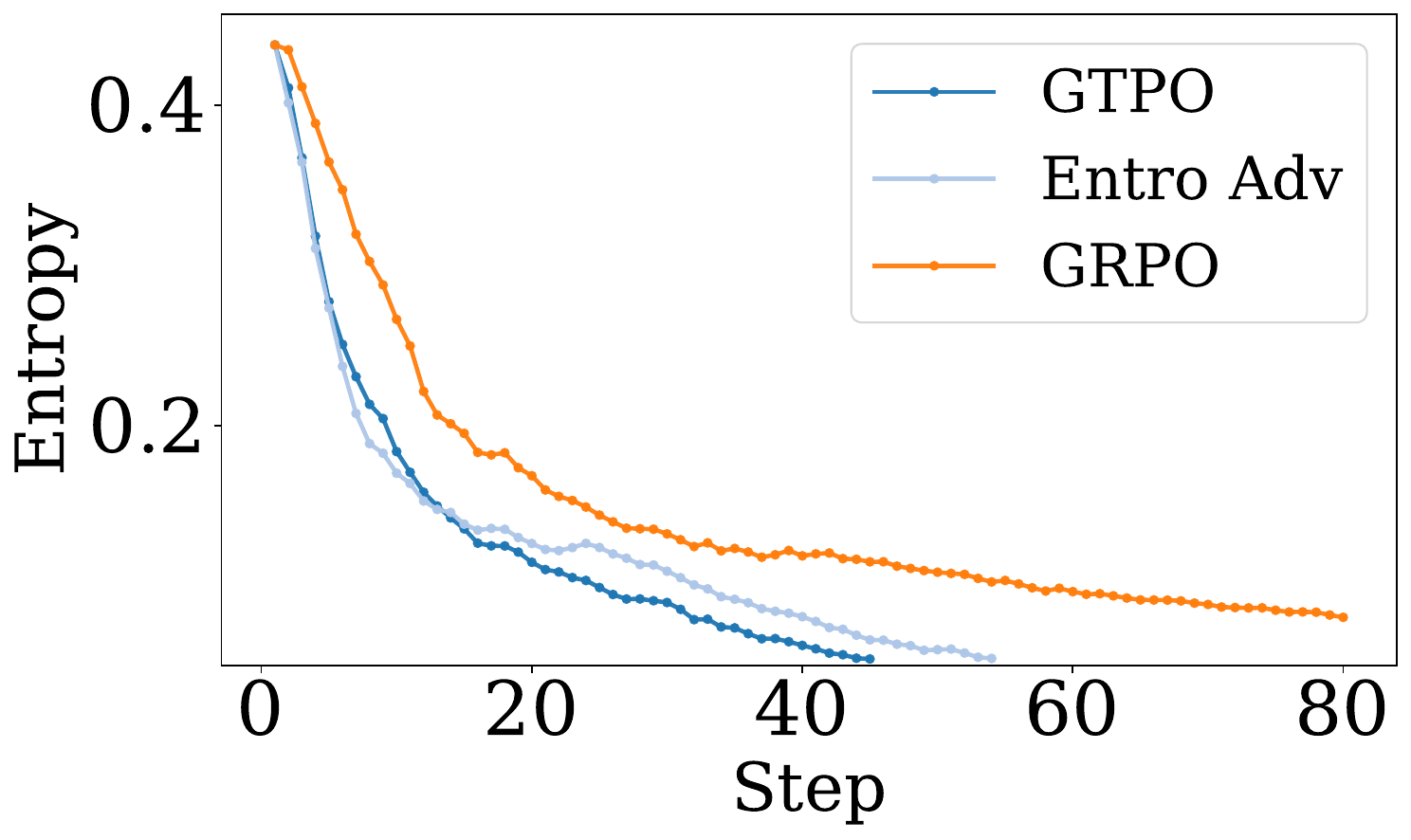}
        \caption{On dataset DAPO-17k}
    \end{subfigure}
    \begin{subfigure}[t]{0.48\linewidth}
        \centering
\includegraphics[width=\linewidth]{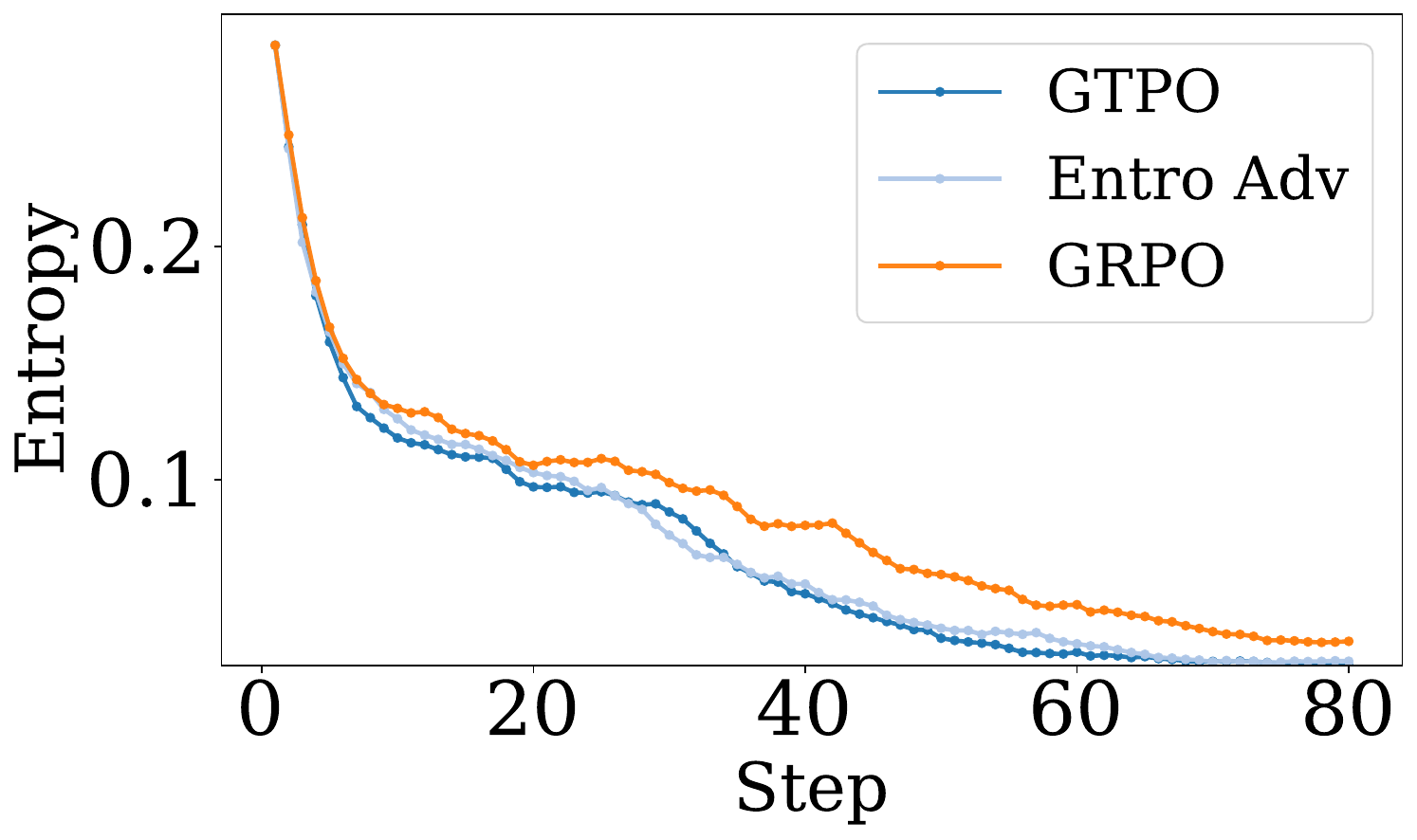}
        \caption{On dataset Math}
    \end{subfigure}
    \vspace{-0.6em}
    \caption{Entropy dynamics with entropy-aware advantage on Qwen2.5-Math-7B backbone.}
    \label{Entropy dynamics with advantage shaping.}
    \vspace{-1.5em}
\end{figure}

\noindent\emph{Empirical Evidence: } We validate this mechanism by examining the entropy dynamics of these methods in Figure~\ref{Entropy dynamics with advantage shaping.}:
compared to GRPO, Entro. Adv. and GTPO both exhibit faster entropy collapse when the policy enters an entropy-decreasing phase.
When policy entropy starts to decline, these methods can even accelerate entropy collapse.

\noindent\emph{Limitation: }This finding highlights a flaw in these methods: rather than reliably encouraging exploration, they may instead aggravate entropy decline.

\paragraph{Limitations of Existing Methods.} While the aforementioned entropy-intervention strategies can partly alleviate entropy collapse in practice, they remain largely heuristic and lack principled guidance.
These strategies involve qualitative adjustments to only  one or two factors in Eq.~\eqref{Omega}, as shown in Table~\ref{factors}, leaving other relevant factors unconsidered and failing to capture their joint impacts on the entropy dynamics.  As a result, there is a substantial gap between their interventions and the actual entropy evolution observed during training. Their effectiveness is compromised.

% \newcommand{\cmark}{\ding{51}} % ✓
% \newcommand{\xmark}{\ding{55}} % ✗
% \begin{table}[htbp]
% \vspace{-0.5em}
%     \centering
%     \setlength{\tabcolsep}{8pt}
%     \renewcommand{\arraystretch}{0.5}
%     \begin{tabular}{lcccc}
%         \toprule
%         \textbf{Method} &$\mathbb{I}_{\text{clip}}$ & $\pi_\theta$ & $A$ & $\mathcal{H}(s)$ \\
%         \midrule
%         DAPO      & \cmark     & \xmark & \xmark & \xmark \\
%         \midrule
%         Unlikeliness  & \xmark   & \cmark & \cmark & \xmark \\
%         \midrule
%         W-REINFORCE  & \xmark    & \xmark & \cmark & \xmark \\
%         \midrule
%         Entropy Adv.  & \xmark   & \xmark & \cmark & \cmark \\
%         \midrule
%         KL Reg.     & \xmark     & \cmark & \xmark & \xmark \\
%         \midrule
%         Entropy Reg.  &\xmark    & \xmark & \xmark & \cmark \\
%         \midrule
%         Edge-GRPO    &\xmark     & \xmark & \xmark & \cmark \\
%         \midrule
%         Forking Tokens  &\xmark  & \xmark & \xmark & \cmark \\
%         \midrule
%         Clip-Cov     &\xmark     & \cmark & \cmark & \xmark \\
%         \midrule
%         \textbf{STEER}  &\cmark  & \cmark & \cmark & \cmark \\
%         \bottomrule
%     \end{tabular}
%     \vspace{-0.5em}
%     \caption{Factors considered by existing entropy-intervention methods and ours.}
%     \label{factors}
% \vspace{-1em}
% \end{table}

\newcommand{\cmark}{{\color{black!80}\ding{51}}}
\newcommand{\xmark}{{\color{black!30}\ding{55}}}
\begin{table}[htbp]
    \centering
    \setlength{\tabcolsep}{10pt}
    \renewcommand{\arraystretch}{1}
    \begin{tabular}{@{}lcccc@{}}
        \toprule
        \textbf{Method} & $\mathbb{I}_{\text{clip}}$ & $\pi_\theta$ & $A$ & $\mathcal{H}(s)$ \\
        \midrule
        DAPO             & \cmark & \xmark & \xmark & \xmark \\
        Unlikeliness     & \xmark & \cmark & \cmark & \xmark \\
        W-REINFORCE      & \xmark & \xmark & \cmark & \xmark \\
        Entro. Adv.     & \xmark & \xmark & \cmark & \cmark \\
        KL Reg.          & \xmark & \cmark & \xmark & \xmark \\
        Entropy Reg.     & \xmark & \xmark & \xmark & \cmark \\
        Edge-GRPO        & \xmark & \xmark & \xmark & \cmark \\
        Forking Tokens   & \xmark & \xmark & \xmark & \cmark \\
        Clip-Cov         & \xmark & \cmark & \cmark & \xmark \\
        \midrule
        \textbf{STEER}   & \cmark & \cmark & \cmark & \cmark \\
        \bottomrule
    \end{tabular}
    \vspace{-0.5em}
    \caption{Factors that governing entropy change considered by existing entropy-intervention methods and ours.}
    \label{factors}
\vspace{-1.5em}
\end{table}

\begin{table*}[t]
\vspace{-1em}
\centering
\definecolor{ourrow}{RGB}{235,243,252}
\fontsize{10}{13.2}\selectfont
{\setlength{\tabcolsep}{6pt}
\renewcommand{\arraystretch}{0.75}
\vspace{-0.5em}
\begin{tabular}{lccccccc}
\toprule
\textbf{Method} & \textbf{AIME24} & \textbf{AIME25} & \textbf{AMC23} & \textbf{MATH500} & \textbf{Minerva} & \textbf{Olympiad} & \textbf{Avg.} \\
\midrule
Qwen2.5-Math-7B & $13.8$ & $5.3$ & $44.6$ & $39.6$ & $9.9$ & $13.8$ & $21.2$ \\
\midrule
\multicolumn{8}{c}{\textbf{Classical RLVR Methods}} \\
\midrule
GRPO                    & $28.0$ & $14.3$ & $66.2$ & $78.6$ & $37.3$ & $40.9$ & $44.2$ \\
SimpleRL-Zoo            & $30.8$ & $14.2$ & $65.4$ & $79.2$ & $37.1$ & $40.8$ & $44.6$ \\
Eurus-PRIME             & $20.9$ & $13.0$ & $65.2$ & $79.8$ & $37.4$ & $40.6$ & $42.8$ \\
OPO                     & $32.2$ & $13.4$ & $71.5$ & $82.2$ & $38.2$ & $41.0$ & $46.4$ \\
\midrule
\multicolumn{8}{c}{\textbf{Entropy Intervention Methods}} \\
\midrule
GRPO w/ clip-high       & $31.7$ & $12.8$ & $66.8$ & $79.0$ & $38.6$ & $39.3$ & $44.7$ \\
GRPO w/ Entro. Loss     & $29.1$ & $14.0$ & $67.6$ & $80.0$ & $38.2$ & $37.9$ & $44.5$ \\
GRPO w/ Fork Tokens     & $31.9$ & $14.3$ & $65.5$ & $79.2$ & $37.1$ & $40.9$ & $44.8$ \\
W-REINFORCE             & $29.2$ & $12.0$ & $64.9$ & $79.2$ & $37.8$ & $40.9$ & $44.0$ \\
Entro. Adv.             & $27.5$ & $13.5$ & $70.2$ & $79.6$ & $36.8$ & $42.8$ & $45.1$ \\
Clip--Cov               & $32.5$ & $12.9$ & $68.4$ & $78.0$ & $40.8$ & $41.3$ & $45.7$ \\
KL--Cov                 & $32.8$ & $14.1$ & $64.2$ & $78.8$ & $37.1$ & $39.4$ & $44.4$ \\
\rowcolor{ourrow} \textbf{STEER (Ours)}  & $\mathbf{36.2}$ & $\mathbf{16.1}$ & $\mathbf{72.1}$ & $\mathbf{82.2}$ & $\mathbf{41.7}$ & $\mathbf{43.0}$ & $\mathbf{48.6}$ \\
\bottomrule
\end{tabular}
\vspace{-0.3em}
\caption{Benchmark results of different methods on Qwen2.5-Math-7B. We report avg@32 for AIME24, AIME25, and AMC23 and avg@1 for other datasets. All results are presented as percentages.}
\label{main_results}
\vspace{-1.5em}
}
\end{table*}

\section{Stabilizing Token-level Entropy-change via Reweighting}
% The preceding analysis motivates us to  develop a more fine-grained and theoretically-driven strategy, termed STEER.
% The key idea is to introduce an adaptive weight $\lambda(s)$ for each token in RLVR to regulate its entropy change. In this way, the entropy change associated with each token is modulated from $\Omega(s)$ to $\lambda(s) \Omega(s)$, thus enabling fine‑grained per‑token entropy modulation rather than only monitoring the global expectation as in prior methods.

% The central question becomes how to design a suitable $\lambda(s)$. Naturally,  $\lambda(s)$ should vary according to $\Omega(s)$ ---  larger entropy changes indicate a higher risk of entropy collapse and should be attenuated.
% To achieve this, we employ an exponential decay function:

The preceding analysis motivates us to  develop a fine-grained and theoretically-grounded method, termed STEER.
The key idea is to directly leverage the theoretically-estimated entropy change $\Omega(s)$, translating it into an adaptive token-level scaling weight $\lambda(s)$, incorporated in the loss function for each token. Accordingly, for tokens susceptible to drastic entropy decay, STEER actively attenuates their gradient contribution, approximately modulating their entropy variations from $\Omega(s)$ to $\lambda(s) \Omega(s)$.  This mechanism empowers STEER to execute precise token-level entropy modulation, moving beyond the coarse monitoring of global expectations by prior methods. 

To achieve this, we introduce $\lambda(s)$ through a simple exponential decay function:
%\vspace{-0.2em}
\begin{equation}\label{weight_mapping}
\lambda(s)
= \exp\!\left(  \, -\alpha \,
\frac{ |\Omega(s)|}{\max_{s \in \mathcal{B}} |\Omega(s)| } \,
\right),
\end{equation}
%\vspace{-0.3em}
where $\alpha > 0$ is the hyperparameter  controlling the decay rate.
This formulation satisfies two desired properties: (1) it is monotonically decreasing with respect to the normalized entropy change; (2) the exponential ensures that all weights remain strictly positive.
Here we apply the absolute value $|\Omega(s)|$ to measure the magnitude of entropy-change, since large entropy increases may also be sub‑optimal.
Controlling both increases and decreases within a stable range leads to more stable training dynamics.
The normalization by the batch maximum ensures numerical stability. 

The hyperparameter $\alpha$ governs the slope of the decay curve.
Equivalently, $\alpha$ determines the minimum attainable weight, $\lambda_{\min}=\exp(-\alpha)$.
Thus, Eq.~\eqref{weight_mapping} can be interpreted as an inverse mapping of entropy changes into the range $[\lambda_{\min}, 1]$, with  larger entropy changes corresponding to smaller weights. In practice, we find it more convenient to tune $\lambda_{\min}$ instead of $\alpha$, due to its better interpretability.

% Overall, STEER offers several compelling advantages:  (1) {Theoretically Grounded:} Built upon our rigorous theoretical analyses, $\Omega(s)$ serves as a precise estimation of token-level entropy variations. This empowers STEER to adaptively pinpoint tokens susceptible to drastic entropy change and impose fine-grained reductions on their update magnitudes. (2) {Lightweight and Plug-and-Play:} STEER incurs minimal computational cost and can be easily integrated into existing methods. It only involves the additional calculation of $\Omega(s)$, the complexity of which is negligible. Notably, STEER can be seamlessly applied to various LLM architectures and RL algorithms by simply incorporating a token-level reweighting coefficient.
Notably, STEER also enjoys significant practical merits, incurring minimal computational cost and allowing for easy integration into existing methods. It only involves the additional calculation of $\Omega(s)$, the complexity of which is negligible. Besides, STEER can be seamlessly applied to various LLM architectures and RL algorithms.

\section{Experiments and Results}
In this section, we evaluate STEER from the following perspectives.
First, we compare STEER against strong baselines on six math reasoning benchmarks and three coding benchmarks.
Second, we examine STEER's entropy control ability in RLVR training.
Third, we test STEER's generalization across multiple model scales, families, and RL algorithms.
% Fourth, we conduct sensitivity analyses on hyperparameters and design choices.
Further results are deferred to Appendix~\ref{Supplementary Performance Evaluation}.

\begin{table*}[t]
\vspace{-0.5em}
\centering
\definecolor{ourrow}{RGB}{235,243,252}
\fontsize{10}{13.2}\selectfont
{\setlength{\tabcolsep}{6pt}
\renewcommand{\arraystretch}{0.75}
\vspace{-0.5em}
\begin{tabular}{lccccccc}
\toprule
\textbf{Method} & \textbf{AIME24} & \textbf{AIME25} & \textbf{AMC23} & \textbf{MATH500} & \textbf{Minerva} & \textbf{Olympiad} & \textbf{Avg.} \\
\midrule
Base         & $3.9$  & $2.6$  & $25.8$ & $52.6$ & $15.4$ & $23.0$ & $20.6$ \\
GRPO         & $17.2$ & $13.2$ & $66.3$ & $80.6$ & $38.0$ & $42.2$ & $42.9$ \\
OPO          & $17.8$ & $12.6$ & $68.2$ & $78.6$ & $37.7$ & $42.6$ & $42.9$ \\
Entro. Adv.  & $14.6$ & $9.8$  & $65.6$ & $78.8$ & $36.5$ & $40.9$ & $41.0$ \\
Clip-Cov     & $14.1$ & $13.6$ & $59.8$ & $78.2$ & $38.6$ & $43.2$ & $41.2$ \\
\rowcolor{ourrow} \textbf{STEER} & $\mathbf{19.3}$ & $\mathbf{14.3}$ & $\mathbf{70.3}$ & $\mathbf{81.4}$ & $\mathbf{39.4}$ & $\mathbf{46.7}$ & $\mathbf{45.2}$ \\
\bottomrule
\vspace{-1.4em}
\end{tabular}
\caption{Benchmark results of different methods on Qwen2.5-14B. We report avg@32 for AIME24, AIME25, and AMC23 and avg@1 for other datasets. All results are presented as percentages.}
\label{14B_results}
\vspace{-1.2em}
}
\end{table*}

\subsection{Experimental Setup}
\label{setup}

\paragraph{Training.}
We follow the default  training recipe of standard GRPO in verl~\citep{sheng2025hybridflow}.
For fair comparisons, we closely follow the experimental setup used in recent work~\citep{yu2025dapo, wang2025scheduling}.
The only hyperparameter introduced by STEER, $\lambda_{\min}$, is set to $0.7$.
To ensure the reliability, all results reported in the main experiments are averaged over two independent training runs.

For math reasoning task, we conduct experiments on four different models, including Qwen2.5-Math-1.5B/7B, Qwen2.5-14B~\citep{yang2024qwen2} and Llama-3.2-3B~\citep{grattafiori2024llama}.
We train on the widely-used DAPO-Math-17k~\citep{yu2025dapo}, which consists of 17K prompts, each paired with a ground-truth integer answer.

For coding tasks, we conduct comparison experiments on four different models, including Qwen2.5-Coder-3B/7B/14B ~\citep{hui2024qwen2} and Mistral-7B~\citep{jiang2023mistral7b}.
For code generation task, we adopt ArcherCodeR\footnote{Available at \url{https://huggingface.co/datasets/Fate-Zero/ArcherCodeR-Dataset}.} for training.
As for code editing task, since there is no large-scale open-source dataset available, we adopt our in-house practical code editing benchmark for training.

\paragraph{Evaluation.} 
For math reasoning task, we evaluate our models and baselines on six widely used mathematical reasoning benchmarks: AIME24, AIME25, AMC23, MATH-500, Minerva Math, and OlympiadBench.
For code generation task, we adopt the widely used LiveCodeBench v5~\citep{jain2024livecodebench} for the evaluation of code generation.
We report avg@4 for code generation task following~\citep{wang2025stabilizing}.
For code editing, we evaluate the model on both our internal held-out test split and Zeta~\citep{zed_zeta_dataset_2025}.
We report exact-match accuracy for measuring code edit task.
More training and evaluation details of our method and baselines are listed in Appendix~\ref{training details}.

\paragraph{Baselines.}
For a thorough comparison, we compare STEER on math reasoning task against 11 baselines, including standard GRPO~\citep{shao2024deepseekmath}, SimpleRL-Zoo~\citep{zeng2025simplerl}, Eurus-PRIME~\citep{cui2025process}, OPO~\citep{hao2025policy}, GRPO with clip-high~\citep{yu2025dapo}, GRPO with entropy loss~\citep{schulman2017proximal}, GRPO with Fork Tokens~\citep{wang2025beyond}, W-REINFORCE~\citep{zhu2025surprising}, Entro. Adv.~\citep{cheng2025reasoning}, Clip-Cov and KL-Cov ~\citep{cui2025entropy}.
And we compare STEER with GRPO on coding tasks.
For all baselines, the default training hyperparameters in RLVR are consistent with STEER, while the newly introduced hyperparameters are configured following the original implementations, detailed in Appendex~\ref{Training Details for our method and baselines}.

\begin{table}[t]
\vspace{-0.3em}
    \centering
    \setlength{\tabcolsep}{10pt}
    \renewcommand{\arraystretch}{1}
    \small
    \definecolor{ourrow}{RGB}{235,243,252}
    \fontsize{10}{13.2}\selectfont
    \begin{tabular}{@{}llccc}
        \toprule
        \textbf{Dataset} & \textbf{Method} & \textbf{3B} & \textbf{7B} & \textbf{14B} \\
        \midrule
        \multirow{2}{*}{Internal}
            & GRPO  & $39.7$ & $40.1$ & $42.6$ \\
            & \cellcolor{ourrow}\textbf{STEER} & \cellcolor{ourrow}$\mathbf{41.2}$ & \cellcolor{ourrow}$\mathbf{41.9}$ & \cellcolor{ourrow}$\mathbf{45.1}$ \\
        \midrule
        \multirow{2}{*}{Zeta}
            & GRPO  & $17.4$ & $22.0$ & $22.3$ \\
            & \cellcolor{ourrow}\textbf{STEER} & \cellcolor{ourrow}$\mathbf{19.3}$ & \cellcolor{ourrow}$\mathbf{24.0}$ & \cellcolor{ourrow}$\mathbf{24.1}$ \\
        \midrule
        \multirow{2}{*}{LCB-v5}
            & GRPO  & $24.4$ & $28.5$ & $29.3$ \\
            & \cellcolor{ourrow}\textbf{STEER} & \cellcolor{ourrow}$\mathbf{24.9}$ & \cellcolor{ourrow}$\mathbf{29.2}$ & \cellcolor{ourrow}$\mathbf{31.8}$ \\
        \bottomrule
    \end{tabular}
    \vspace{-0.3em}
    \caption{Performance on real-world coding tasks.}
    \label{Performance on real-world coding tasks.}
    \vspace{-1.5em}
\end{table}

\subsection{Main Results}

\paragraph{Math Reasoning Tasks.}
The main results in math reasoning task are shown in Table~\ref{main_results}.
STEER outperforms classical RLVR baselines as well as existing entropy intervention baselines across all datasets.
STEER improves average performance by $2.2$ points over the second runner-up (OPO) and by $2.9$ points over the third runner-up (Clip-Cov) across all baselines.
The performance experiments on  Qwen2.5-14B shown in Table~\ref{14B_results} are compared with the top three competitors in Table~\ref{main_results} (i.e., OPO, Clip-Cov, and Entro. Adv.).
STEER also consistently achieves the highest average performance, demonstrating its superior generalization in improving math reasoning abilities.

\begin{figure}[htbp]
\vspace{-0.5em}
\centering
\includegraphics[width=0.95\linewidth]{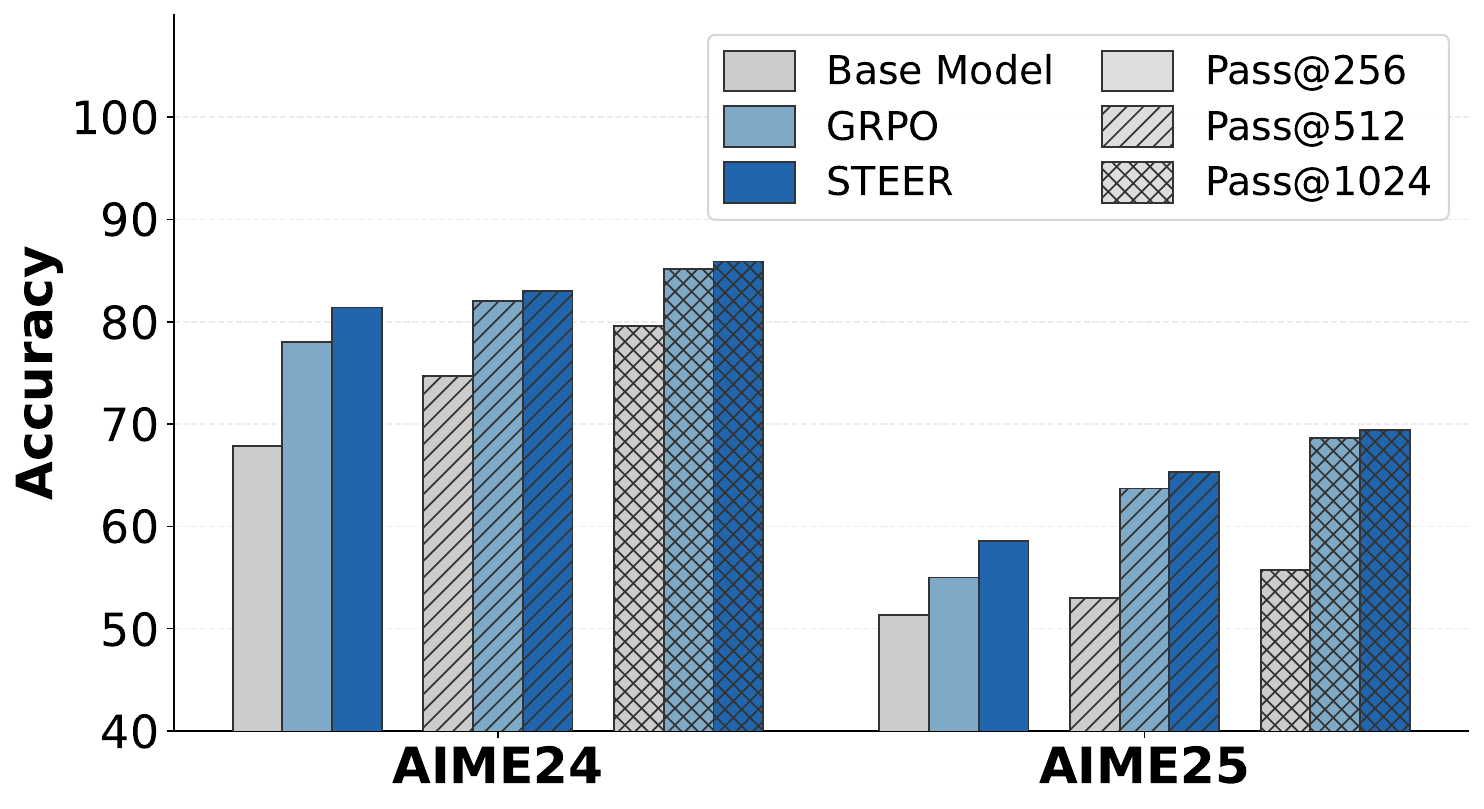}
\vspace{-0.5em}
\caption{Pass@256/512/1024 performance on AIME24 and AIME25.}
\label{fig:pass1024}
\vspace{-1em}
\end{figure}

Beyond metric avg@32 and avg@1 in Table~\ref{main_results}, the results of Pass@256/512/1024 in math reasoning are shown in Figure~\ref{fig:pass1024}.
As observed, STEER delivers the best Pass@256/512/1024 on both AIME24/25 and the highest average. By explicitly regulating entropy, STEER preserves sufficient exploration to discover more informative trajectories, thereby achieving better performance.

\paragraph{Coding Tasks.}
Beyond math reasoning, we evaluate our method on real-world code tasks.
Table~\ref{Performance on real-world coding tasks.} compares STEER with GRPO on three Qwen2.5-coder models.
We observe that STEER exceeds GRPO by at least $1\%$ in each model and test set.
These results demonstrate the superior performance of STEER in various scenarios.
More performance evaluation is detailed in Appendix~\ref{Supplementary Performance Evaluation}.

\begin{figure}[t]
    \centering
    \includegraphics[width=0.85\linewidth, height=3.7cm]{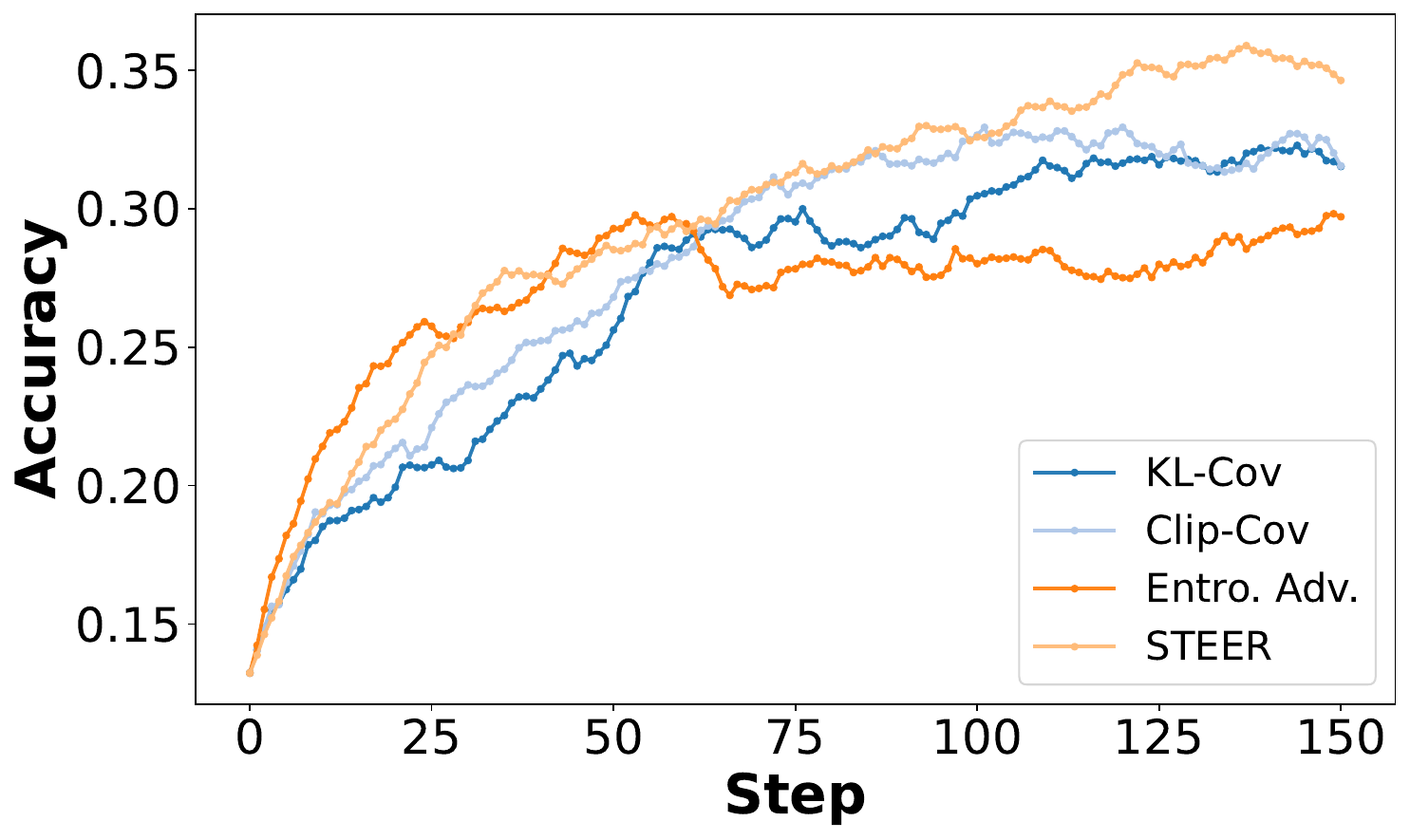}
    \vspace{-0.8em}
\captionsetup{justification=centering,singlelinecheck=false}
    \caption{Test accuracy dynamics comparison.}
\label{test_acc_plot_EMA_s0.8_max150}
    \vspace{-1.2em}
\end{figure}

\paragraph{Training Dynamics.}
Figure~\ref{test_acc_plot_EMA_s0.8_max150} shows the curves of test accuracy during training. 
Notably, STEER exhibits a stable upward trajectory, ultimately achieving superior final performance compared to the baselines.

% We then investigate whether better entropy regulation increases the probability of rare-but-correct trajectories that the base model hardly reaches even with large sampling budgets.

% 主表 Table 1（Qwen2.5-Math-7B on 6 math benchmarks, 10 baselines）
% 关键 claims:
% - STEER best on all 6
% - +2.2 over OPO (strongest baseline)
% - +2.9 over Clip-Cov (strongest entropy-intervention)

% Figure: test curves

% === 延伸实验 ===

\begin{table*}[t]
\vspace{-0.5em}
\centering
\definecolor{ourrow}{RGB}{235,243,252}
\fontsize{10}{13.2}\selectfont
{\setlength{\tabcolsep}{6pt}
\renewcommand{\arraystretch}{0.78}
\begin{tabular}{lcccccccc}
\toprule
\textbf{Method} & \textbf{AIME24} & \textbf{AIME25} & \textbf{AMC23} & \textbf{MATH500} & \textbf{Minerva} & \textbf{Olympiad} & \textbf{Avg.} \\
\midrule
GRPO        & $31.6$ & $12.8$ & $66.7$ & $79.0$ & $39.3$ & $40.1$ & $44.9$ \\
Entro. Adv. & $34.8$ & $13.4$ & $64.3$ & $77.6$ & $37.6$ & $39.9$ & $44.6$ \\
Entro. Loss & $32.7$ & $14.7$ & $71.3$ & $79.0$ & $36.8$ & $41.4$ & $46.0$ \\
Clip-Cov    & $30.4$ & $14.0$ & $72.3$ & $79.6$ & $37.1$ & $41.7$ & $45.8$ \\
\rowcolor{ourrow} \textbf{STEER (Ours)} & $\mathbf{36.1}$ & $\mathbf{16.0}$ & $\mathbf{76.3}$ & $\mathbf{80.5}$ & $\mathbf{39.5}$ & $\mathbf{42.3}$  & $\mathbf{48.5}$ \\
\bottomrule
\end{tabular}
\vspace{-0.8em}
\caption{Performance on test datasets in extreme scenarios.}
\label{Performance on test datasets in extreme scenarios.}
\vspace{-1.5em}
}
\end{table*}

\subsection{Empirical Results for Entropy Modulation}
\begin{figure}[t]
    \centering
    \includegraphics[width=0.87\linewidth, height=4cm]{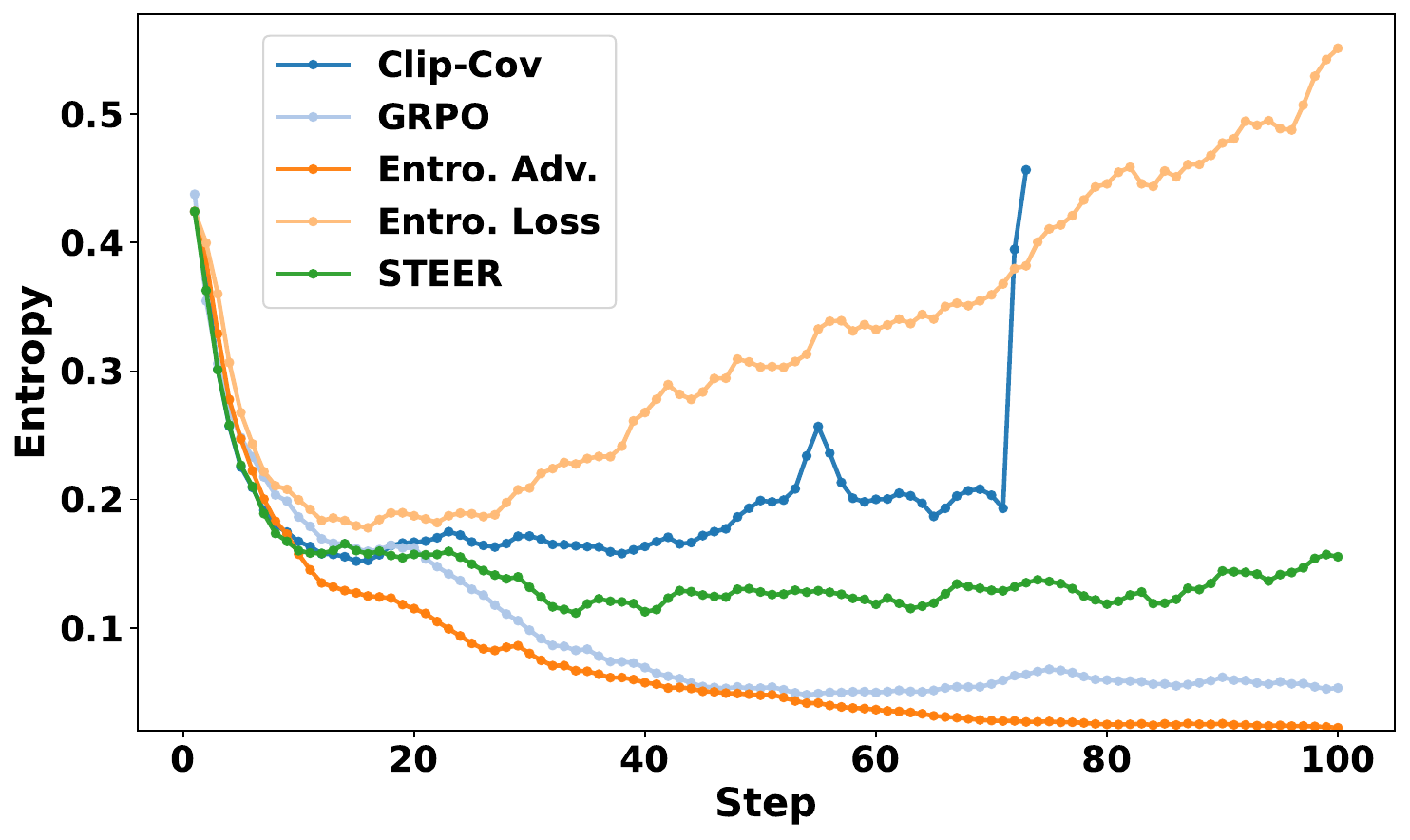}
\captionsetup{justification=centering,singlelinecheck=false}
\vspace{-0.7em}
    \caption{Entropy dynamics in extreme scenarios.}
    \label{entropy-high-5-low-0.99}
    \vspace{-0.8em}
\end{figure}

% The strength of STEER is not only reflected in its performance but also in its ability to regulate entropy across a wide range.
To better illustrate STEER's ability to regulate entropy across a wide range, 
we adopt an extreme training setup with $\varepsilon_{\text{high}} = 5$ and $\varepsilon_{\text{low}}=0.99$, where almost no ratio clipping is applied.
In such scenarios, RL training is vulnerable due to unstable gradient updates under extreme clipping ratios.
The results are shown in Figure~\ref{entropy-high-5-low-0.99}.
Most methods fail to maintain stable entropy: GRPO and Entro. Adv. tend toward entropy collapse; adding an Entropy Loss drives entropy up rapidly, leading to excessive uncertainty; and Clip-Cov cannot reliably control entropy. By contrast, STEER stabilizes after an initial decline and maintains steady entropy subsequently.
We test entropy intervention methods in this training setup and test set accuracy shown in Table~\ref{Performance on test datasets in extreme scenarios.}.
It can be observed that, even in training scenarios where the clipping operation is almost completely removed, STEER maintains relatively stable performance compared to other entropy intervention methods and achieves the highest accuracy across all test sets.

\begin{figure}[t]
% \vspace{-0.5em}
    \centering
\includegraphics[width=0.87\linewidth, height=4.4cm]{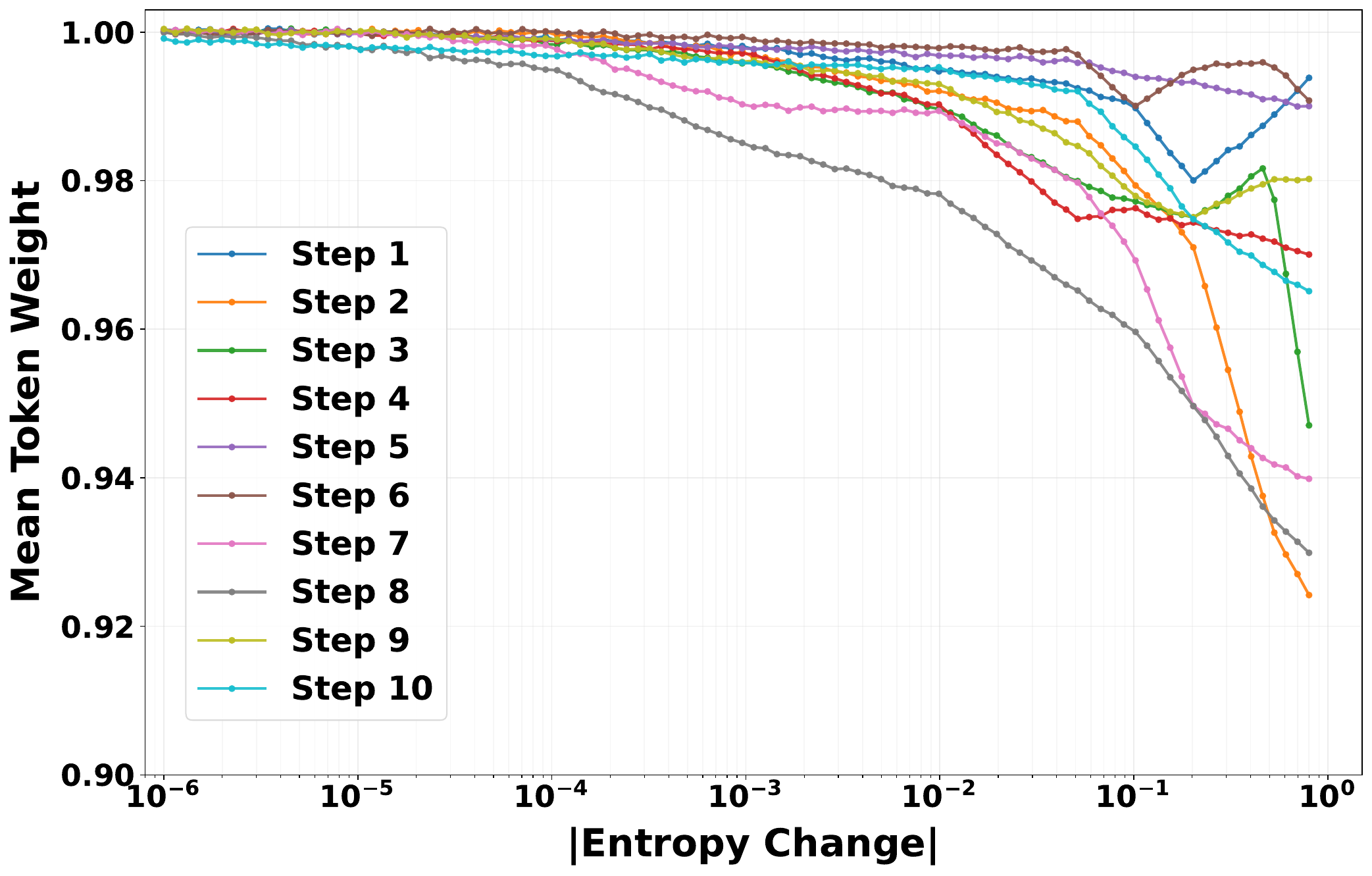}\captionsetup{justification=centering,singlelinecheck=false}
\vspace{-0.5em}
    \caption{Relationship between mean token weight and entropy change across steps.}
    \label{exp_token_weight}
    \vspace{-1.5em}
\end{figure}

Besides, Figure~\ref{exp_token_weight} depicts the trend of average token weight as a function of the absolute token entropy change in the first $10$ steps.
When entropy changes are small, most weights remain near $1$; only tokens with large entropy changes receive substantially reduced weights, indicating that STEER stabilizes training without impeding learning.

% \begin{figure}[htbp]
%     \centering
% \includegraphics[width=0.85\linewidth, height=4.2cm]{figs/exp_token_weight.pdf}\captionsetup{justification=centering,singlelinecheck=false}
%     \caption{Relationship between mean token weight and entropy change across steps.}
%     \label{exp_token_weight}
% \end{figure}

\vspace{-0.2em}
\subsection{Generalization Study}
We conduct experiments on additional backbones, including Llama-3.2-3B-Instruct~\citep{grattafiori2024llama} for math reasoning and Mistral-7B-v0.3~\citep{jiang2023mistral7b} for code editing. For each task, we use the same STEER hyperparameter as in the main experiments, rather than re-tuned per backbone. The results are reported in Appendix~\ref{Empirical Results Extended to Other Base Models and Other RL Algorithms}, and STEER yields consistent performance improvements.
We also evaluate STEER's performance with varying model size including 1.5B/7B/14B, as reported in Table~\ref{main_results}, Table~\ref{14B_results} and Appendix~\ref{Empirical Results Extended to Other Base Models and Other RL Algorithms}. We observe STEER yields superior performance across multiple model sizes.

Besides, we evaluate STEER under two other representative RL algorithms, RLOO~\citep{ahmadian2024back} and OPO~\citep{hao2025policy} with the default setting in verl implementations. The results are also shown in Appendix~\ref{Empirical Results Extended to Other Base Models and Other RL Algorithms}.
STEER improves the average score from $45.8 \to 46.8$ on RLOO and $46.4 \to 47.5$ on OPO, indicating that STEER can be effectively applied to other RLVR algorithms rather than specific to GRPO.

\vspace{-0.3em}
\section{Conclusion}
\vspace{-0.2em}
In this work, we conduct comprehensive theoretical and empirical analyses of entropy dynamics in RLVR. We derive a tight approximation for token-level entropy change at each update step, revealing four governing factors and providing a unified view to demystify the mechanisms of existing strategies.
We further introduce STEER, a principled entropy-modulation strategy that adaptively reweights tokens based on the theoretically-estimated entropy variations.
% Our extensive experiments demonstrate that STEER effectively mitigates entropy collapse and boosts reasoning performance across diverse benchmarks, LLMs and RL algorithms. 

\vspace{-0.2em}
\section{Limitations}
\vspace{-0.2em}
This study focuses on settings with verifiable rewards, while RL training without verifiable rewards is not explored.
Extending our analysis to more general scenarios, such as environments with limited verifiability, would be an interesting direction for future work.
It would be also valuable to investigate whether our findings generalize to other reinforcement learning settings, such as agentic and long-horizon RL.

\vspace{-0.2em}
\section{Acknowledgements}
\vspace{-0.3em}
This work is supported by the Starry Night Science Fund of Zhejiang University Shanghai Institute for Advanced Study (SN-ZJU-SIAS-001), and the National Natural Science Foundation of China (62476244,62372399).
We also sincerely thank Yi Liu and the CodeBuddy Team in Tencent (https://www.codebuddy.ai/) for their valuable assistance throughout this work.

% It would be highly interesting to extend our analyses to more general scenarios such as environments with limited verifiability.
% Besides, it would be also meaningful to extend our analyses to other RL scenarios, such as agentic and long-horizon RL.
% agentic and long-horizon RL
% This study mainly focuses on verifiable rewards, while RL training without verifiable rewards are not explored. It would be highly interesting to extend our analyses to more general scenarios such as environments with limited verifiability or paradigms with process-based rewards.

% This study mainly focuses on verifiable rewards, while RL training without verifiable rewards are not explored. It would be highly interesting to extend our analyses to more general cases with .

% It would be highly interesting to extend our analyses to other RL scenarios
% agentic and long-horizon RL

\clearpage
\clearpage

% \section{Acknowledgements}
% \vspace{-0.3em}
% This work is supported by the Starry Night Science Fund of Zhejiang University Shanghai Institute for Advanced Study (SN-ZJU-SIAS-001), and the National Natural Science Foundation of China (62476244,62372399).
% We also sincerely thank Tencent CodeBuddy (https://www.codebuddy.ai/), especially Yi Liu and the CodeBuddy Team, for their generous support and valuable assistance throughout this work.

% \newpage

% \section*{Ethics statement}

% We have manually reevaluated the dataset we created to ensure it is free of any potential for discrimination, human rights violations, bias, exploitation, and any other ethical concerns.

% \section*{Reproducibility statement}

% To ensure the reproducibility of our findings, all source code and datasets used in our experiments are included in the supplementary material. The provided materials are sufficient to replicate the main results presented in this paper.

\bibliography{custom}
% \bibliography{iclr2026_conference}
% \bibliographystyle{iclr2026_conference}

\clearpage
% \onecolumn
\appendix
\setcounter{theorem}{0}

\appendixtableofcontents

\clearpage

\appsection{Usage of LLMs}
Throughout the preparation of this manuscript, Large Language Models (LLMs) were utilized as a writing and editing tool. Specifically, we employed LLMs to improve the clarity and readability of the text, refine sentence structures, and correct grammatical errors. All final content, including the core scientific claims, experimental design, and conclusions, was conceived and written by us, and we take full responsibility for the final version of this paper.

\appsection{Related Work}\label{Related work}

\appsubsection{Reinforcement Learning with Verifiable Rewards in LLMs}

Reinforcement Learning with Verifiable Rewards (RLVR)~\citep{chang2024survey}, in which rewards are derived from a rule-based verifier, has recently emerged as an effective paradigm for enhancing the reasoning performance of large language models~\citep{zhang2025survey, jaech2024openai, lambert2024tulu,gao2023alleviating,wang2025polo, liu2025spark} beyond SFT~\citep{chang2025lora, chang2026balora}. In most RLVR settings, the reward is determined by checking whether the model output matches a reference answer, typically yielding a binary signal that reflects success or failure.
The development of core algorithms, notably PPO and GRPO~\citep{shao2024deepseekmath}, has substantially advanced RLVR. Several follow-up works further refine the optimization procedure, including reward-scheduling~\citep{wang2025scheduling}, improved advantage estimation~\citep{wu2026step, xie2025unlocking}, and extensions to multi-turn settings~\citep{ma2026tspo}.
Another line of work applies RLVR at scale or to diverse tasks, such as DeepSeek-R1~\citep{guo2025deepseek}, tool-augmented reasoning~\citep{zhang2025looptool}, and sample-efficient exploration~\citep{zhang2026expseek, zhao2026reinforced}.
Complementary approaches integrate curriculum learning~\citep{li2025curriculum}, adaptive temperature control~\citep{zhou2026look}, and benchmark-driven evaluation~\citep{an2025amo, wu2025table} into the RLVR pipeline. Recent studies also explore the interplay between logical structure and RL dynamics~\citep{zhang2026semanticawarelogicalreasoningsemiotic, zhang2026logicalphasetransitionsunderstanding, chen2023bias, lin2025recommendation, cui2025hatllm, wang2025survey, gao2023cirs, gao2023alleviating, bai2026ttvs}.

\appsubsection{Entropy-Oriented RL Methods for LLM Reasoning}

Entropy regularization~\citep{mnih2016asynchronous, haarnoja2018soft,chen2023bias,lin2025recommendation,gao2023cirs}, an early line of work in traditional RL, may mislead actions at critical states~\citep{zhang2025maximum} and has been shown to be highly sensitive to the coefficient in LLM training~\citep{cheng2025reasoning, cui2025entropy,cui2025hatllm,yang2024psl,yang2025breaking,zhang2026talos,yang2026bear}.
\citep{liu2025prorl} argues that the KL penalty preserves entropy and acts as a regularizer, ensuring that the online policy remains close to a stable reference, which stabilizes learning and reduces overfitting to misleading reward signals.
Nevertheless, the KL divergence term between the current policy $\pi_{\theta}$ and the reference policy $\pi_{\text{ref}}$ in the original form~\citep{shao2024deepseekmath} is excluded in our work, since its practical impact is often negligible or counterproductive for reasoning tasks, as demonstrated in recent works~\citep{yu2025dapo, chu2025gpg, hu2025open}.
One typical approach to address entropy collapse is by raising the sampling temperature during inference.
However, recent findings in~\citep{luo2025deepscaler}  suggest that while this method postpones the onset of entropy collapse, it does not prevent it, as entropy continues to decrease progressively throughout the training process.
Recent studies have sought to mitigate entropy collapse by adjusting key elements of policy optimization, such as PPO-style ratio clipping~\citep{yu2025dapo, yang2025dcpo}, balancing positive and negative samples~\citep{zhu2025surprising}, and applying KL regularization~\citep{liu2025prorl}. However, these methods are broad and lack fine-grained control at the token level, with their mechanisms often not fully explained in a unified or principled way.
Several methods attempt to encourage exploration via an entropy-induced advantage~\citep{cheng2025reasoning, tan2025gtpo, wang2025beyond, wang2025stabilizing, deng2025decomposing}, with the intuition that emphasizing uncertain states will promote exploration and raise overall policy entropy.
In practice, however, we found this design often fails to reliably mitigate entropy collapse because it disproportionately strengthens learning on high-entropy tokens and thereby magnifies entropy change, leading to unreliable entropy control.
Although prior work~\citep{cui2025entropy} considers entropy change, the resulting estimation is distorted due to its unreasonable state-equivalence assumption.
Notably, its entropy-control scheme (i) enforces a hard binary split by entropy change without considering their intra-group differentiation, and (ii) may hinder the learning process, since high-entropy-change tokens that are informative for exploration are over-penalized.

\begin{table*}[h]
\centering
\begin{tabular}{p{0.38\linewidth} l} % 左列定宽，按需要调 0.30~0.38
\toprule
\textbf{Methods} & \textbf{Intervention Strategies} \\
\midrule
\shortstack[l]{\textbf{DAPO / DCPO}\\ ~\citep{yu2025dapo}\\
~\citep{yang2025dcpo}}
  & $\mathbb{I}_{\text{clip}}=
    \begin{cases}
      0, & A_{i,t}>0 \text{ and } r_{i,t}>1+\varepsilon_{\text{high}},\\
      0, & A_{i,t}<0 \text{ and } r_{i,t}<1-\varepsilon_{\text{low}},\\
      1, & \text{otherwise}
    \end{cases}$ \\
\midrule
\shortstack[l]{\textbf{KL penalty}\\~\citep{shao2024deepseekmath} }
  & $\mathcal{R} \left( \pi_\theta \right) = \frac{\pi_{\text{ref}}(o_{i,t}\mid q,o_{i,<t})}{\pi_\theta(o_{i,t}\mid q,o_{i,<t})}$ \\
\midrule
\shortstack[l]{\textbf{Entropy Regularization}\\~\citep{he2025skywork} }
   & $\mathcal{R} \left( \pi_\theta \right) = - \log \pi_\theta(o_{i,t}\mid q,o_{i,<t})$ \\
  \midrule
\shortstack[l]{\textbf{Unlikeliness}\\~\citep{he2025rewarding} }
   &  $\hat{R}_{i,t} = R_{i,t} \left(1- \beta_{\text{rank}} \frac{G-\text{rank}(o_i)}{G}\right)$, $\,\, \beta_{\text{rank}} > 0$ \\
  \midrule
\shortstack[l]{\textbf{W-REINFORCE}\\~\citep{zhu2025surprising} }
   &
  $\hat{A}_{i,t} =
  \begin{cases}
  \lambda, & A_{i,t} > 0 \\
  1, & A_{i,t} < 0
  \end{cases}, \quad  \lambda < 1$ \\
  \midrule
\shortstack[l]{\textbf{Entropy Advantage}\\~\citep{cheng2025reasoning} }
   & 
  $\hat{A}_{i,t} = A_{i,t} + \min \left( \alpha \cdot \mathcal{H}_{i,t}^{\text{detach}}, \frac{|A_{i,t}|}{\kappa} \right),$ \;\; $\alpha > 0, \kappa > 1$ \\
  \midrule
  \shortstack[l]{\textbf{GTPO}\\ ~\citep{tan2025gtpo} }
   & $ \hat{R}_{i,t} = R_{i,t} + \alpha \dfrac{\mathcal{H}_{i,t}}{\frac{1}{d_t}\sum_{k=1}^{d_t} \mathcal{H}_{k,t}}  , \text{ for } \, R_{i,t}>0$ \\
     \midrule
       \shortstack[l]{\textbf{EDGE-GRPO}\\~\citep{zhang2025edge} }
   & $ \hat{A}_{i} = \frac{A_{i}}{\hat{\mathcal{H}}_{i}}, \,\, \hat{\mathcal{H}}_{i}: \text{normalized entropy across the group}$ \\
     \midrule
\shortstack[l]{\textbf{PPL-based}\\~\citep{deng2025decomposing} }
   & $ \hat{A}_{i,t} = A_{i,t} (1-\alpha \text{log-PPL} (o_i))$,$\,\, \alpha > 0$ \\
  \midrule
\shortstack[l]{\textbf{Position-based}\\~\citep{deng2025decomposing} }
   & $ \hat{A}_{i,t} = A_{i,t} + \gamma \text{sign}(A_{i,t}) \sigma (r_{it})$ \;\; $r_{it}$: token’s relative position \\
  \midrule
\shortstack[l]{\textbf{Forking Tokens}\\~\citep{wang2025beyond} }
   &  $\mathbb{I}_{\text{clip}} = \mathbb{I}_{\text{clip}} \land \mathbb{I}( \mathcal{H}_{i,t}  > \tau_{\mathcal{B}} ), \,\, \tau_{\mathcal{B}} \text{: threshold in batch } \mathcal{B}$ \\
%   \midrule
% \shortstack[l]{\textbf{Clip-Cov}\\~\citep{cui2025entropy} }
%    &  $\mathbb{I}_{\text{clip}} = \mathbb{I}_{\text{clip}}\land \mathbb{I}\!\left( \bigl(\log \pi_\theta(o_{i,t}) - \tfrac{1}{N}\sum_{j=1}^N \log \pi_\theta(y_j)\bigr)
%   \bigl(A(o_{i,t}) - \tfrac{1}{N}\sum_{j=1}^N A(y_j)\bigr)  > \tau_{\mathcal{D}} \right)$ \\
%     \midrule
% \shortstack[l]{\textbf{KL-Cov}\\~\citep{cui2025entropy} }
%    &  $\mathcal{R} \left( \pi_\theta \right) = \frac{\pi_{\text{old}}(o_{i,t}\mid q,o_{i,<t})}{\pi_\theta(o_{i,t}\mid q,o_{i,<t})}-1$ \\
\bottomrule
\end{tabular}
\caption{A Token-level Gradient Reweighting Perspective for Existing Entropy Shaping Methods.}
\label{A Token-level Gradient Reweighting Perspective for Shaping Policy Entropy}
\end{table*}

\appsection{Formulations of Entropy Shaping}\label{Appendix A Token-level Gradient Reweighting Perspective for Shaping Policy Entropy}

\appsubsection{A Unified Token-level Gradient Reweighting Framework}
We show that existing entropy-intervention methods, 
despite their diverse designs, all operate by modifying a 
single quantity: the \emph{token-level gradient weight}. 
This unified view directly connects to the four governing 
factors identified in Theorem~\ref{theorem_entropy_change}.

The policy gradient for a broad family of RL algorithms can be expressed as follows:
\begin{equation}\label{app_gradient}
\begin{aligned}
    \nabla_\theta J(\theta)
= &\mathbb{E}_{q \sim \mathcal D,\ \{o_i\}\sim\pi_{\text{old}}(\cdot\mid q)} \Bigg[ \frac{1}{\sum_{i=1}^G|o_i|} \sum_{i=1}^G\sum_{t=1}^{|o_i|} \\  &w_{i, t}(q)
\nabla_\theta\log\pi_\theta(o_{i,t}\mid q,o_{i,<t}) \Bigg].
\end{aligned}
\end{equation}
where $w_{i, t}(q) =  \mathbb{I}_{\text{clip}} r_{i,t}  A_{i,t} + \beta \mathcal{R} \left( \pi_\theta \right)$.
Specifically, $r_{i,t} = \frac{\pi_\theta(o_{i,t}|q,o_{i,<t})}{\pi_{\text{old}}(o_{i,t}|q,o_{i,<t})}$ denotes the importance sampling ratio, advantage $A_{i,t}$ is calculated by reward $R_{i,t}$ and 
\begin{equation}
\mathbb{I}_{\text{clip}}  =
\begin{cases}
0, & A_{i,t} > 0 \;\;\text{and}\;\; r_{i,t} > 1+\varepsilon, \\
0, & A_{i,t} < 0 \;\;\text{and}\;\; r_{i,t} < 1-\varepsilon, \\
1, & \text{otherwise}.
\end{cases}
\end{equation}
$\mathcal{R} \left( \pi_\theta \right)$ is an optional regularization, where some algorithms set $\beta=0$, such as GRPO in our main text.
Table~\ref{A Token-level Gradient Reweighting Perspective for Shaping Policy Entropy} maps each existing entropy-intervention method 
to its specific modification of $w_{i,t}$.
These methods adjust only a subset of the four 
governing factors, which explains both their partial 
effectiveness and their inherent limitations.

\appsubsection{Theoretical Analyses Extended to Other RL Algorithms}
\label{Theoretical analyses extended to other RL algorithms}
While our main theoretical analyses are presented under the GRPO update, they are not limited to GRPO and can be extended to other policy-gradient RL algorithms. In particular, our derivation is based on the policy-gradient formulation; and the GRPO policy-gradient expression can be adapted to other methods by making minor modifications based on Eq.~\eqref{app_gradient}.
Specifically, different RL algorithms can be recovered by:
\begin{itemize}
    \item GRPO~\citep{shao2024deepseekmath}: $A_{i,t} = A_{\text{group}}$.
    \item PPO~\citep{schulman2017proximal}: PPO shares the same clipped-ratio structure as GRPO; the difference is the advantage estimator $A_{i,t} = A_{\text{GAE}}$.
    \item RLOO~\citep{ahmadian2024back}: It is REINFORCE-style with a leave-one-out baseline, i.e., $r_{i,t} = 1$ and
    \begin{equation}
    A_{i,t} = R_i - \frac{1}{G-1} \sum_{j \neq i} R_j
    \end{equation}
    \item OPO~\citep{hao2025policy}: It is on-policy RL with an optimal reward baseline, so typically $r_{i,t} = 1$ and $A_{i,t}$ is a baseline-corrected advantage.
\end{itemize}
Substituting the above generalized policy-gradient form into the proof of Theorem~\ref{theorem_entropy_change} yields a corresponding generalized estimator of the entropy change:
\begin{equation}
\begin{split}
\Omega(s) &= -\frac{\eta}{L} \mathbb{E}_{\pi_\theta(\cdot|s)} \Big[ \mathbb{I}_{\text{clip}} r_{i,t} A_{i,t} \\
& \quad \pi_\theta (1 - \pi_\theta) \left( \log \pi_\theta + \mathcal{H}(s) \right) \Big].
\end{split}
\end{equation}
This shows that a unified entropy-change estimator can be derived for multiple RL algorithms within the same analytical framework, with the specific choice of advantage function given by the definitions above. Consequently, the theoretical insights developed for GRPO are also applicable to other RL algorithms.

% 下面我们首先将现有熵相关工作都统一在同一框架下，并解释他们对熵的影响。然后我们再逐一分析熵相关的现象

\appsection{Supplementary Empirical Analysis for RLVR Entropy}
This section presents supplementary experiments that further support the analyses and claims in the main text.

\begin{figure*}[htbp]
    \centering
    \begin{subfigure}[t]{0.3\linewidth}
      \includegraphics[height=3.5cm]{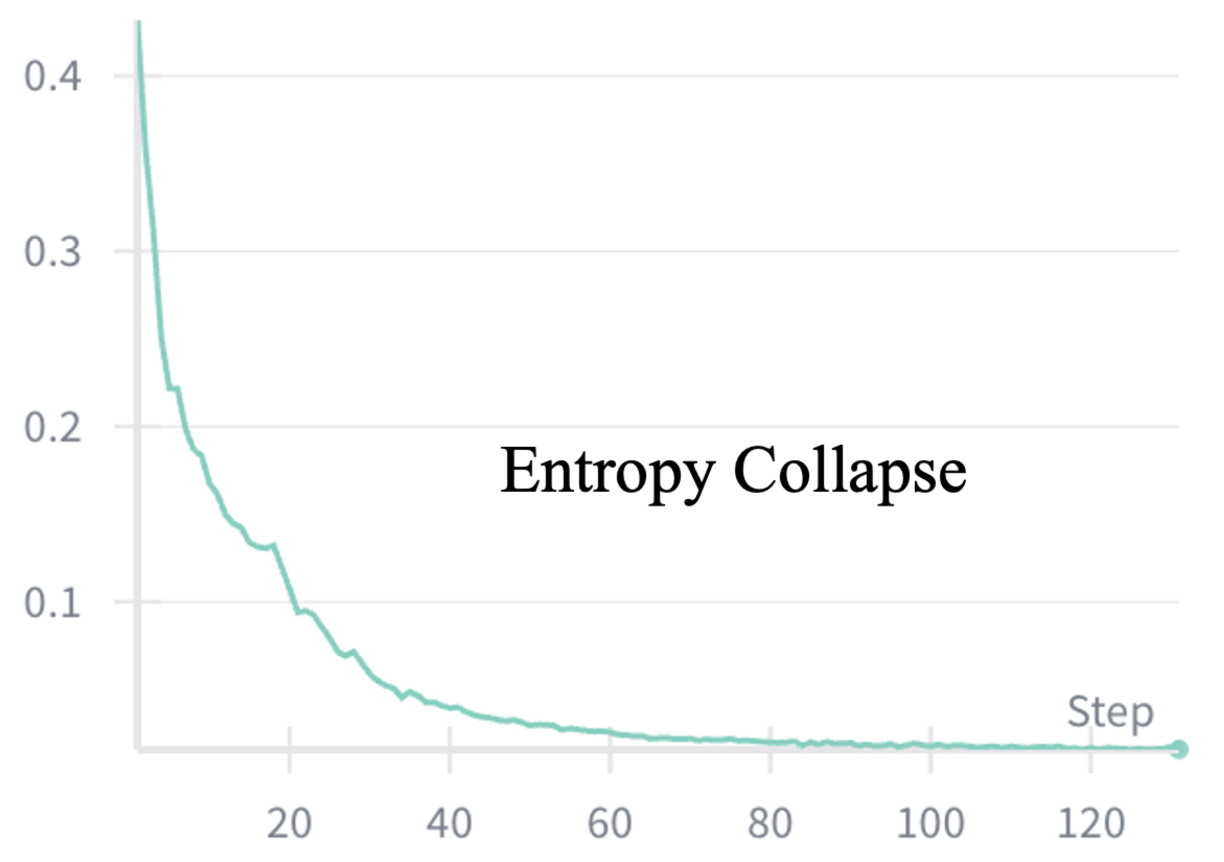}
      \caption{Continuity across batches}
    \end{subfigure}
    \begin{subfigure}[t]{0.3\linewidth}
      \includegraphics[height=3.5cm]{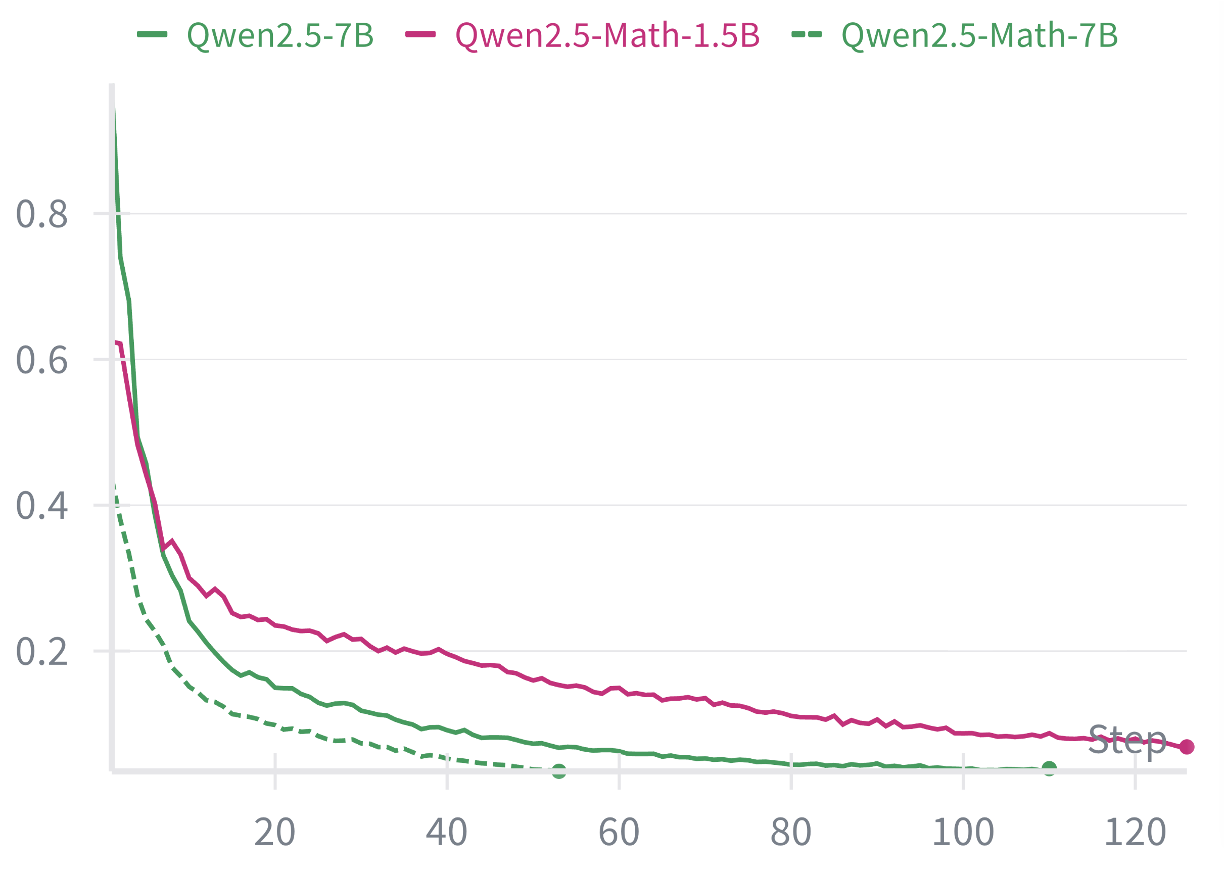}
      \caption{Dependence on model}
    \end{subfigure}
    \begin{subfigure}[t]{0.3\linewidth}
      \includegraphics[height=3.5cm]{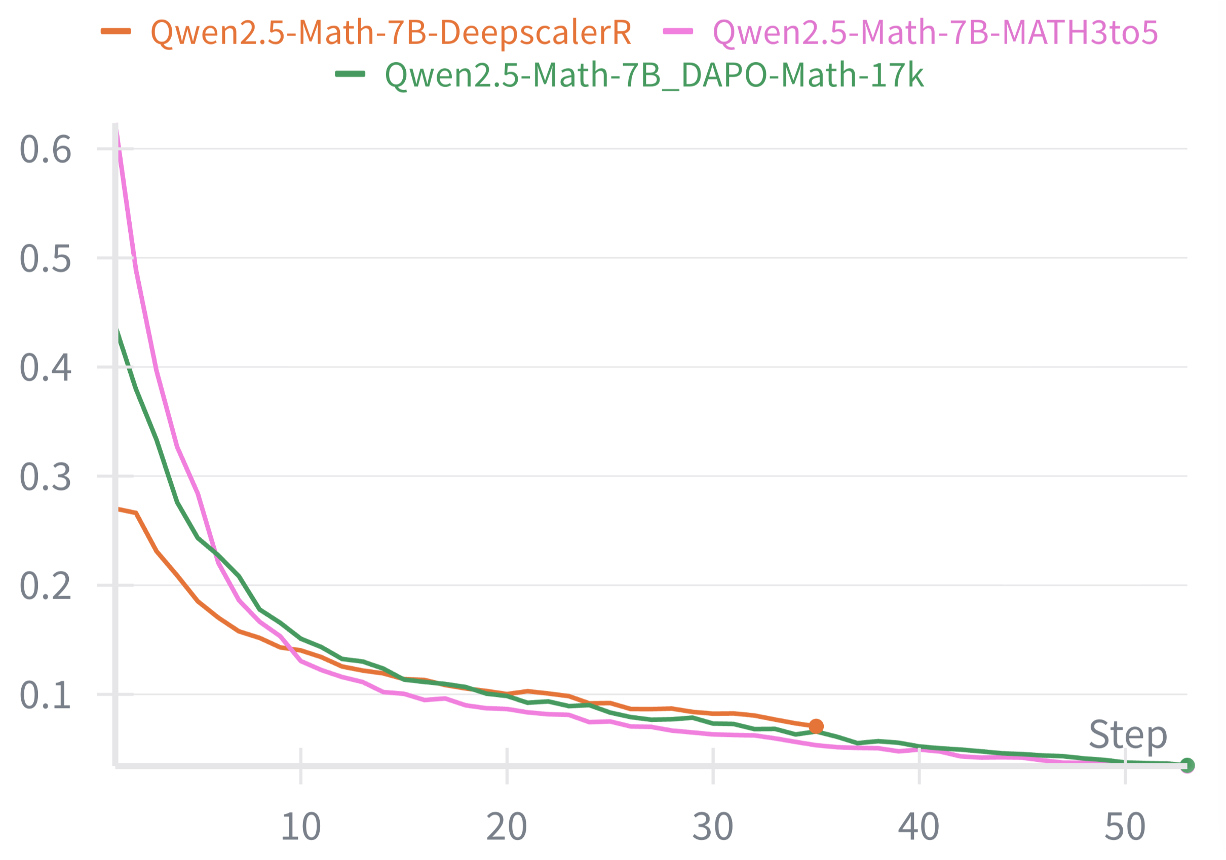}
      \caption{Dependence on dataset}
    \end{subfigure}
    \caption{Empirical properties of policy entropy in LLM RL training.}
    \label{Empirical properties of policy entropy.}
\end{figure*}

\begin{figure*}[htbp]
    \centering
    \begin{subfigure}[t]{0.24\linewidth}
      \includegraphics[height=2.8cm]{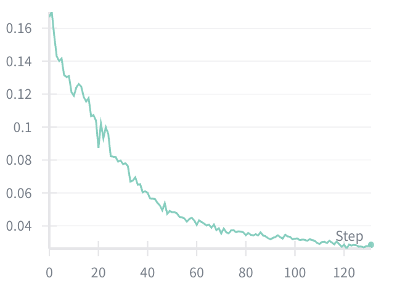}
      \caption{Minerva}
    \end{subfigure}
    \begin{subfigure}[t]{0.24\linewidth}
      \includegraphics[height=2.8cm]{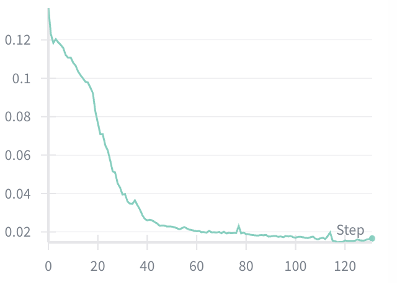}
      \caption{AMC23}
    \end{subfigure}
    \begin{subfigure}[t]{0.24\linewidth}
      \includegraphics[height=2.8cm]{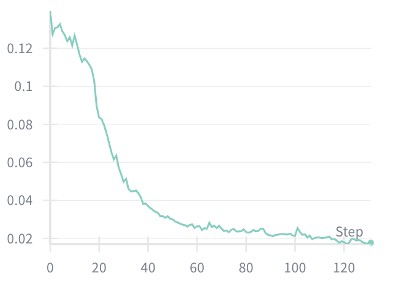}
      \caption{AIME24}
    \end{subfigure}
    \begin{subfigure}[t]{0.24\linewidth}
      \includegraphics[height=2.8cm]{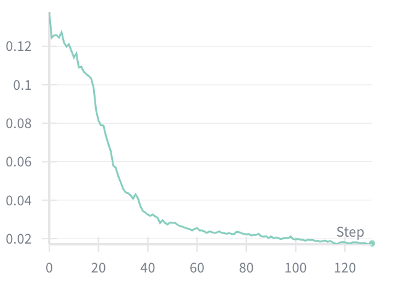}
      \caption{AIME25}
    \end{subfigure}
    \caption{In-domain consistency.}
    \label{In-domain consistency.}
\end{figure*}

\appsubsection{Empirical Properties of Policy Entropy}\label{Empirical Properties of Policy Entropy}
In this subsection, we present several basic empirical properties of policy entropy dynamics in RLVR, providing a global view of how entropy behaves during training.

Figure~\ref{Empirical properties of policy entropy.} summarizes several empirical properties of policy entropy in RLVR.
First, continuity across batch (panel 1) shows that, under a standard GRPO run, policy entropy decays smoothly step by step and entropy collapse appears as a gradual trend rather than sudden jumps.
Second, dependence on model (panel 2) compares different backbones (e.g., Qwen2.5-Math-1.5B / 7B vs. Qwen2.5-7B) and reveals that while all models eventually experience entropy collapse, the initial level and decay speed vary with model size and pretraining.
Third, dependence on dataset (panel 3) indicates that training on different math datasets (Math-3to5, DAPO-MATH, DeepScaleR) leads to distinct entropy curves, suggesting that entropy dynamics is also shaped by data distribution and difficulty.
Finally, in-domain consistency (Figure~\ref{In-domain consistency.}) shows that, within a fixed training run, entropy measured on several test sets in the same domain (Minerva, AMC23, AIME24, AIME25) follows highly similar monotonically decreasing trajectories, implying that a single global entropy trend governs a wide range of tasks.
Together, these observations provide an overall picture of how policy entropy behaves in RLVR before any additional intervention is applied.

\appsubsection{Entropy Change Estimation Comparison}\label{Entropy Change Estimation Comparison}

The comparison results on three metrics are shown in Table~\ref{MSE_and_PCC} in the main text strongly validate the effectiveness of our estimator derived in Theorem~\ref{theorem_entropy_change}.

\begin{figure*}[htbp]
    \centering
    \begin{subfigure}[t]{0.3\linewidth}
      \includegraphics[height=3.5cm, width=4.5cm]{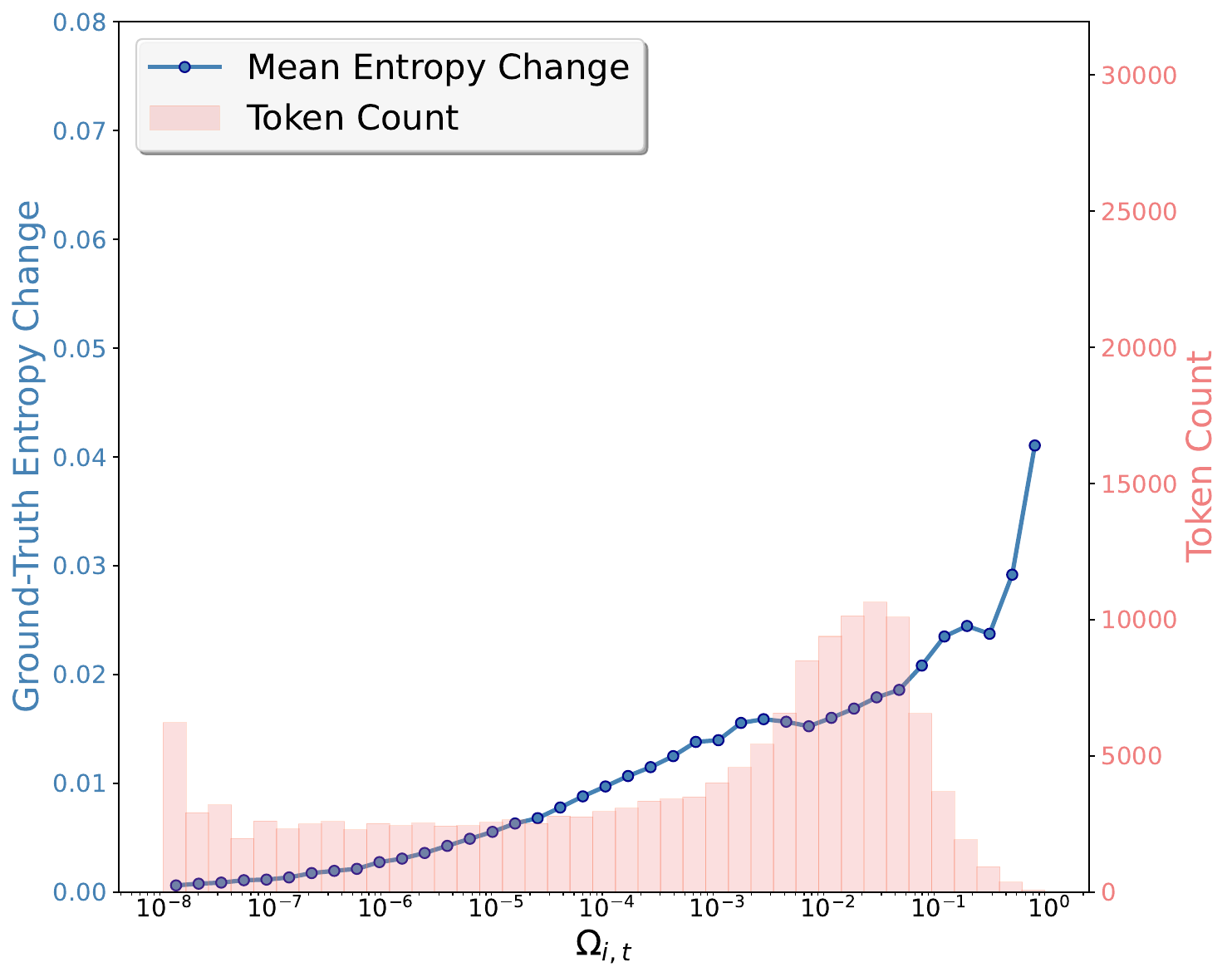}
      \caption{Ours on Math-1.5B}
    \end{subfigure}
    \begin{subfigure}[t]{0.3\linewidth}
      \includegraphics[height=3.5cm, width=4.5cm]{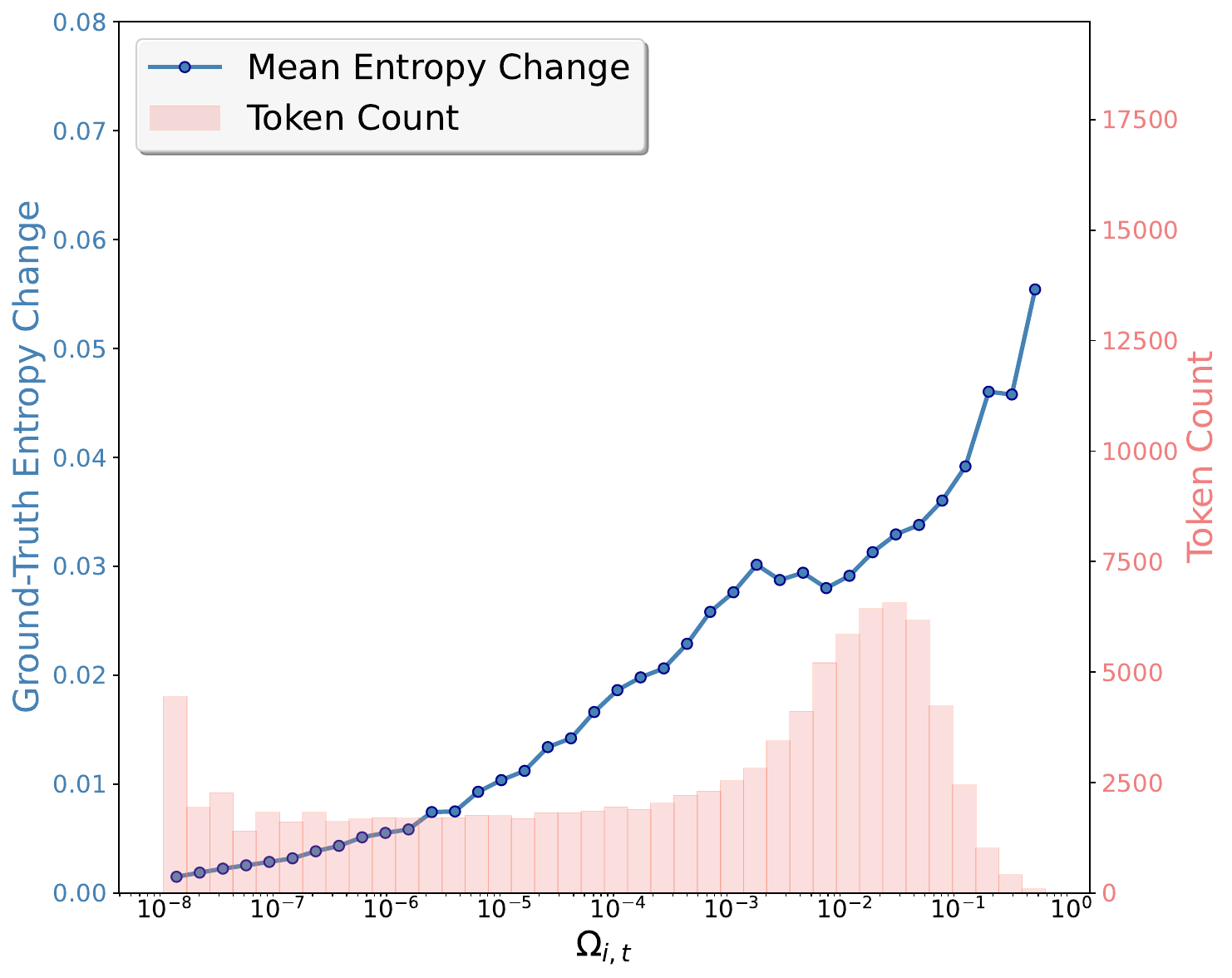}
      \caption{Ours on 7B}
    \end{subfigure}
    \begin{subfigure}[t]{0.3\linewidth}
      \includegraphics[height=3.5cm, width=4.5cm]{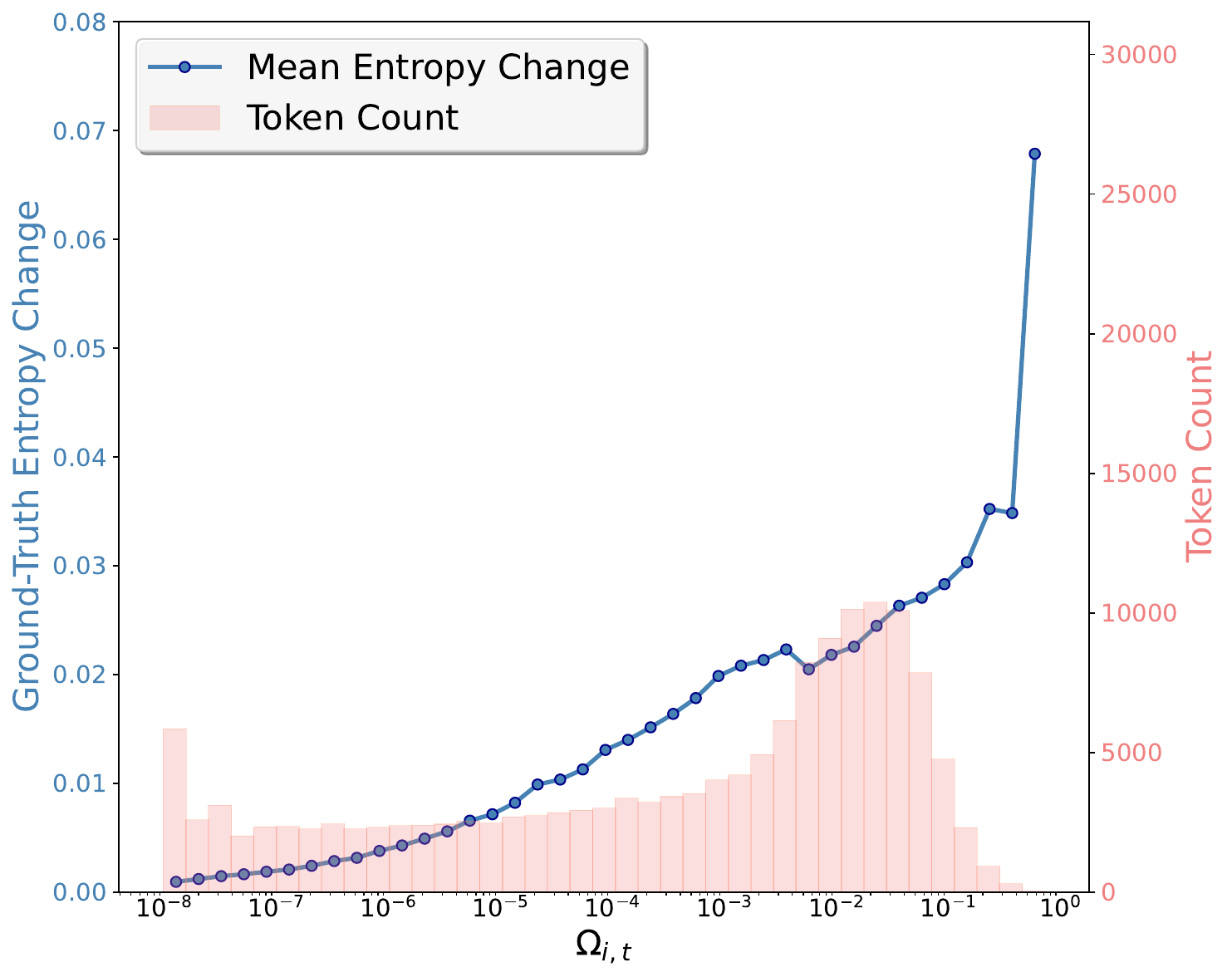}
      \caption{Ours on Math-7B}
    \end{subfigure}
    \begin{subfigure}[t]{0.3\linewidth}
      \includegraphics[height=3.5cm, width=4.5cm]{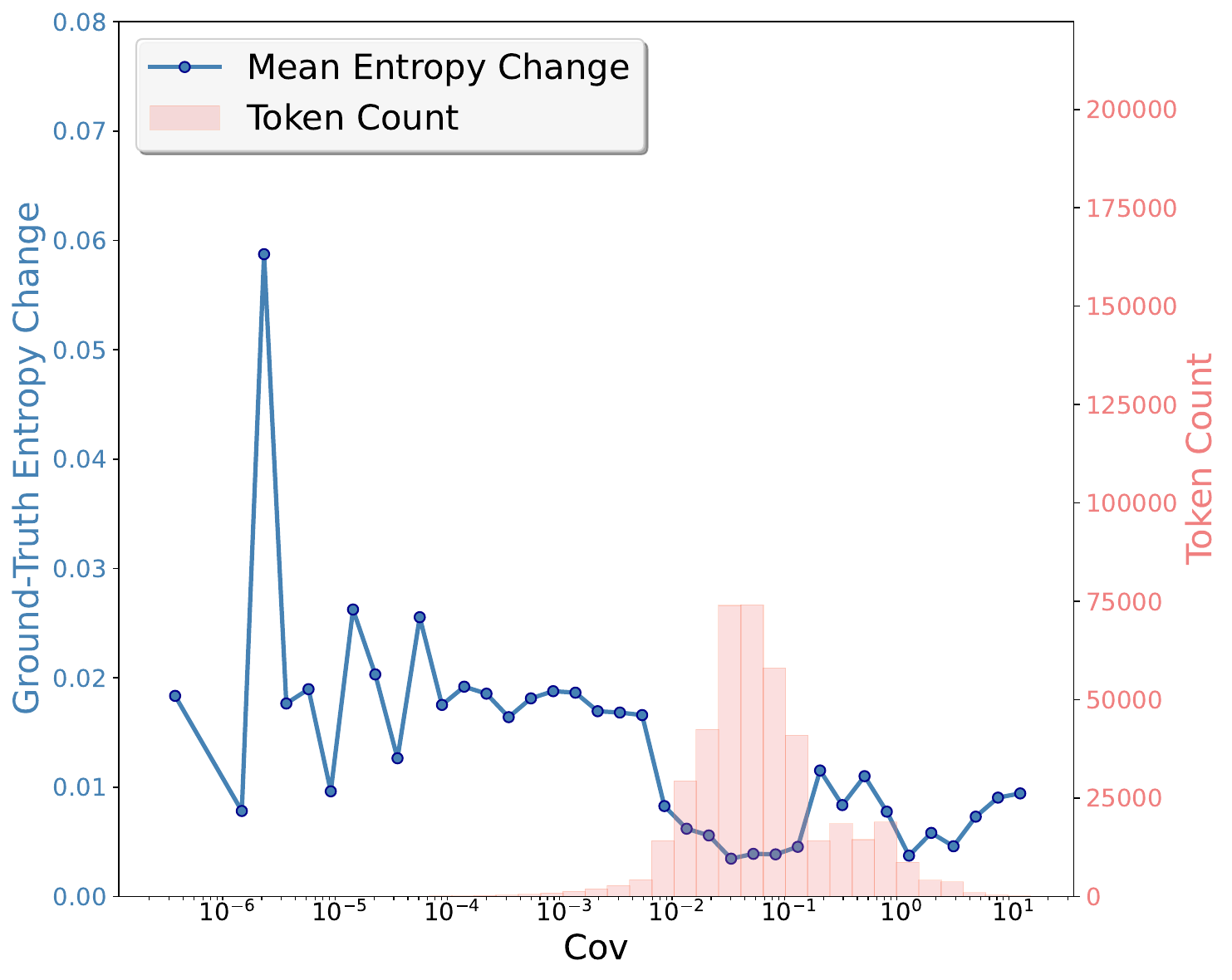}
      \caption{Cov on Math-7B}
    \end{subfigure}
    \begin{subfigure}[t]{0.3\linewidth}
      \includegraphics[height=3.5cm, width=4.5cm]{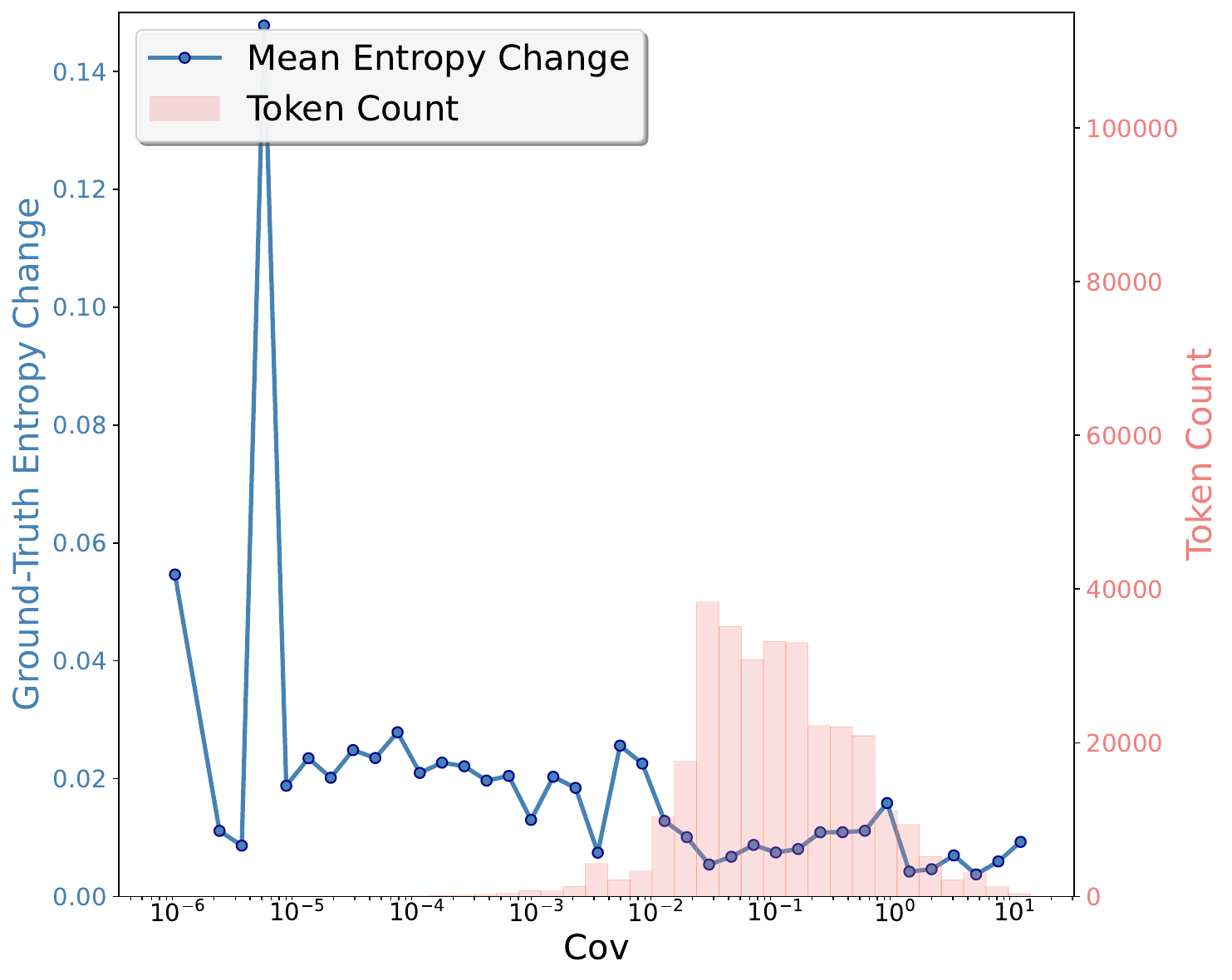}
      \caption{Cov on Math-7B}
    \end{subfigure}
    \begin{subfigure}[t]{0.3\linewidth}
      \includegraphics[height=3.5cm, width=4.5cm]{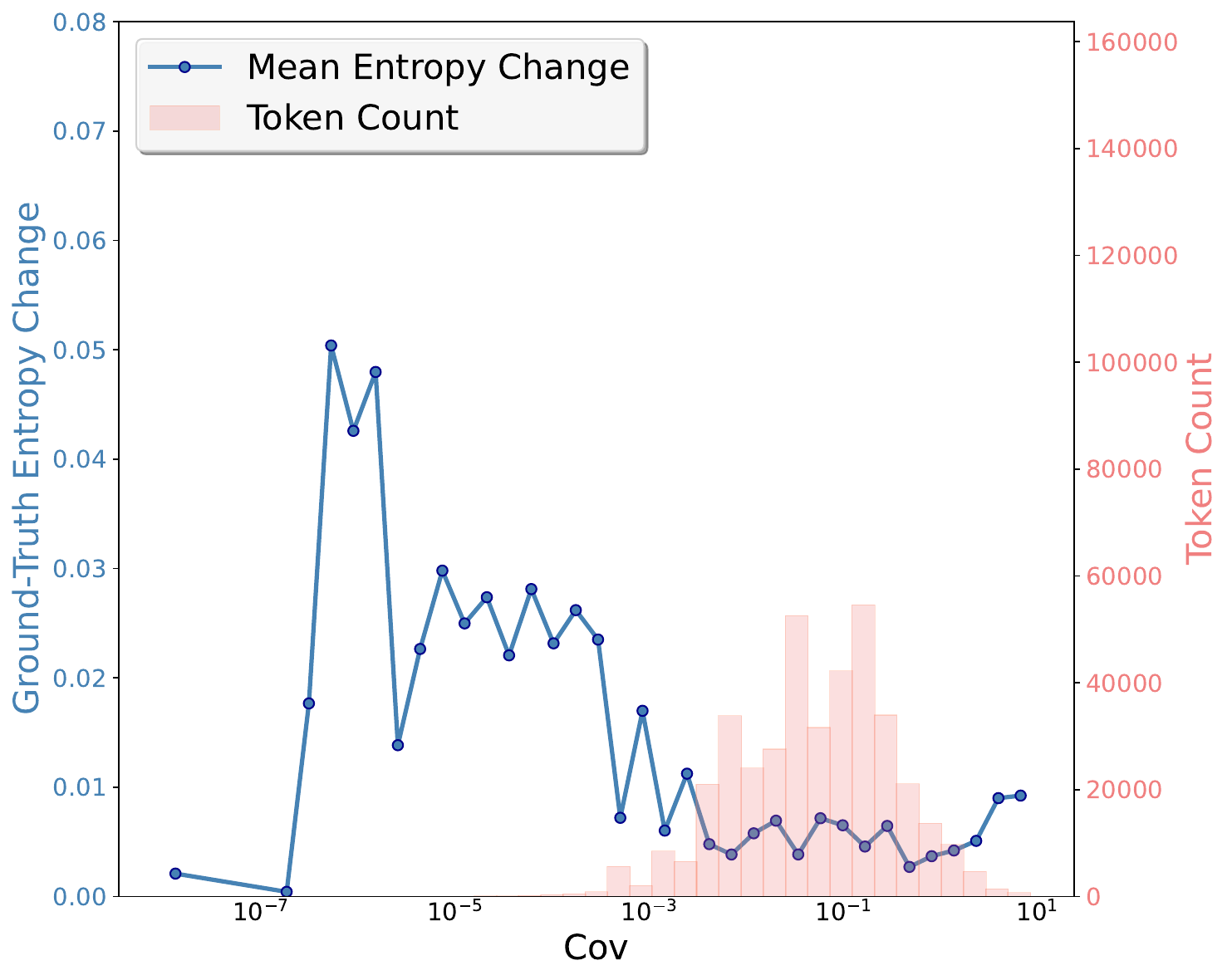}
      \caption{Cov on Math-7B}
    \end{subfigure}
    \caption{Entropy change comparison on dataset DAPO-Math-17k.}
    \label{Entropy Change on DAPO-Math-17k.}
\end{figure*}

\begin{figure*}[htbp]
    \centering
    % ---- 第一行 ----
    \begin{subfigure}[t]{0.3\linewidth}
      \includegraphics[height=3.5cm, width=4.5cm]{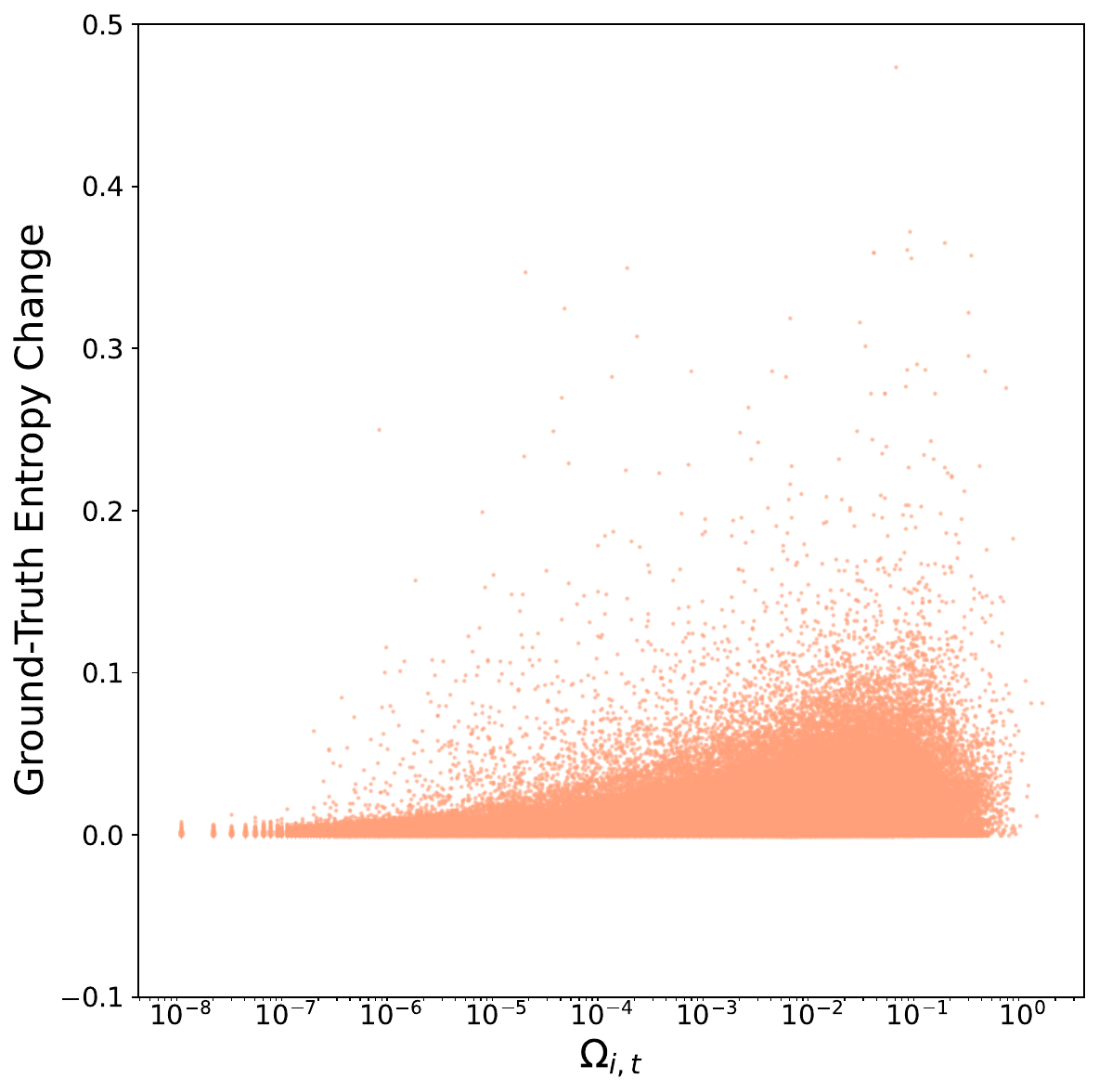}
      \caption{Ours on Math-1.5B}
    \end{subfigure}
    \begin{subfigure}[t]{0.3\linewidth}
      \includegraphics[height=3.5cm, width=4.5cm]{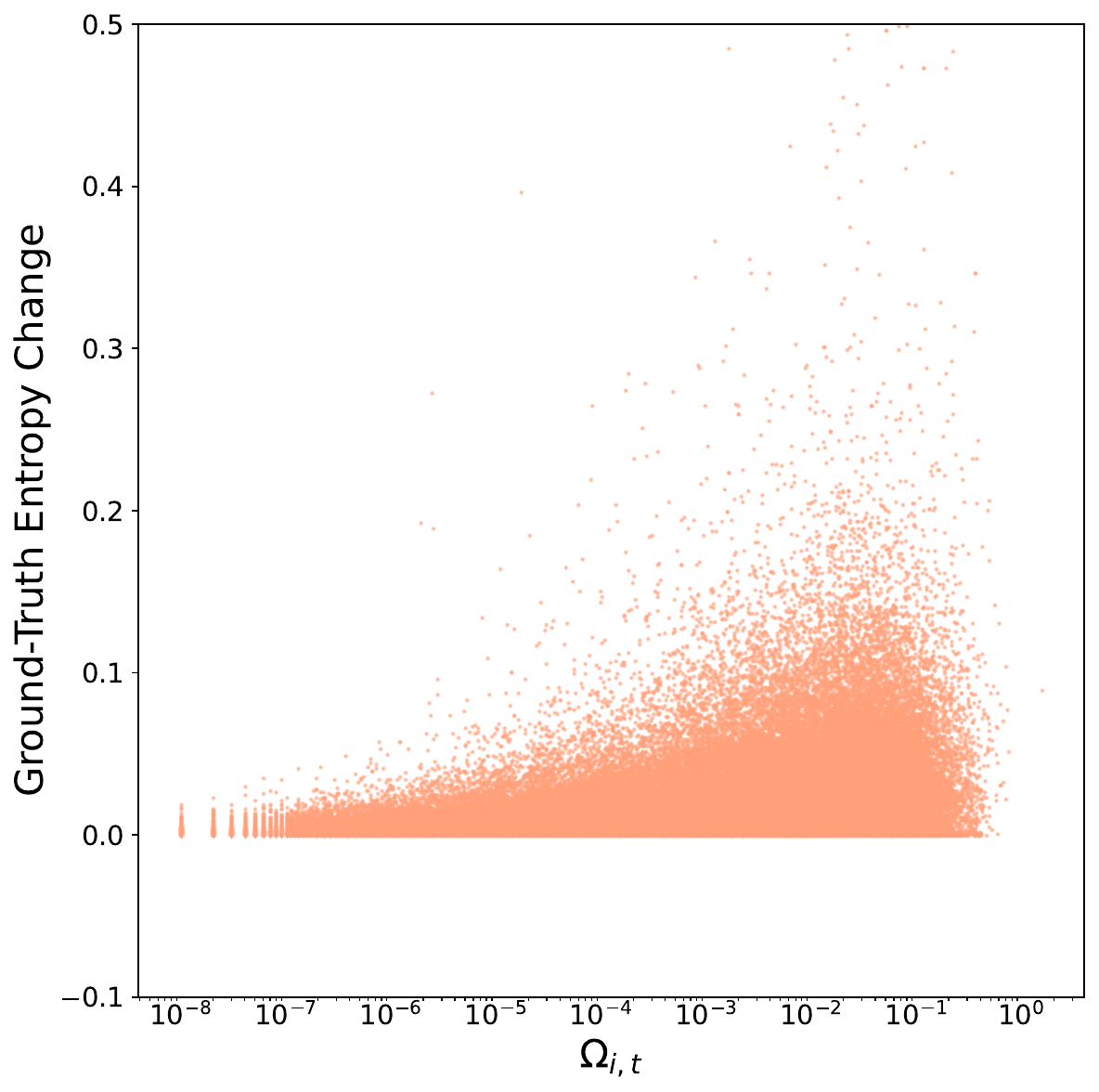}
      \caption{Ours on 7B}
    \end{subfigure}
    \begin{subfigure}[t]{0.3\linewidth}
      \includegraphics[height=3.5cm, width=4.5cm]{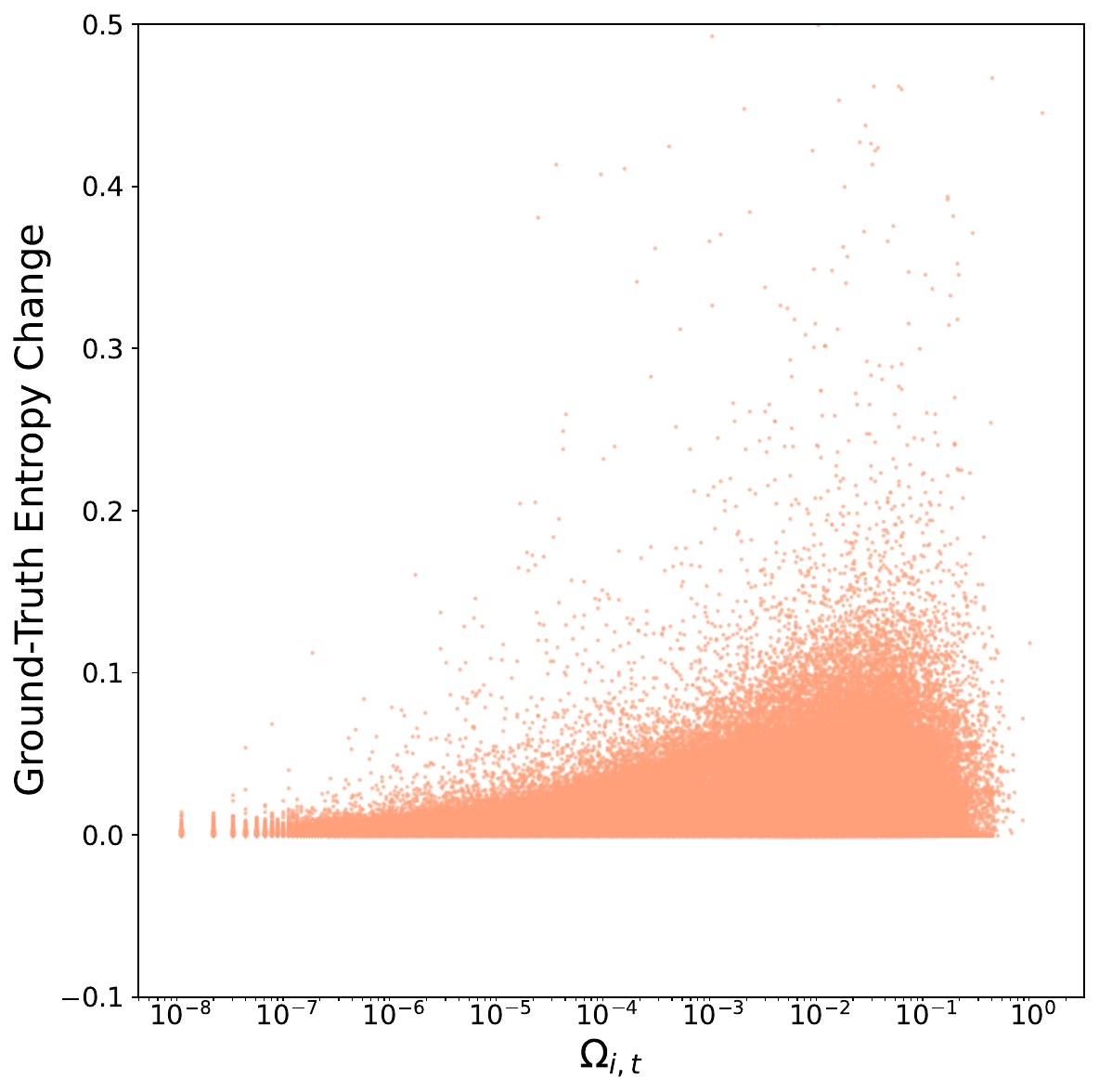}
      \caption{Ours on Math-7B}
    \end{subfigure}
    \begin{subfigure}[t]{0.3\linewidth}
      \includegraphics[height=3.5cm, width=4.5cm]{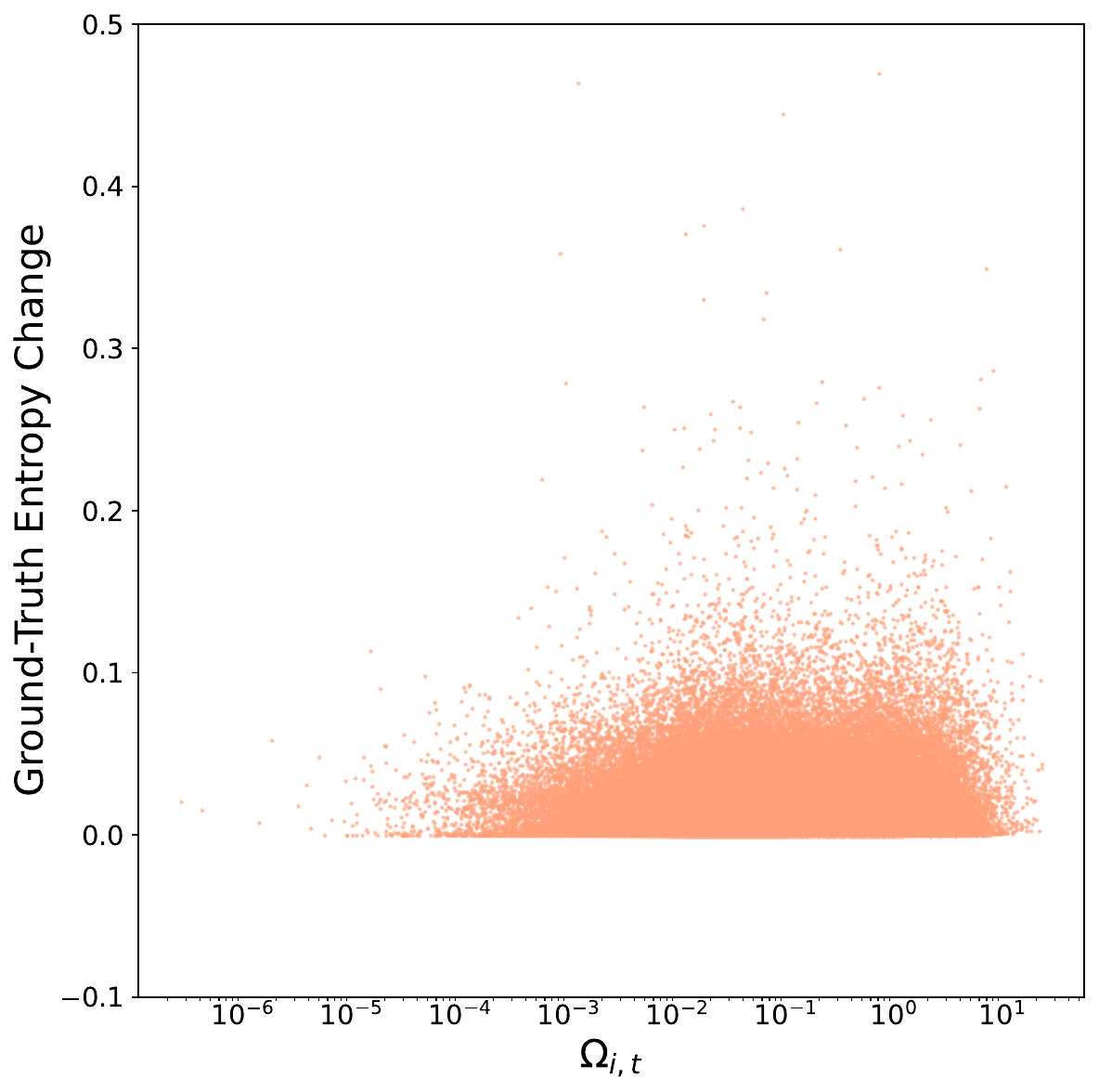}
      \caption{Cov on Math-1.5B}
    \end{subfigure}
    \begin{subfigure}[t]{0.3\linewidth}
      \includegraphics[height=3.5cm, width=4.5cm]{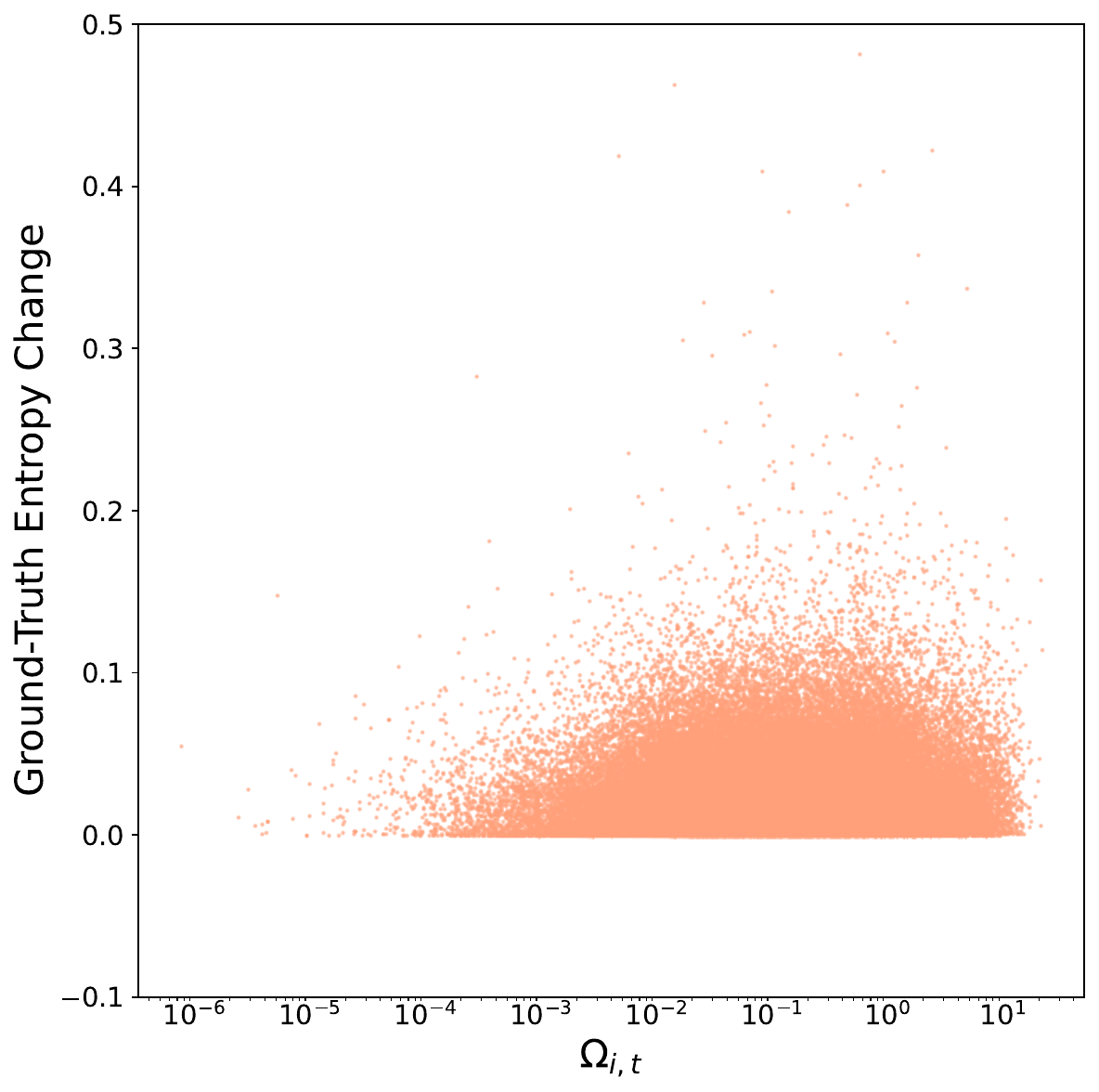}
      \caption{Cov on 7B}
    \end{subfigure}
    \begin{subfigure}[t]{0.3\linewidth}
      \includegraphics[height=3.5cm, width=4.5cm]{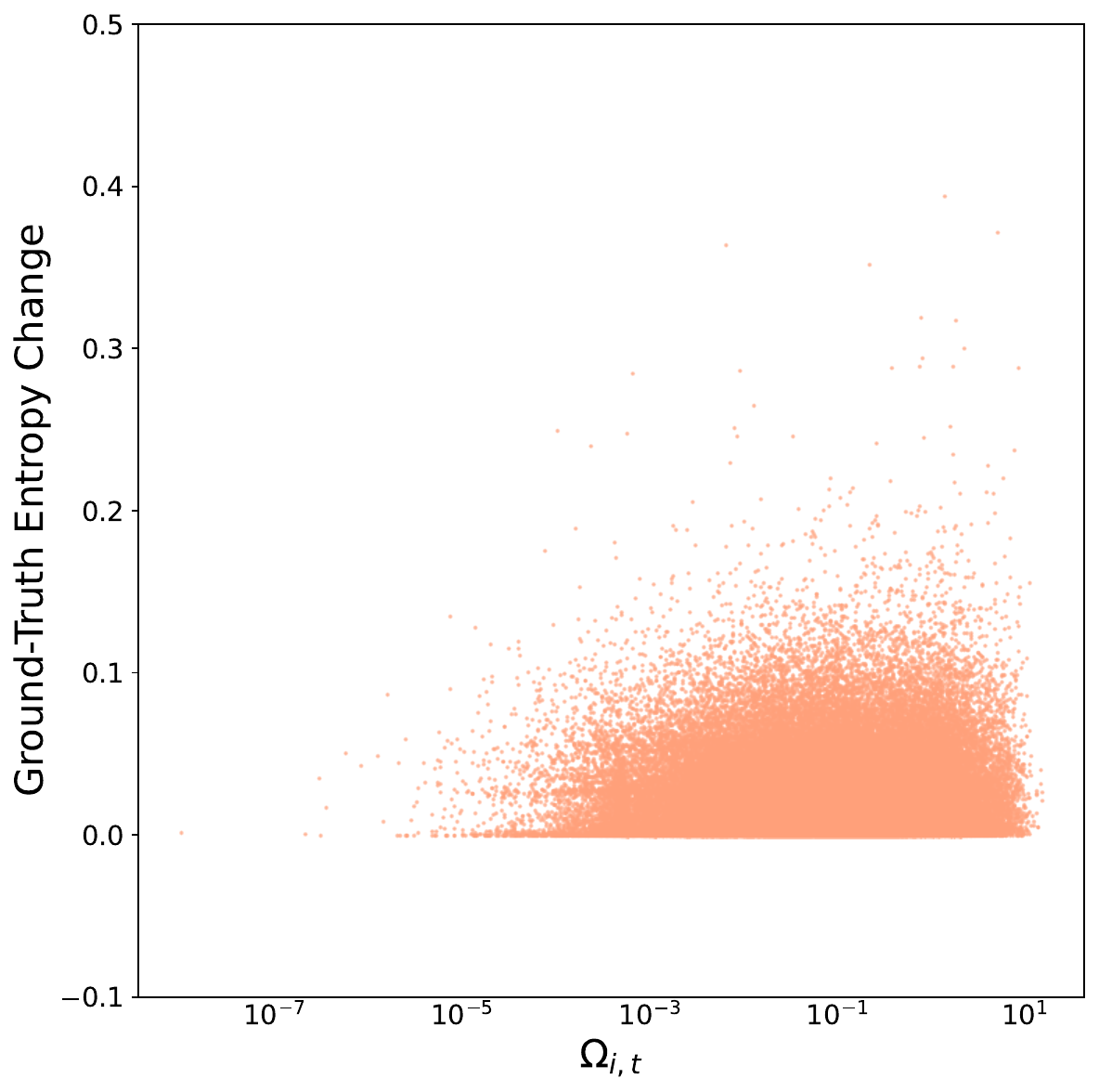}
      \caption{Cov on Math-7B}
    \end{subfigure}
    \caption{Entropy change scatters on dataset DAPO-Math-17k.}
    \label{Entropy Change scatters on DAPO-Math-17k.}
\end{figure*}

Beyond these, we record the token entropy changes for the first $10$ training steps across different models and datasets.
Figure~\ref{Entropy Change on DAPO-Math-17k.} and~\ref{Entropy Change scatters on DAPO-Math-17k.} show the results on dataset DAPO-Math-17k.
The curve denotes the estimated vs. ground-truth entropy change
(left axis) and histograms show token counts per bin (right axis).
It can be observed that our method exhibits a clear positive correlation with ground-truth entropy change, which strongly supports our theoretical framework.
By contrast, the estimation scheme in~\citep{cui2025entropy} exhibits no clear correlation.

\begin{figure}[t]
    \centering
    \includegraphics[width=0.98\linewidth, height=4.5cm]{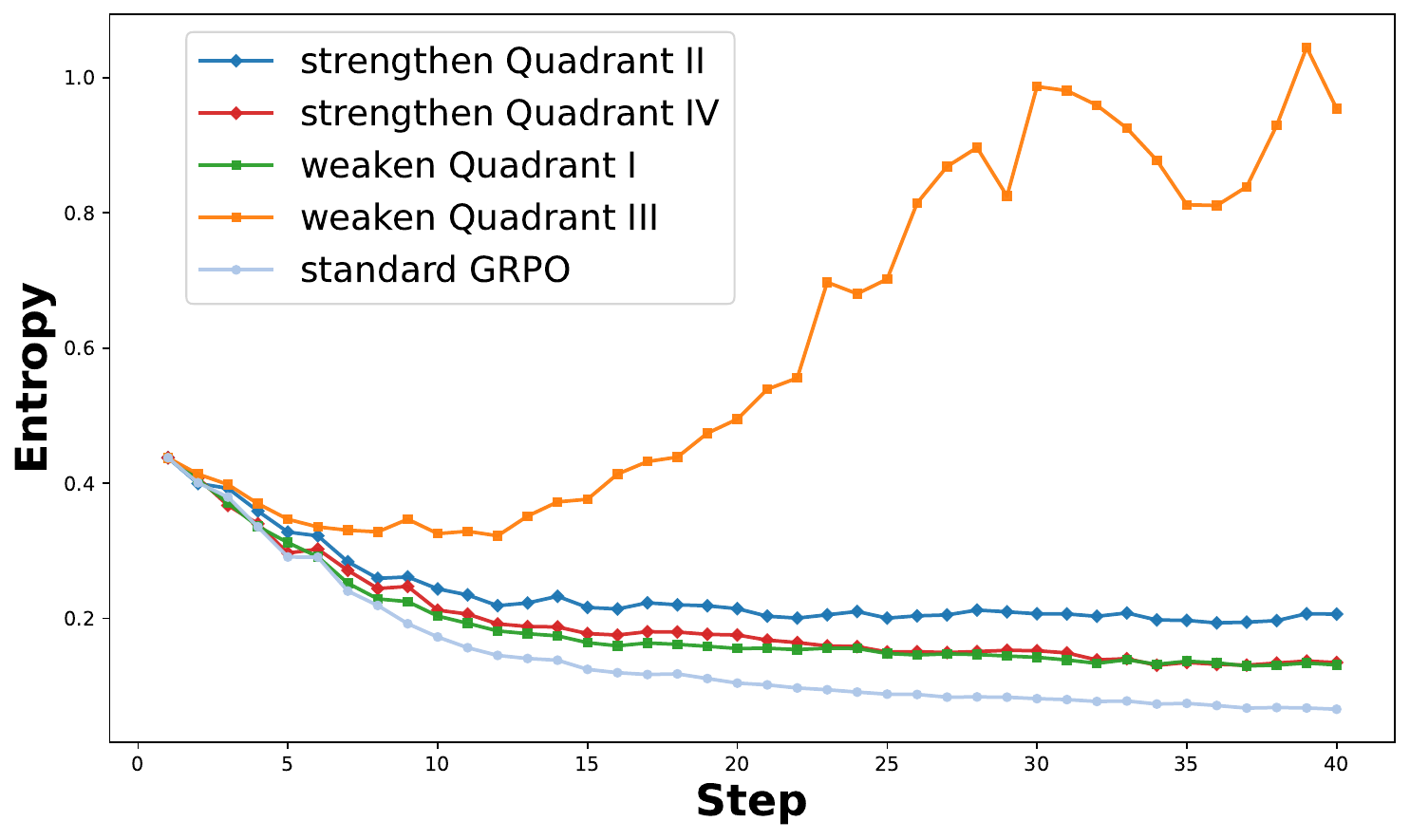}
    \caption{Four schemes to uplift entropy based on advantage-probability effect.}
    \label{strength_weaken_EMA_s0.4_max40}
\end{figure}

\appsubsection{Influencing Entropy Dynamics by Strengthening or Weakening the Quadrants} 
\label{appendix_strength_weaken}
We next ask whether these theoretical findings of quadrant-level tendencies can be used to actively steer entropy in practice.
Guided by the above quadrant-level tendencies, we design a simple intervention on 10\% of tokens in each quadrant to increase entropy: for the entropy-increasing quadrants (II and IV), we double-weight their updates, whereas for the entropy-decreasing quadrants (I and III), we mask their updates, and then track the resulting policy entropy.
As shown in Figure~\ref{strength_weaken_EMA_s0.4_max40}, all four interventions consistently increase policy entropy compared to the standard GRPO baseline, supporting our analysis.
For experiments in Figure~\ref{strength_weaken_EMA_s0.4_max40}, we randomly select samples with a generation probability greater than $0.8$ and an advantage greater than $0$, as well as those with a generation probability less than $0.2$ and an advantage less than 0, and randomly mask $10\%$ of such tokens.
Similarly, for samples with a generation probability greater than $0.8$ and an advantage less than $0$, or a generation probability less than $0.2$ and an advantage greater than $0$, we set the token weight for $10\%$ of such tokens to twice the original token weight.

\begin{figure*}[t]
  \centering
  \begin{subfigure}{0.45\textwidth}
    \centering
    \includegraphics[width=0.95\linewidth]{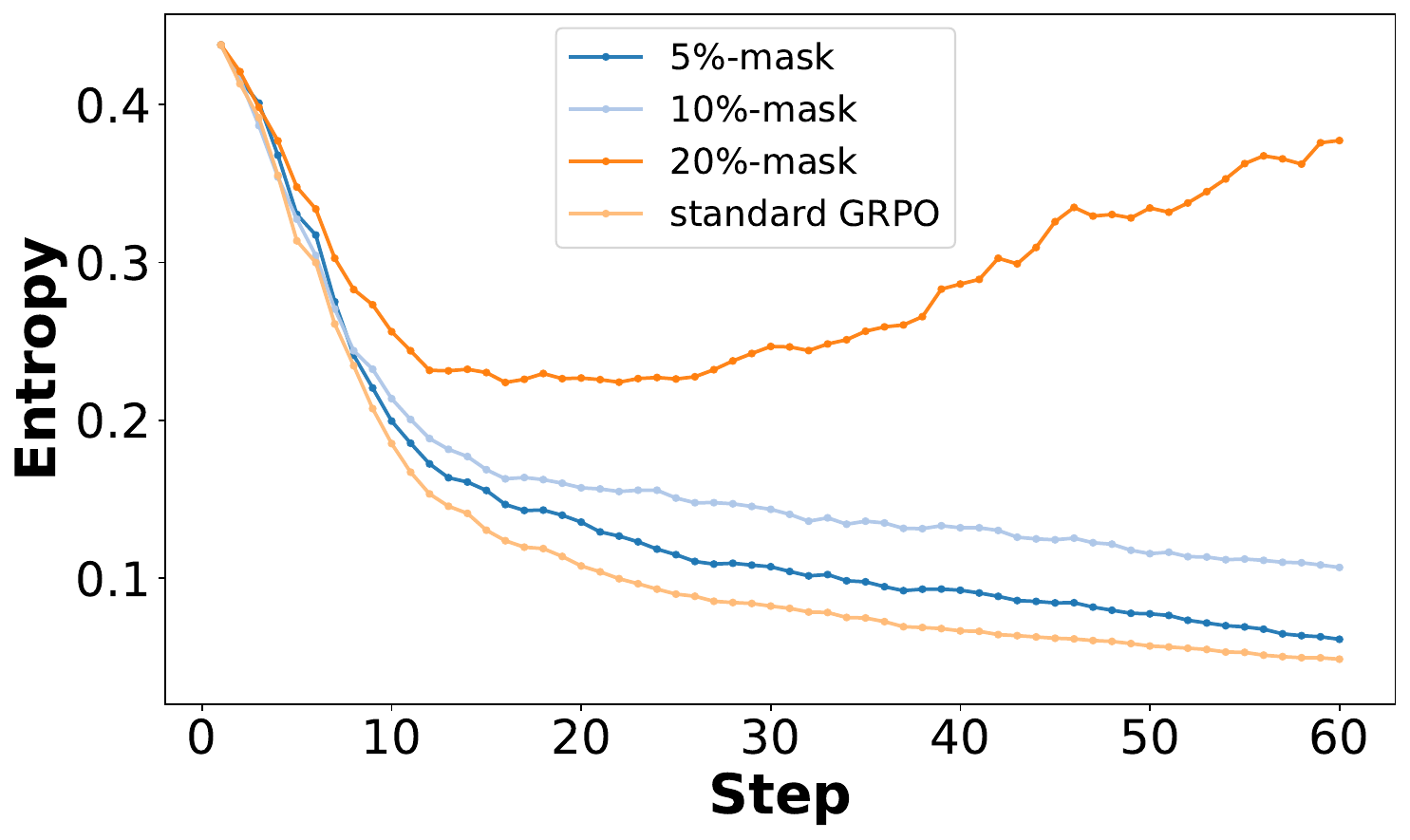}
    \caption{Masking Quadrant I.}
    \label{Masking Quadrant I.}
  \end{subfigure}
  \hfill
  \begin{subfigure}{0.45\textwidth}
    \centering
    \includegraphics[width=0.95\linewidth]{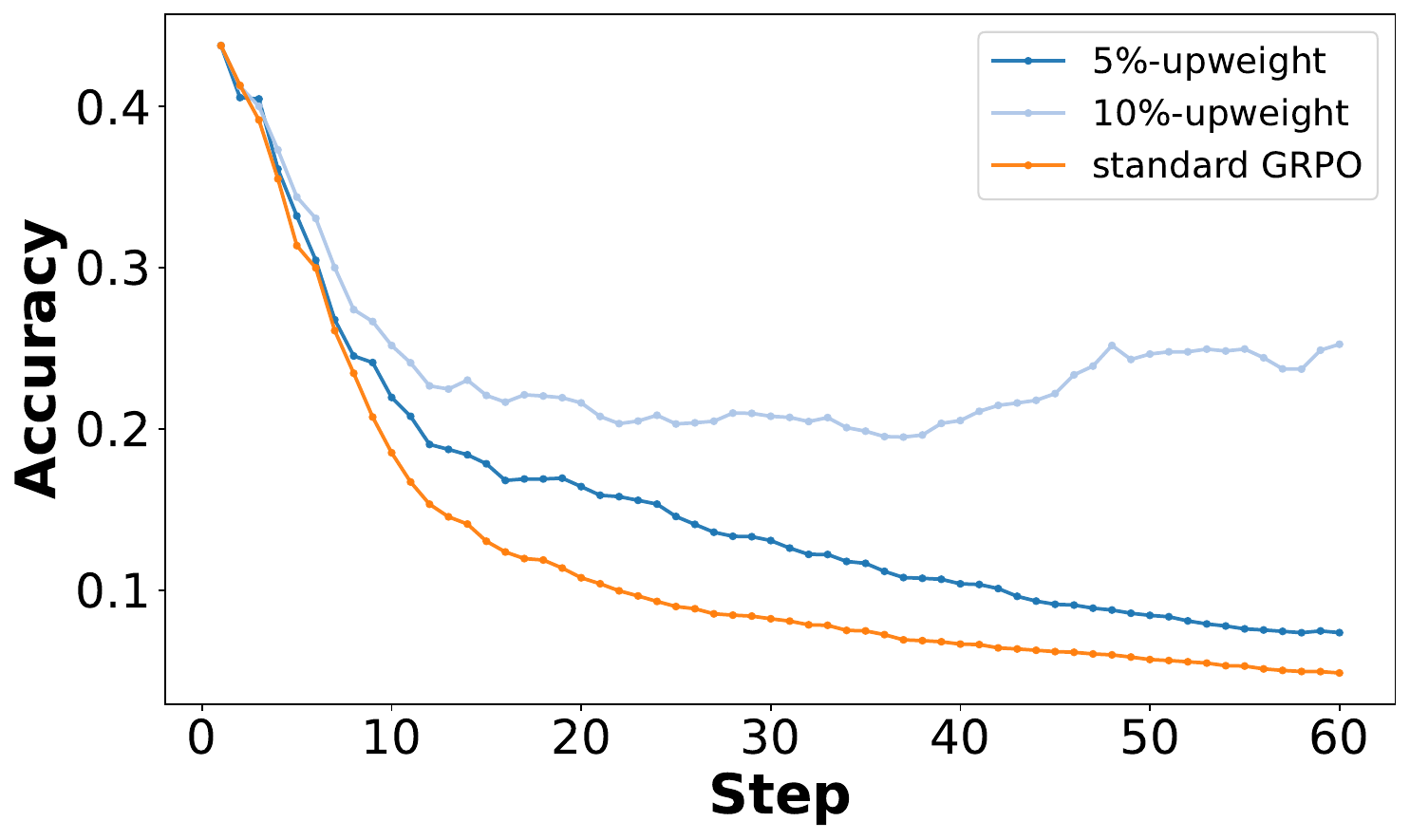}
    \caption{Up-weighting Quadrant II.}
    \label{Up-weighting Quadrant II.}
  \end{subfigure}

  \begin{subfigure}{0.45\textwidth}
    \centering
    \includegraphics[width=0.95\linewidth]{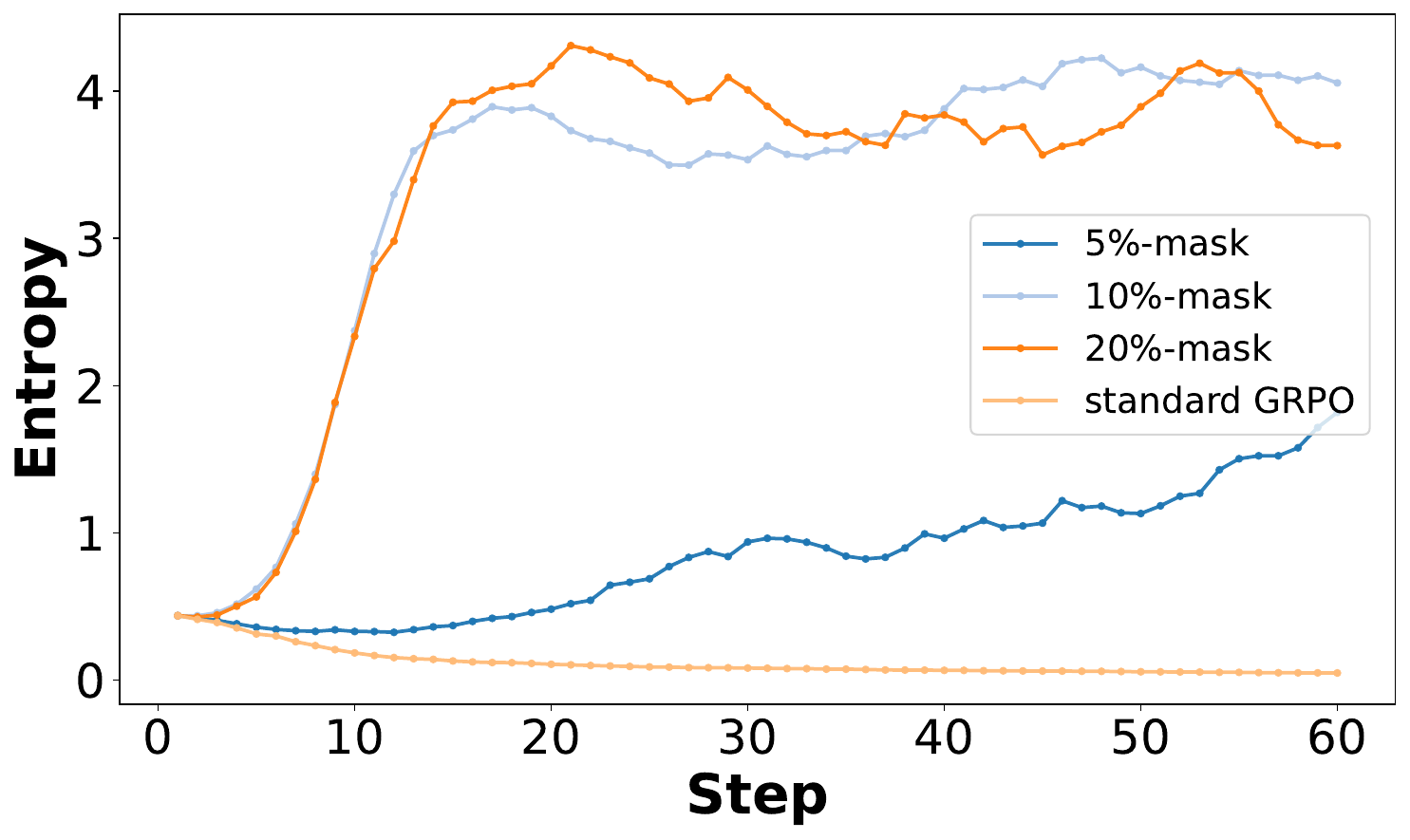}
    \caption{Masking Quadrant III.}
    \label{Masking Quadrant III.}
  \end{subfigure}
  \hfill
  \begin{subfigure}{0.45\textwidth}
    \centering
    \includegraphics[width=0.95\linewidth]{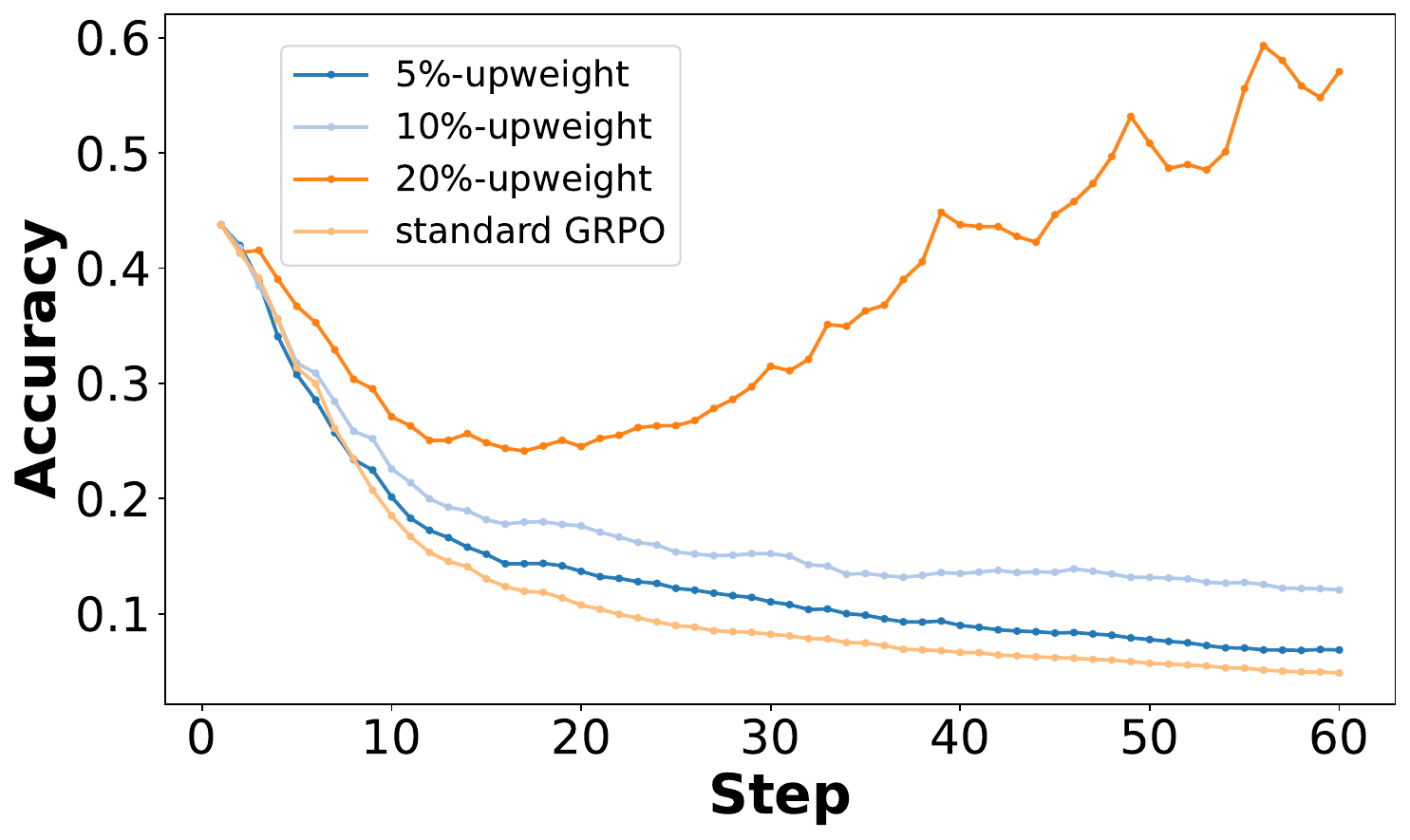}
    \caption{Up-weighting Quadrant IV.}
    \label{Up-weighting Quadrant IV.}
  \end{subfigure}
  \caption{Four cases that increase policy entropy.}
  \label{Increasing entropy in four cases.}
\end{figure*}

\begin{figure*}[t]
  \centering
  % 第一行
  \begin{subfigure}{0.45\textwidth}
    \centering
    \includegraphics[width=0.95\linewidth]{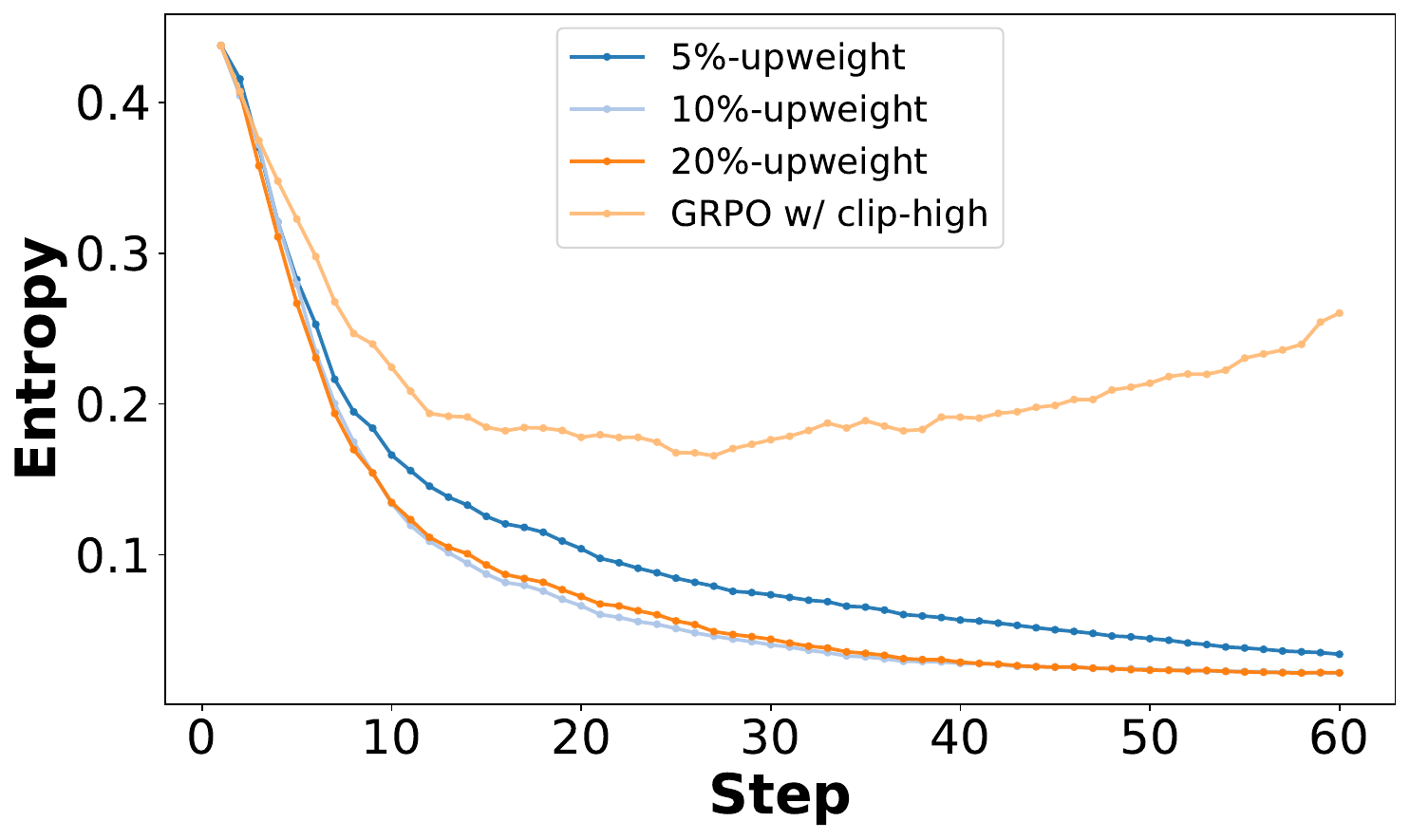}
    \caption{Up-weighting Quadrant I.}
    \label{Up-weighting Quadrant I.}
  \end{subfigure}
  \hfill
  \begin{subfigure}{0.45\textwidth}
    \centering
    \includegraphics[width=0.95\linewidth]{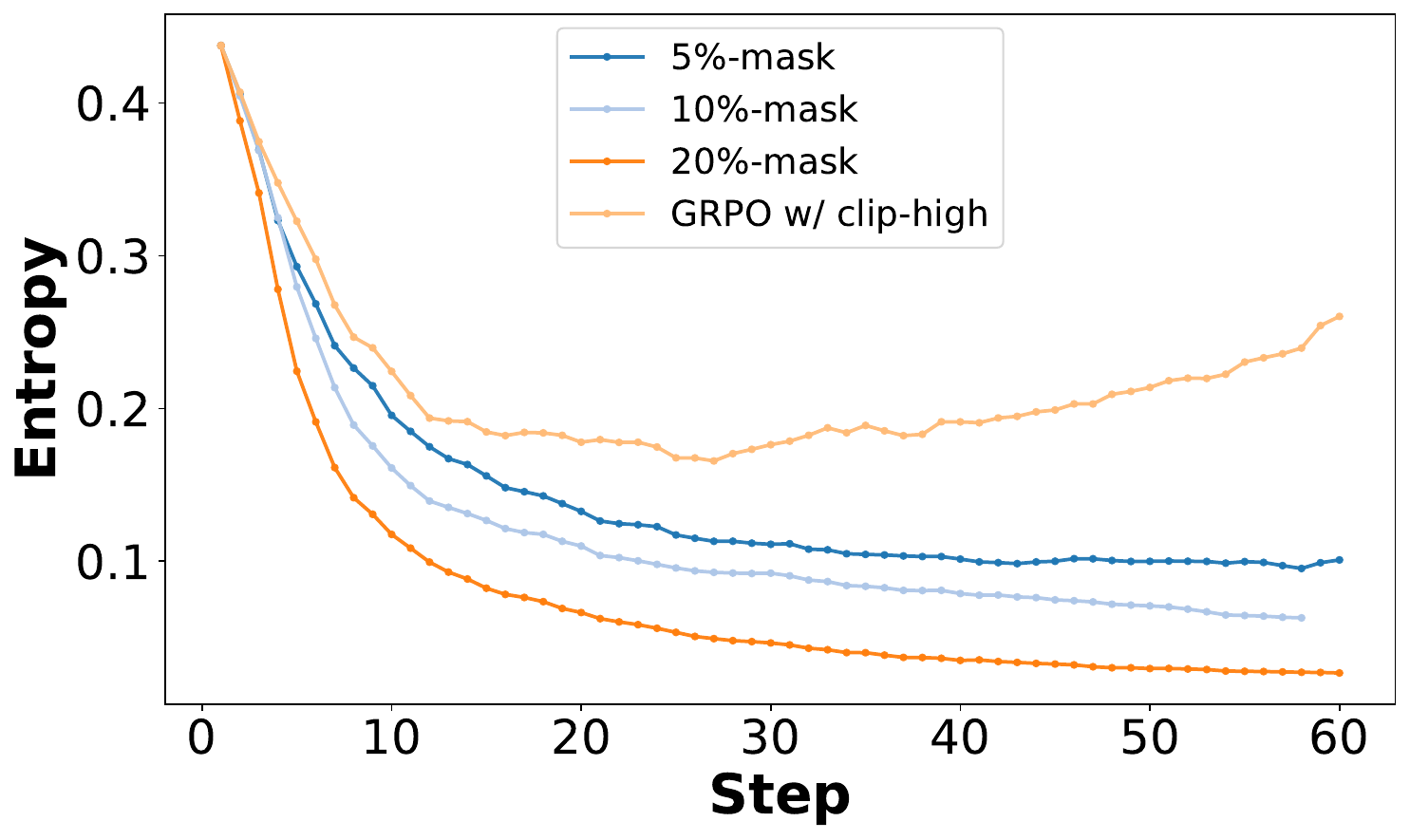}
    \caption{Masking Quadrant II.}
    \label{Masking Quadrant II.}
  \end{subfigure}
  % 第二行
  \begin{subfigure}{0.45\textwidth}
    \centering
    \includegraphics[width=0.95\linewidth]{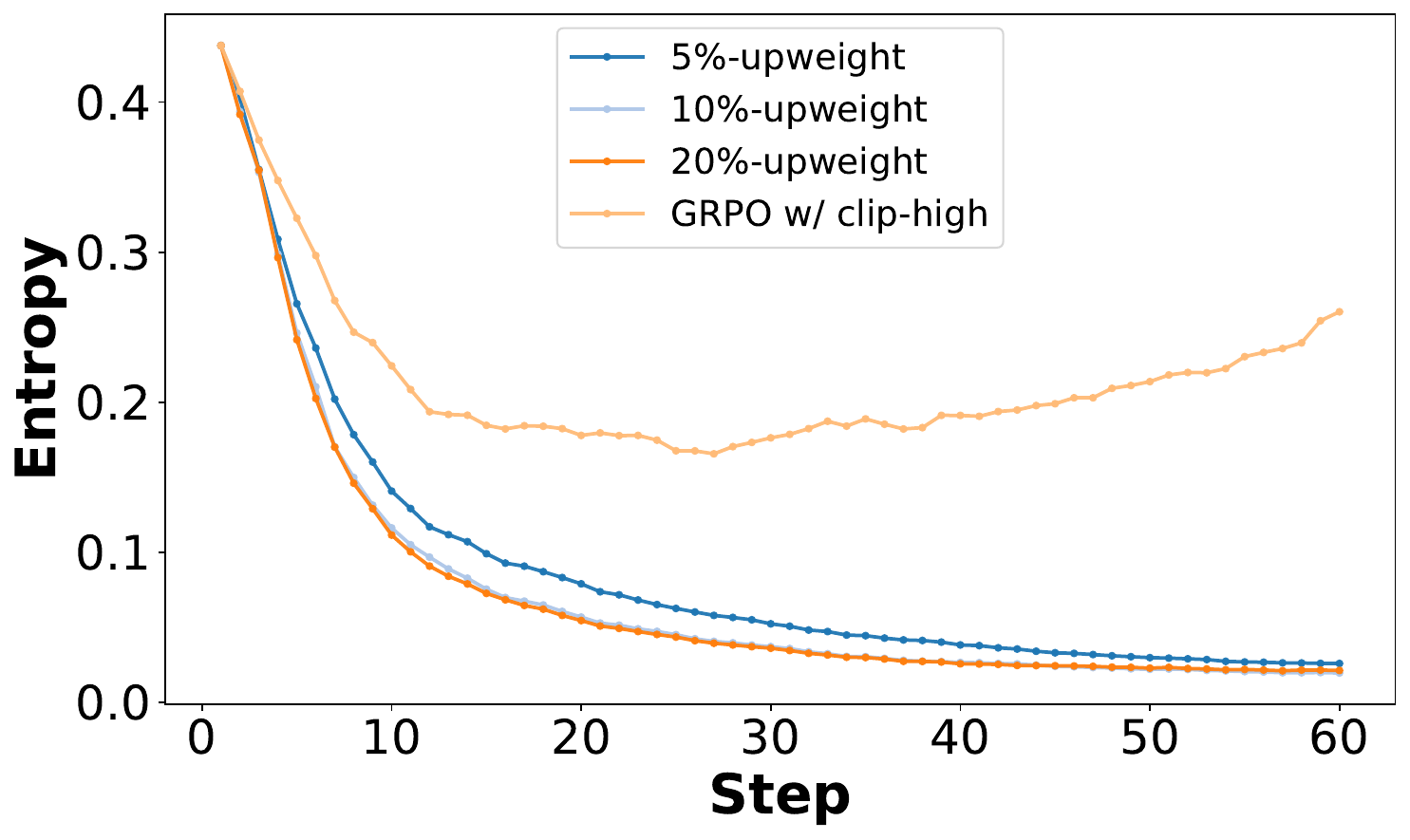}
    \caption{Up-weighting Quadrant III.}
    \label{Up-weighting Quadrant III.}
  \end{subfigure}
  \hfill
  \begin{subfigure}{0.45\textwidth}
    \centering
    \includegraphics[width=0.95\linewidth]{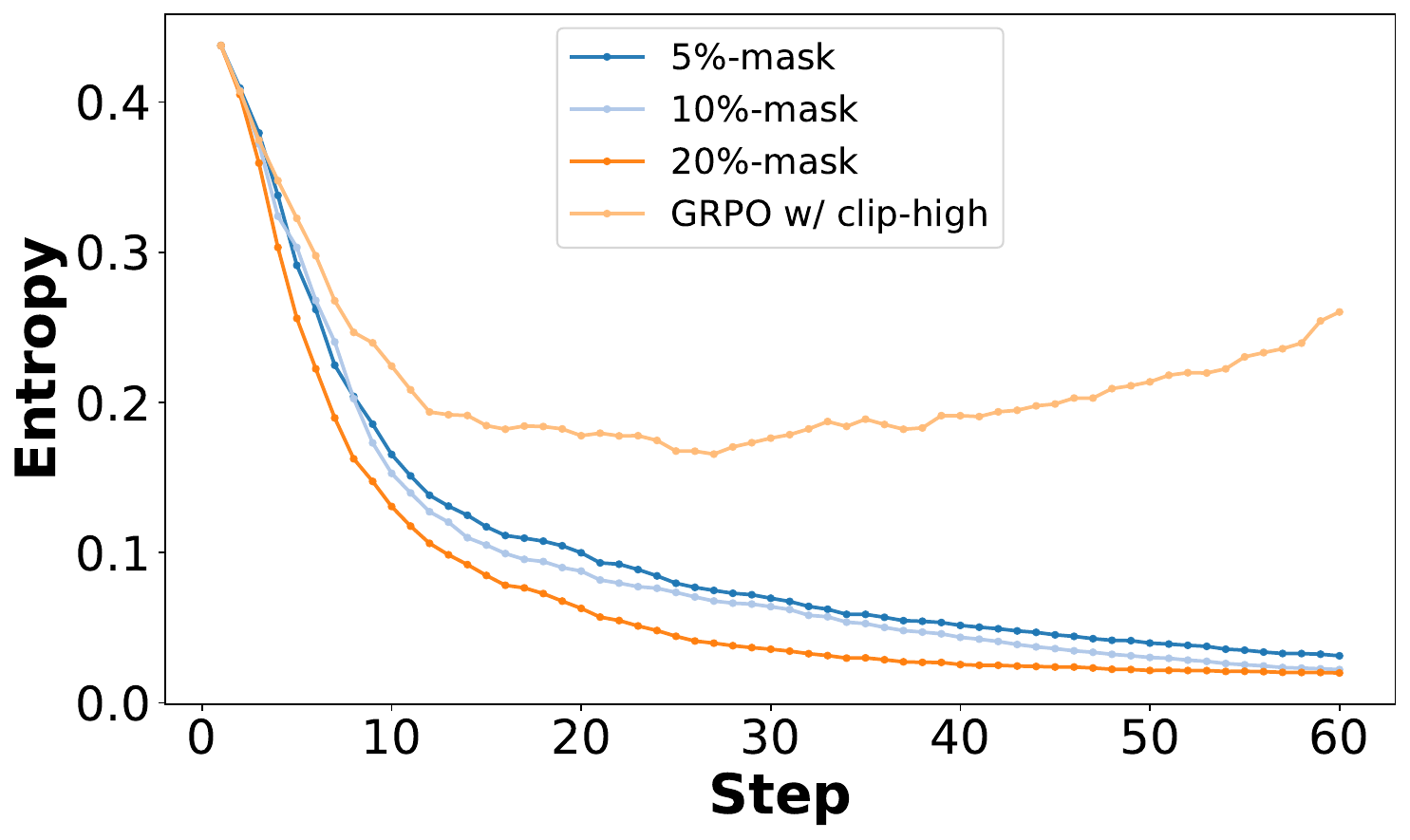}
    \caption{Masking Quadrant IV.}
    \label{Masking Quadrant IV.}
  \end{subfigure}
  \caption{Four cases that decrease policy entropy.}
  \label{Decreasing entropy in four cases.}
\end{figure*}

\begin{table*}[t]
  \centering
  \setlength{\tabcolsep}{10pt}
  \renewcommand{\arraystretch}{1.1}
  \begin{tabular}{@{}l l l@{}}
    \toprule
    \textbf{Dataset} & \textbf{Samples} & \textbf{Level} \\
    \midrule
    \multicolumn{3}{@{}l}{\textit{Training Dataset}} \\
    \quad DAPO-Math-17k~\citep{yu2025dapo}       & 17,398 & Olympiad \\
    \midrule
    \multicolumn{3}{@{}l}{\textit{Test Datasets}} \\
    \quad AIME24~\citep{li2024numinamath}         & 30     & Olympiad \\
    \quad AIME25~\citep{li2024numinamath}         & 30     & Olympiad \\
    \quad AMC23~\citep{li2024numinamath}          & 40     & Intermediate \\
    \quad MATH500~\citep{hendrycks2021measuring}  & 500    & Advanced \\
    \quad Minerva~\citep{lewkowycz2022solving}    & 272    & Graduate \\
    \quad OlympiadBench~\citep{he2024olympiadbench} & 675  & Olympiad \\
    \bottomrule
  \end{tabular}
  \caption{The statistics of dataset for math reasoning.}
  \label{dataset-stats}
\end{table*}

\begin{table*}[t]
  \centering
  \setlength{\tabcolsep}{10pt}
  \renewcommand{\arraystretch}{1.1}
  \begin{tabular}{@{}l l l l@{}}
    \toprule
    \textbf{Dataset} & \textbf{Samples} & \textbf{Task} & \textbf{Source} \\
    \midrule
    \multicolumn{4}{@{}l}{\textit{Training Datasets}} \\
    \quad ArcherCodeR              & 6,753  & Generation & DeepCoder, CodeContests, CodeForces \\
    \quad Internal Code-Edit Corpus & 51,474 & Editing    & Internal user development \\
    \midrule
    \multicolumn{4}{@{}l}{\textit{Test Datasets}} \\
    \quad LiveCodeBench v5         & 279    & Generation & Competitive programming \\
    \quad Internal Held-out Test   & 3,314  & Editing    & Internal user development \\
    \quad Zeta                     & 33     & Editing    & Public code-edit benchmark by Zed \\
    \bottomrule
  \end{tabular}
  \caption{The statistics of datasets for code generation and code editing.}
  \label{code-dataset-stats}
\end{table*}

\coloredtext{To further validate the patterns of entropy change with advantage and probability in Figure~\ref{Advantage_and_Probability}, we strengthen (up-weighting) or weaken (masking) each of the four quadrants at different intensities to induce entropy increases or decreases, respectively.
Unlike the setup in Figure~\ref{strength_weaken_EMA_s0.4_max40} where $10\%$ tokens are intervened, we present a more comprehensive validation here.}

\coloredtext{Figure~\ref{Increasing entropy in four cases.} shows interventions applied to each quadrant with the goal of increasing entropy, using standard GRPO ($\varepsilon_{\text{high} = 0.2}, \varepsilon_{\text{low} = 0.2}$) as the baseline; while Figure~\ref{Decreasing entropy in four cases.} presents interventions with the goal of  decreasing entropy across the four quadrants, using GRPO w/ clip-high ($\varepsilon_{\text{high} = 0.28}, \varepsilon_{\text{low} = 0.2}$) as the baseline;
In each case, the proportion of tokens masked or up-weighted ranges from $5\%$ to $20\%$.
Across all cases, it can be observed that the token-level intervention effects on entropy align with our quantitative analysis framework, and the impact becomes more pronounced as the intervention ratio increases (from $5\%$ to $20\%$).
For example, in Figure~\ref{Up-weighting Quadrant II.}, compared to standard GRPO, up-weighting Quadrant \hyperlink{Quadrant II}{II} yields a marked increase in policy entropy over standard GRPO (we exclude the $20\%$ up-weight case because it produces excessively high entropy).
This indicates that the clip-high mechanism in DAPO~\citep{yu2025dapo} and unlikeliness~\citep{he2025rewarding} can be viewed as a special instance of this intervention.
In summary, the overall entropy dynamics arise from the joint contributions of the four quadrants; perturbing any one of them can induce a predictable change in the total entropy from our analysis framework.}

\appsubsection{Entropy Effect of Clipping Operation}
In this subsection, we adjust the clipping thresholds to steer entropy change in GRPO, and evaluate RLVR performance under different entropy levels.

The entropy dynamics induced by clip operation is shown in Figure~\ref{clip-high} and~\ref{clip-low} in the main text.
\emph{Clip-high} (\ie $\varepsilon_{\text{high}}$): entropy decreases in the early phase for all settings; larger $\varepsilon_{\text{high}}$ leads to a clear late-stage entropy rebound and sustained growth, whereas small $\varepsilon_{\text{high}}$ yields continued decay and low final entropy.
\emph{Clip-low} (\ie $\varepsilon_{\text{low}}$): the behavior is more bifurcated---with $\varepsilon_{\text{low}}=0.1$, entropy increases monotonically over training, while $\varepsilon_{\text{low}} \geq 0.2$ drives entropy rapidly toward (near) zero, exhibiting a much stronger tendency toward entropy collapse.

\begin{table}[htbp]
    \centering
    \small
    \setlength{\tabcolsep}{10pt}
    \renewcommand{\arraystretch}{1.15}
    \begin{tabular}{lcccc}
        \toprule
        \textbf{Parameters} & $\mathbf{0.1}$ & $\mathbf{0.2}$ & $\mathbf{0.5}$ & $\mathbf{0.8}$ \\
        \midrule
        $\varepsilon_{\text{high}}$ & $44.0$ & $\mathbf{44.2}$ & $43.7$ & $42.3$ \\
        $\varepsilon_{\text{low}}$  & $43.7$ & $\mathbf{44.2}$ & $42.5$ & $42.0$ \\
        \bottomrule
    \end{tabular}
    \caption{Average math reasoning performance with different clip operations.}
    \label{Average math reasoning performance with different clip operations.}
\end{table}

We evaluate these runs individually; the average math reasoning performance under different clipping operations is reported in Table~\ref{Average math reasoning performance with different clip operations.}.
We observe that performance degrades under both entropy collapse and entropy explosion, whereas maintaining entropy within a stable range yields consistently better results.
This highlights the importance of stabilizing entropy dynamics, which is aligned with our proposed STEER.

Why large entropy growth can hurt training? This phenomenon can be explained as follows: excessively high entropy makes the policy overly stochastic, reducing the likelihood of sampling informative reasoning trajectories that provide useful training signals. In GRPO-style training, this may lead to a large fraction of rollouts receiving near-zero rewards, hindering optimization. Similar concerns on “entropy explosion” have also been raised in prior work~\citep{wu2025quantile}.

% \begin{figure*}[htbp]
%     \centering
%     \begin{subfigure}[t]{0.45\textwidth}
%         \centering
%         \includegraphics[width=\linewidth]{figs/clip-high_EMA_s0.4_max80.pdf}
%         \caption{Effect of $\varepsilon_{\text{high}}$}
%         \label{clip-high_app}
%     \end{subfigure}
%     \hfill
%     \begin{subfigure}[t]{0.45\textwidth}
%         \centering
%         \includegraphics[width=\linewidth]{figs/clip-low_EMA_s0.4_max80.pdf}
%         \caption{Effect of $\varepsilon_{\text{low}}$}
%         \label{clip-low_app}
%     \end{subfigure}
%     \caption{Empirical validation of entropy-change mechanisms via ratio clipping.}
%     \label{entropy-psr-nsr-all}
% \end{figure*}

\appsection{Training Settings} \label{training details}

\appsubsection{Detailed Information for Dataset}

\paragraph{Math Reasoning}
We use DAPO-Math-17k as the training dataset for enhancing model's math reasoning, which is a competition-style math reasoning dataset introduced in the DAPO work.
It contains roughly 17k problem–solution pairs of olympiad-level mathematics (covering algebra, number theory, combinatorics, geometry, etc.), derived from standard public math-reasoning benchmarks and reformatted into supervised trajectories suitable for RLVR training.

Table~\ref{dataset-stats} reports detailed statistics of the datasets used in math reasoning experiments, including the number of questions and difficulty levels. These benchmarks are all widely used to evaluate mathematical reasoning ability.

\paragraph{Coding Tasks}
For code generation task, we adopt ArcherCodeR\footnote{Available at \url{https://huggingface.co/datasets/Fate-Zero/ArcherCodeR-Dataset}.} for RLVR training, which contains 6753 code generation tasks sourced 
from DeepCoder\footnote{\url{https://huggingface.co/datasets/agentica-org/DeepCoder-Preview-Dataset}}, 
CodeContests\footnote{\url{https://huggingface.co/datasets/deepmind/code_contests}}, 
and CodeForces\footnote{\url{https://huggingface.co/datasets/open-r1/codeforces}}, with regenerated 
test cases to reduce false positives during RL training.
We adopt the widely used LiveCodeBench v5~\citep{jain2024livecodebench} for evaluation.
As for code edit task, we build a large-scale corpus of practical code-editing examples—over 50000 instances—collected from internal users and representative of their day-to-day software development workflows.
We then evaluate the model on both our internal held-out test split (3314 cases) and the Zeta benchmark. Internal refers to our in-house, real-world code-editing evaluation set.
Zeta is a public code-edit benchmark~\citep{zed_zeta_dataset_2025} and we evaluate on its “dpo” subset, since the “eval”
subset contains only 33 examples.
Table~\ref{code-dataset-stats} reports detailed statistics of the datasets used in coding experiments.

Then we detail the construction of the internal training and test code-edit data.
Each example contains two fields: \texttt{<prompt>} and \texttt{<edit>}. The \texttt{<prompt>} field bundles all required inputs, including the system prompt, the surrounding code context, an ordered sequence of prior edits, the target edit range annotated with the cursor position, and any user-provided hints. The \texttt{<edit>} field provides the reference (ground-truth) edit.

\begin{table*}[t]
\centering
\renewcommand{\arraystretch}{1}
\begin{tabular}{lcccccccc}
\toprule
\textbf{Mapping} & \textbf{AIME24} & \textbf{AIME25} & \textbf{AMC23} & \textbf{MATH500} & \textbf{Minerva} & \textbf{Olympiad} & \textbf{Avg.} \\
\midrule
exponential & $36.2$ & $16.1$ & $72.1$ & $82.2$ & $41.7$ & $43.0$ & $48.6$ \\
linear      & $36.0$ & $15.7$ & $73.6$ & $81.8$ & $39.4$ & $41.8$ & $48.0$ \\
binary      & $32.5$ & $14.7$ & $71.3$ & $80.9$ & $38.2$ & $41.5$ & $46.5$ \\
\bottomrule
\end{tabular}
\caption{Ablation study on different weight mapping modes.}
\label{Ablation Study}
\end{table*}

\begin{figure*}[htbp]
    \centering
    \begin{minipage}{0.3\textwidth}
        \centering
    \includegraphics[width=0.9\linewidth, height=4.1cm]{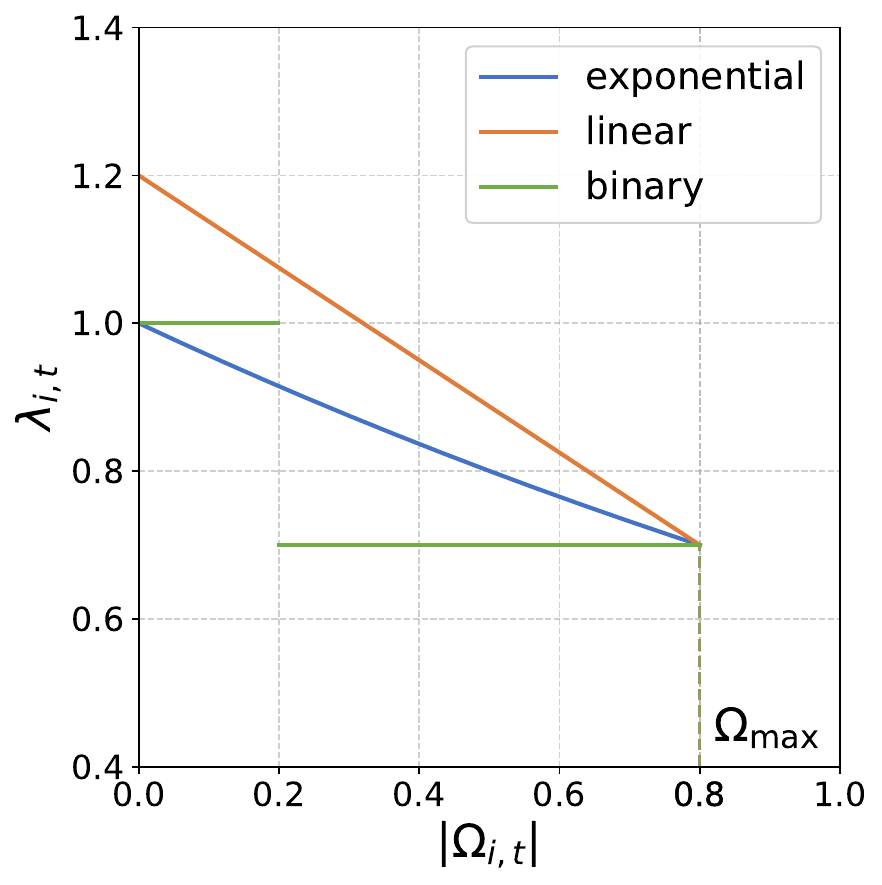}
    \caption{Weight Mapping.}
    \label{Mapping modes.}
    \end{minipage}
    \begin{minipage}{0.68\textwidth}
        \centering
    \includegraphics[width=0.95\linewidth, height=4.1cm]{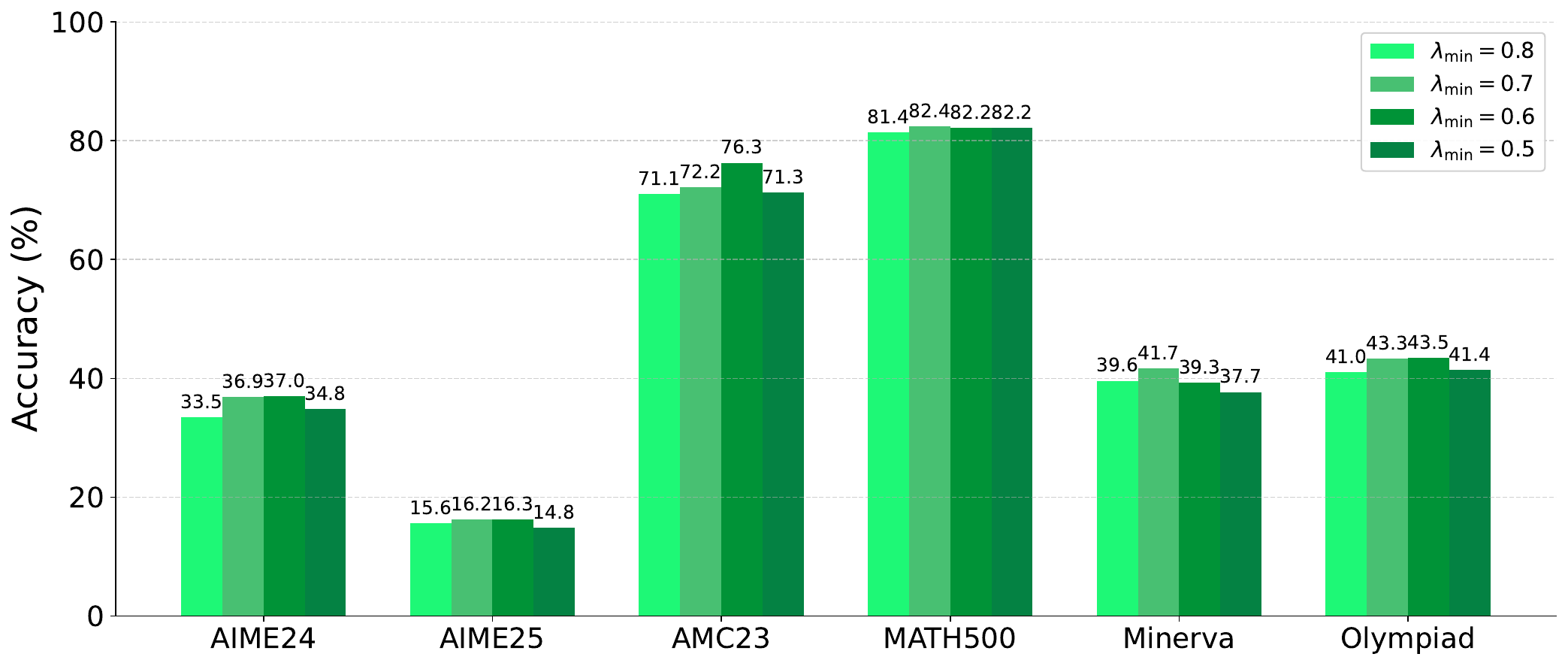}
    \caption{Hyperparameter Sensitivity on $\lambda_\text{min}$.}
    \label{Hyperparameter Sensitivity.}
    \end{minipage}
\end{figure*}

\appsubsection{Training Details for Our Method and Baselines}\label{Training Details for our method and baselines}

All algorithms are implemented based on the official GRPO codebase within the verl framework.

\paragraph{Training settings}
Generation batch size is set to $512$, and update batch size is set to $32$.
The number of rollouts is set to $8$.
Both the KL-divergence and entropy loss terms are removed in our experiments.
Training is performed with top-p value of $1.0$ and temperature is $1.0$. 
We use a learning rate of 1e-6 without warm-up across all experiments.
At each rollout step, we generate $8$ rollouts for each of $512$ sampled questions with a mini-batch size of $32$ for updating policy model.
Models are trained for at most $200$ rollout steps.
Unless otherwise specified, we follow GRPO’s default design choices with token-level loss normalization without dynamic sampling and KL regularization.
For Qwen2.5 series models, the maximum input length is $1024$ and the maximum output length is $3072$.
For Qwen2.5-Coder series models in code editing task, the maximum input length is $4096$ and the maximum output length is $1024$.

\paragraph{Evaluation settings}
Validation is performed with a top-p value of $0.7$ and temperature is $1.0$ across all models and test sets.
We use Math-Verify and Qwen-Verify for both validation during training and final evaluation.
All evaluations are \textit{zero-shot} with no additional prompts.
\coloredtext{All methods save a checkpoint every $10$ steps, and the checkpoint achieving the highest AIME24 accuracy is selected for test.}
All experiments were conducted on a cluster equipped with NVIDIA H20 GPUs.

\paragraph{Specific settings for baselines}
Table~\ref{main_results} reports the main results, and this subsection details the training setups for all compared baselines.
For GRPO~\citep{shao2024deepseekmath}, we follow the official VeRL training recipe and keep all hyperparameters unchanged.
For SimpleRL-Zoo~\citep{zeng2025simplerl}, we adopt the original training recipe from the official repository, replacing the original math corpus with DAPO-Math-17k as the training data.
For Eurus-PRIME~\citep{cui2025process}, we directly use the publicly released checkpoint that is trained on Qwen2.5-Math-7B with process reward.
For OPO~\citep{hao2025policy}, we follow the reference implementation in VeRL without new hyperparameters introduced.
GRPO w/ clip-high~\citep{yu2025dapo} sets the upper clipping threshold to $\varepsilon_{\text{high}} = 0.28$ while keeping all other settings identical to GRPO.
GRPO w/ Entro. Loss~\citep{schulman2017proximal} augments the GRPO objective with an entropy loss term; we tune its weight over $\{0.01, 0.001, 0.0001\}$ and report the best result.
GRPO w/ Fork Tokens follows the strategy in~\citep{wang2025beyond}:
using only policy gradients of the top 20\%, 30\%, or 40\% highest-entropy tokens and report the best-performing configuration.
W-REINFORCE~\citep{zhu2025surprising} is implemented by assigning a reduced weight to positive samples, tuning $\lambda \in \{0.1, 0.2\}$ and reporting the best result.
Entro. Adv.~\citep{cheng2025reasoning} is reproduced following the original paper, fixing $\kappa = 2$ for all experiments and setting $\alpha = 0.4$.
Clip-Cov and KL-Cov~\citep{cui2025entropy} apply clipping or KL-penalty constraints only to a small subset of generated tokens.
Following the original implementation, we set the fraction of constrained tokens to 0.0002.

\begin{table*}[t]
\vspace{-0.5em}
\centering
\definecolor{ourrow}{RGB}{235,243,252}
{\setlength{\tabcolsep}{5pt}
\renewcommand{\arraystretch}{0.85}
\vspace{-0.5em}
\begin{tabular}{lccccccc}
\toprule
\textbf{Method} & \textbf{AIME24} & \textbf{AIME25} & \textbf{AMC23} & \textbf{MATH500} & \textbf{Minerva} & \textbf{Olympiad} & \textbf{Avg.} \\
\midrule

\multicolumn{8}{c}{\textbf{Qwen2.5-Math-1.5B}} \\
\midrule
Base         & $4.1$  & $2.1$  & $24.7$ & $29.0$ & $9.2$  & $20.5$ & $14.9$ \\
GRPO         & $16.2$ & $7.6$  & $56.0$ & $74.4$ & $26.1$ & $34.6$ & $35.8$ \\
OPO          & $14.8$ & $9.0$  & $58.2$ & $72.2$ & $26.1$ & $35.9$ & $36.0$ \\
Entro. Adv.  & $15.0$ & $9.1$  & $55.7$ & $70.2$ & $26.8$ & $34.9$ & $35.3$ \\
Clip-Cov     & $14.7$ & $8.4$  & $56.0$ & $72.8$ & $26.4$ & $34.9$ & $35.5$ \\
\rowcolor{ourrow} \textbf{STEER} & $\mathbf{17.4}$ & $\mathbf{9.7}$ & $\mathbf{61.6}$ & $\mathbf{75.5}$ & $\mathbf{28.2}$ & $\mathbf{36.6}$ & $\mathbf{38.2}$ \\

\bottomrule
\end{tabular}
\caption{Benchmark results of different methods on Qwen2.5-Math-1.5B. We report avg@32 for AIME24, AIME25, and AMC23 and avg@1 for others. All results are presented as percentages.}
\label{1.5B_results}
}
\end{table*}

\appsection{Supplementary Performance Evaluation}\label{Supplementary Performance Evaluation}
This section presents additional performance results that assess the robustness and generality of STEER under varied settings.

\begin{figure}[t]
    \centering
    \includegraphics[width=0.85\linewidth, height=3.8cm]{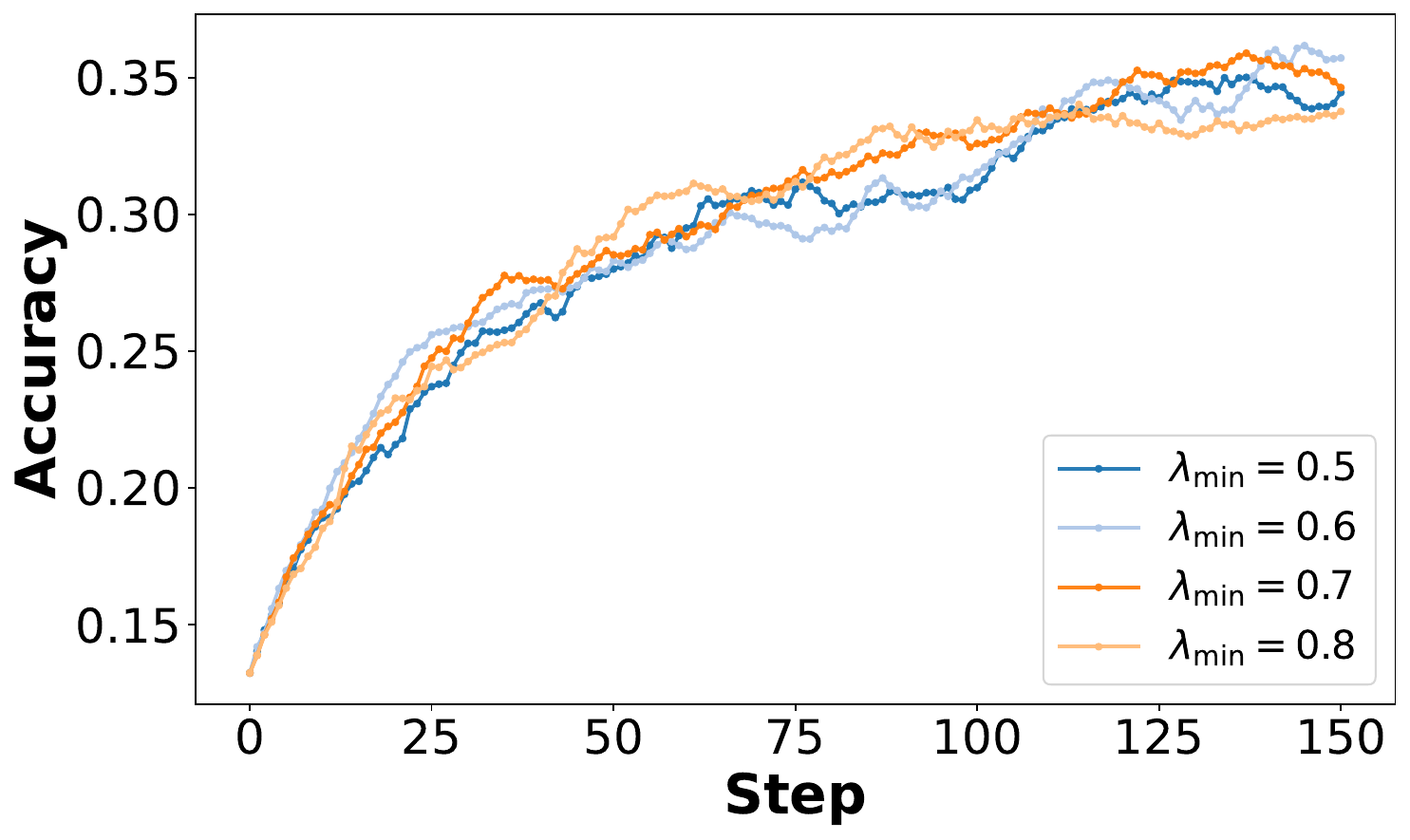}
\captionsetup{justification=centering,singlelinecheck=false}
    \caption{STEER's test accuracy dynamics under different hyperparameter $\lambda_{\min}$.}
    \label{test_acc_plot_para_EMA_s0.8_max150}
\end{figure}

\begin{figure}[t]
\vspace{-0.5em}
    \centering
    \includegraphics[width=0.98\linewidth, height=7.5cm]{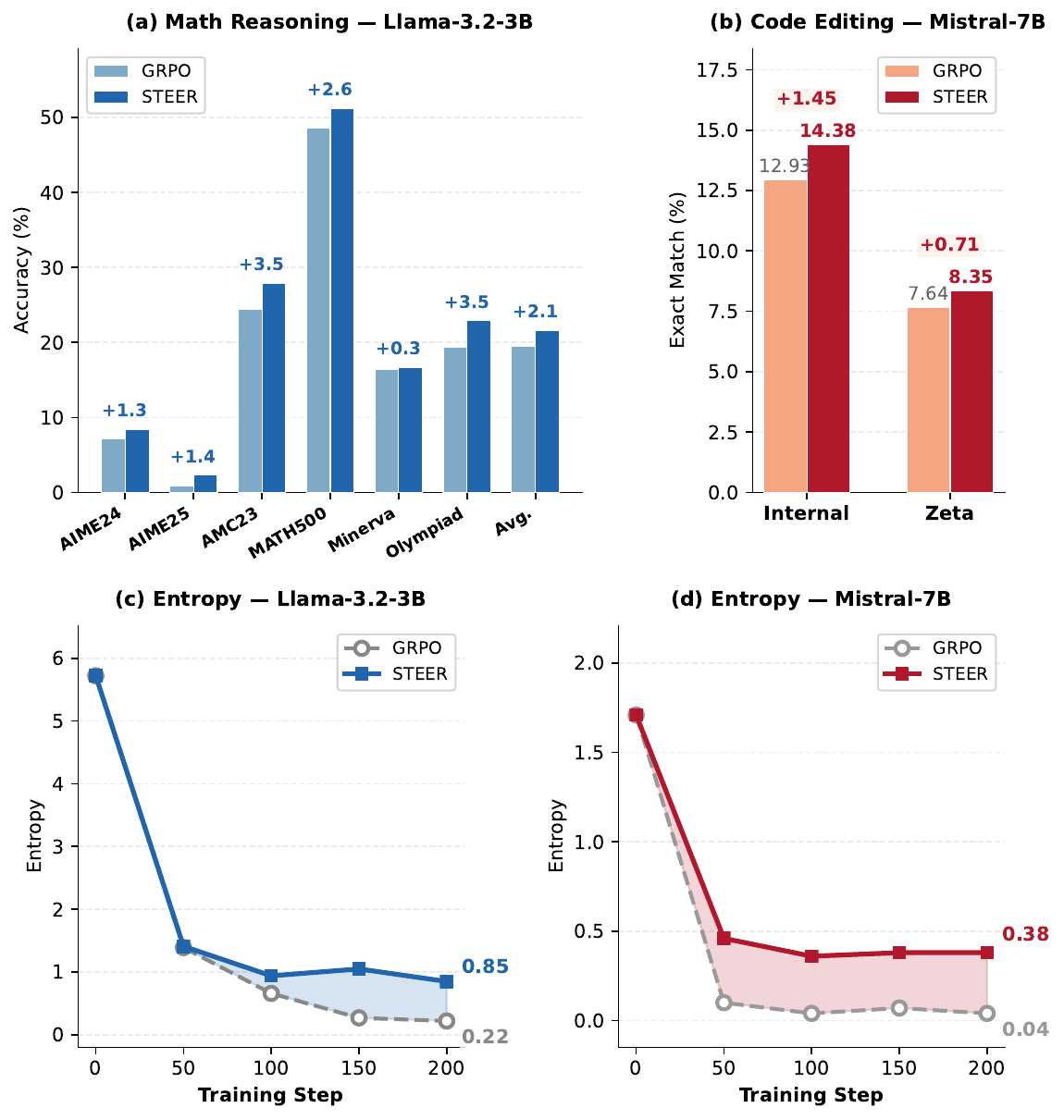}
    \vspace{-0.5em}
    \caption{Performance comparison and entropy on math reasoning and coding tasks on other base models.}
    \label{Results on math and code tasks on other base models.}
    \vspace{-1em}
\end{figure}

\begin{figure}[t]
\vspace{-0.8em}
    \centering
    \includegraphics[width=0.95\linewidth, height=8cm]{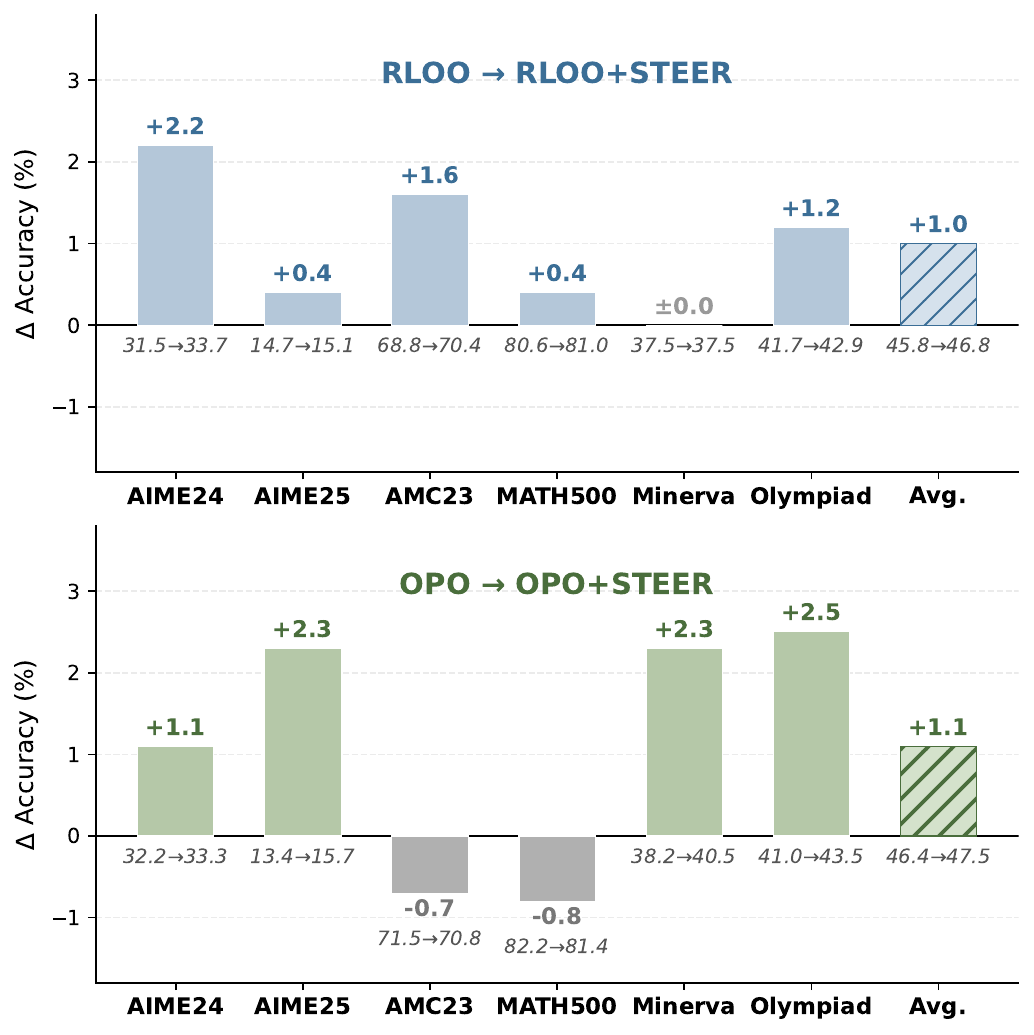}
    \vspace{-0.5em}
    \caption{Results on math reasoning tasks based on other RL algorithms.}
    \label{Results on math reasoning tasks based on other RL algorithms.}
    \vspace{-0.8em}
\end{figure}

\appsubsection{Ablation Study}
Besides the exponential mapping in Eq.~\eqref{weight_mapping}, we consider the following linear mapping and binary mapping for ablation:
\begin{equation*}
\begin{aligned}
&\text{linear:}\quad
\lambda_{i,t} = \lambda_{\text{max}} - 
    \frac{\lambda_{\text{max}} - \lambda_{\text{min}}}{\Omega_{\text{max}} - \Omega_{\text{min}}}
    \left( \Omega_{i,t} - \Omega_{\text{min}} \right), \\
&\text{binary:}\quad
\lambda_{i,t} =
\begin{cases}
\lambda_{\text{min}}, & \Omega_{i,t} > Q_{\xi}(\Omega), \\
1, & \text{otherwise}.
\end{cases}
\end{aligned}
\end{equation*}
We set $\lambda_{\text{min}}=0.7$ throughout and $\lambda_{\text{max}} = 1.2$ for the linear mapping.
For the binary mapping, the quantile threshold $Q_{\xi}$ is set to $Q_{0.8}$—i.e., the top $20\%$ of tokens by $\Omega_{i,t}$ are assigned weight $\lambda_{\min}$ and the remaining $80\%$ of tokens keep weight $1$.
The three mapping schematics are illustrated in Figure~\ref{Mapping modes.}, and their performance on Qwen2.5-Math-7B is reported in Table~\ref{Ablation Study}, each is the average of two runs.
It can be seen that the binary mapping degrades performance, whereas the linear mapping does not materially harm performance.
This highlights the necessity of continuous token-level reweighting, as truncation cannot precisely control entropy change.

\appsubsection{Hyperparameter Sensitivity}
We also assess the sensitivity of the experimental results to hyperparameters $\lambda_\text{min}$ in Eq.~\eqref{weight_mapping}.
An excessively small $\lambda_\text{min}$ may hinder the model's learning and lead to unstable training, while an excessively large $\lambda_\text{min}$ reduces the model's ability to control entropy.
As shown in Figure~\ref{Hyperparameter Sensitivity.}, our method performs consistently well when $\lambda_\text{min} \in [0.5, 0.8]$.
Figure~\ref{test_acc_plot_para_EMA_s0.8_max150} presents the test accuracy curves for different hyperparameters, demonstrating both stability and superiority of STEER.

\appsubsection{Empirical Results Extended to Other Base Models and Other RL Algorithms}\label{Empirical Results Extended to Other Base Models and Other RL Algorithms}
We add experiments on additional backbones, including Qwen2.5-Math-1.5B, Llama-3.2-3B-Instruct~\citep{grattafiori2024llama} for math reasoning and Mistral-7B-v0.3~\citep{jiang2023mistral7b} for code editing. For each task, we use the same STEER hyperparameter as in the main experiments, rather than re-tuned per backbone. The results are reported in Figure~\ref{Results on math and code tasks on other base models.}, and we find that STEER yields consistent performance improvements.

As observed, STEER improves the average score from $19.5 \to 21.6$ and clearly mitigates entropy collapse ($0.22 \to 0.85$ at Step~200). On Mistral-7B-v0.3 for code-edit task, STEER improves exact-match on both Internal ($12.93 \to 14.38$) and Zeta ($7.64 \to 8.35$), again keeping entropy much higher at the end ($0.04 \to 0.38$ at Step~200).
These improvements suggest our proposed method and its hyperparameters transfer beyond the Qwen2.5 family.

We evaluate STEER under RLOO and OPO with the default setting in verl implementations and the rest of the core setup is the same as in the main results.
The results are shown in Figure~\ref{Results on math reasoning tasks based on other RL algorithms.}. 
As shown, STEER improves the average score from $45.8 \to 46.8$ on RLOO and $46.4 \to 47.5$ on OPO, indicating the effectiveness of STEER is not GRPO-specific.

\onecolumn
\clearpage

\appsection{Theorem Proof Details}\label{proof}
During the RLVR training, token logits are shaped by entangled internal parameters, making entropy change difficult to quantify.
To capture the essence of distribution shifts during training, we adopt the following weak assumption.
\begin{assumption}[Parameter-independent softmax]
\label{Parameter independence assumption}
For any context (state) $s = (q,o_{<t})$, each token (action) $a$ in the vocabulary $\mathcal{V}$
is associated with an independent logit parameter $z_{s,a}(\theta)$.
At update step $k$ in training, the next-token distribution of $\pi_\theta^k$ then follows
\[
\pi_\theta^k(\cdot\mid s)=\mathrm{softmax}(z^k(s)),
\]
where $z^k(s)$ is the vector of logit parameters for all actions under state $s$.
\end{assumption}

Assumption~\ref{Parameter independence assumption} states that a gradient step on the sampled token does not substantially affect the logits of the other tokens in the vocabulary.
Given this assumption, we derive the following theorem on token entropy change.

\begin{theorem}[First–order entropy change estimation]
\label{theorem_entropy_change_ap}
Let the policy model $\pi_\theta$ satisfy
Assumption~\ref{Parameter independence assumption}.
For any context (state) $s = (q, o_{<t})$, define the
token–level entropy change between two update steps as
\[
\Delta \mathcal{H}(s)
\triangleq
\mathcal{H}(\pi_{\theta}^{k+1} \mid s)
-
\mathcal{H}(\pi_{\theta}^{k} \mid s) .
\]
Under a single GRPO update in Eq.~\eqref{grpo_Eq},
$\Delta \mathcal{H}(s)$ admits the decomposition
\begin{equation}
\label{eq:delta_H_decomp}
\Delta \mathcal{H}(s)
=
\Omega(s) + \Phi(s),
\end{equation}
where the first–order estimation term is
\begin{equation}
\label{eq:Omega_expanded}
\Omega(s)
=
-\,\frac{\eta}{L}\;
\mathbb{E}_{a \sim \pi_\theta^k(\cdot \mid s)}
\Big[
\frac{\mathbb{I}_{\text{clip}}(s,a)\, A(s,a)}{\pi_{\text{old}}(a\mid s)}
\,
\pi_\theta^k(a\mid s)\,
\bigl(1-\pi_\theta^k(a\mid s)\bigr)\,
\bigl(
\log \pi_\theta^k(a\mid s)
+
\mathcal{H}(\pi_\theta^k\mid s)
\bigr)
\Big],
\end{equation}
and the higher–order remainder term $\Phi(s)$ satisfies
\begin{equation}
\label{eq:remainder_bound}
|\Phi(s)|
\;\le\;
C \,\eta^{2}
\left[
   \frac{A_{\max}\,r_{\max}}{L}
\right]^{\!2},
\end{equation}
where $L$ is the total decoded length in the GRPO update and
$A_{\max}, r_{\max}$ bound the token–level advantage and
importance ratio and $C>0$ is a constant depending on the policy parameterization.
\end{theorem}

\begin{proof}

We prove the theorem in five steps as follows.

\paragraph{Step 1: First-order Taylor expansion.}
Taking the first-order Taylor expansion of $\Delta \mathcal{H}(s)$ around
$z^{k}(s)$, we have
\begin{align}
\Delta \mathcal{H}(s)
&= \mathcal{H}(\pi_{\theta}^{k+1} \mid s)
 - \mathcal{H}(\pi_{\theta}^{k} \mid s) \notag \\
&=
\underbrace{
  \left\langle
    \frac{\partial \mathcal{H}(\pi_{\theta}^{k} \mid s)}{\partial z},
    \, z^{k+1}(s) - z^{k}(s)
  \right\rangle
}_{\text{first-order term } \Omega(s)}
+
\underbrace{
  \mathcal{O}\bigl(\|z^{k+1}(s) - z^{k}(s)\|_{2}^{2}\bigr)
}_{\text{higher-order remainder } \Phi(s)} ,
\end{align}
where $z(s) = (z_{s,a})_{a\in \mathcal{V}}$ is the logit vector for all actions
under state $s$.

\paragraph{Step 2: Gradient of entropy w.r.t. logits.}
For a fixed state $s$, the conditional entropy of the policy
$\pi_\theta^k(\cdot\mid s)$ is
\[
\mathcal{H}(\pi_\theta^k\mid s)
=
- \sum_{a\in \mathcal{V}}
  \pi_\theta^k(a\mid s)\,
  \log \pi_\theta^k(a\mid s).
\]
Under the parameter–independent softmax parameterization,
each token $a\in\mathcal{V}$ has a logit $z_{s,a}^k$ and
\[
\pi_\theta^k(a\mid s)
=
\frac{\exp(z_{s,a}^k)}{\sum_{b \in\mathcal{V}} \exp(z_{s,b}^k)}.
\]
We differentiate $\mathcal{H}(\pi_\theta^k\mid s)$ with respect to a
single logit $z_{s,b}$. Using the softmax derivative, we have
\[
\frac{\partial \pi_\theta^k(a\mid s)}{\partial z_{s,b}}
=
\pi_\theta^k(a\mid s)\,
\bigl(\mathbf 1\{a=b\}-\pi_\theta^k(b\mid s)\bigr).
\]
Then, we obtain
\begin{align*}
\frac{\partial \mathcal{H}(\pi_\theta^k\mid s)}{\partial z_{s,b}}
&=
- \sum_{a\in \mathcal{V}}
   \Bigl(\log \pi_\theta^k(a\mid s)+1\Bigr)
   \frac{\partial \pi_\theta^k(a\mid s)}{\partial z_{s,b}} \\[4pt]
&=
- \sum_{a\in \mathcal{V}}
   \Bigl(\log \pi_\theta^k(a\mid s)+1\Bigr)
   \pi_\theta^k(a\mid s)
   \bigl(\mathbf 1\{a=b\}-\pi_\theta^k(b\mid s)\bigr) \\[4pt]
&=
- \Bigl(\log \pi_\theta^k(b\mid s)+1\Bigr)
   \pi_\theta^k(b\mid s)\bigl(1-\pi_\theta^k(b\mid s)\bigr)
+ \pi_\theta^k(b\mid s)
   \sum_{a\neq b}
   \Bigl(\log \pi_\theta^k(a\mid s)+1\Bigr)
   \pi_\theta^k(a\mid s) \\[4pt]
&=
- \Bigl(\log \pi_\theta^k(b\mid s)+1\Bigr)
   \pi_\theta^k(b\mid s)\bigl(1-\pi_\theta^k(b\mid s)\bigr)  \\[-2pt]
&\qquad
+ \pi_\theta^k(b\mid s)
   \left[
      \sum_{a\in \mathcal{V}}
      \Bigl(\log \pi_\theta^k(a\mid s)+1\Bigr)
      \pi_\theta^k(a\mid s)
      - \Bigl(\log \pi_\theta^k(b\mid s)+1\Bigr)\pi_\theta^k(b\mid s)
   \right] \\[4pt]
&=
- \Bigl(\log \pi_\theta^k(b\mid s)+1\Bigr)\pi_\theta^k(b\mid s)
+ \pi_\theta^k(b\mid s)
   \sum_{a\in \mathcal{V}}
   \Bigl(\log \pi_\theta^k(a\mid s)+1\Bigr)\pi_\theta^k(a\mid s),
\end{align*}
where $\mathcal{V}$ denotes the vocabulary.
Note that
\[
\sum_{a\in \mathcal{V}}\pi_\theta^k(a\mid s)=1,
\qquad
\sum_{a\in \mathcal{V}}
   \log \pi_\theta^k(a\mid s)\,\pi_\theta^k(a\mid s)
= -\mathcal{H}(\pi_\theta^k\mid s).
\]
Therefore,
\[
\sum_{a\in \mathcal{V}}
   \Bigl(\log \pi_\theta^k(a\mid s)+1\Bigr)\pi_\theta^k(a\mid s)
= -\mathcal{H}(\pi_\theta^k\mid s)+1.
\]
Substituting back, we get
\begin{align*}
\frac{\partial \mathcal{H}(\pi_\theta^k\mid s)}{\partial z_{s,b}}
&=
- \Bigl(\log \pi_\theta^k(b\mid s)+1\Bigr)\pi_\theta^k(b\mid s)
+ \pi_\theta^k(b\mid s)\bigl(-\mathcal{H}(\pi_\theta^k\mid s)+1\bigr) \\[4pt]
&=
-\,\pi_\theta^k(b\mid s)\,
   \Bigl(\log \pi_\theta^k(b\mid s)
         + \mathcal{H}(\pi_\theta^k\mid s)\Bigr).
\end{align*}
Thus, for any $a\in \mathcal{V}$,
\begin{equation}
\label{eq:entropy_grad_logit}
\frac{\partial \mathcal{H}(\pi_\theta^k\mid s)}{\partial z_{s,a}}
=
-\,\pi_\theta^k(a\mid s)\,
   \bigl(\log \pi_\theta^k(a\mid s)
         + \mathcal{H}(\pi_\theta^k\mid s)\bigr).
\end{equation}

\paragraph{Step 3: One-step GRPO update in logit space.}
The policy gradient of GRPO in Eq.~\eqref{grpo_Eq} can be written as
\begin{equation}
\label{eq:grpo_pg_detail}
    \nabla_\theta J(\theta)
    =
    \mathbb{E}_{\substack{
        q \sim \mathcal{D}, \\
        \{o_i\}_{i=1}^G \sim \pi_{\text{old}}(\cdot \mid q)
    }}
    \Bigg[
        \frac{1}{\sum_{i=1}^G |o_i|}
        \sum_{i=1}^G \sum_{t=1}^{|o_i|}
        \mathbb{I}_{\text{clip}}(i,t)\,
        \frac{\pi_\theta^k\!\big(o_{i,t}\mid q,o_{i,<t}\big)}
             {\pi_{\text{old}}\!\big(o_{i,t}\mid q,o_{i,<t}\big)}
        A_{i,t}\,
        \nabla_\theta \log \pi_\theta\big(o_{i,t}\mid q,o_{i,<t}\big)
    \Bigg],
\end{equation}
where $\mathbb{I}_{\text{clip}}(i,t)$ is the clipping indicator and
$A_{i,t}$ is the token-level advantage.

For notational convenience, fix a particular token position $(i,t)$ and
write
\[
s \triangleq (q, o_{i,<t}),
\qquad
a \triangleq o_{i,t}.
\]
We also write $\mathbb{I}_{\text{clip}}(s,a)$ and $A(s,a)$ to denote
the same quantities as functions of the state–action pair $(s,a)$.
Under Assumption~\ref{Parameter independence assumption},
each logit $z_{s,a}$ is treated as an independent parameter.
Under the softmax parameterization, the derivative of the log-probability
with respect to its own logit is
\[
\frac{\partial}{\partial z_{s,a}}
\log \pi_\theta^k(a\mid s)
=
1-\pi_\theta^k(a\mid s),
\]
while the derivatives with respect to logits of non-sampled actions
$a'\neq a$ at the same state $s$ are neglected.

An update of $z_{s,a}$ in the direction of the GRPO gradient with
learning rate $\eta$ therefore contributes
\begin{align}
z^{k+1}_{s,a} - z^{k}_{s,a}
&=
\eta\,
\frac{1}{\sum_{i'=1}^G |o_{i'}|}
\,
\mathbb{I}_{\text{clip}}(i,t)\,
\frac{\pi_\theta^k(a\mid s)}{\pi_{\text{old}}(a\mid s)}\,
A_{i,t}\,
\frac{\partial}{\partial z_{s,a}}
\log \pi_\theta^k(a\mid s) \notag\\[4pt]
&=
\eta\,
\frac{1}{\sum_{i'=1}^G |o_{i'}|}
\,
\mathbb{I}_{\text{clip}}(i,t)\,
\frac{\pi_\theta^k(a\mid s)}{\pi_{\text{old}}(a\mid s)}\,
A_{i,t}\,
\bigl(1-\pi_\theta^k(a\mid s)\bigr).
\label{eq:delta_z_sa_raw}
\end{align}
By switching to the $(s,a)$
notation explicitly, we obtain the simplified update
\begin{equation}
\label{eq:delta_z_sa_ratio}
z^{k+1}_{s,a} - z^{k}_{s,a}
=
\eta\, \frac{1}{\sum_{i'=1}^G |o_{i'}|} \,
\mathbb{I}_{\text{clip}}(s,a)\,
\frac{\pi_\theta^k(a\mid s)}{\pi_{\text{old}}(a\mid s)}\,
A(s,a)\,
\bigl(1-\pi_\theta^k(a\mid s)\bigr).
\end{equation}
For non-sampled actions $a'$ at the same state $s$, the corresponding
logits $z_{s,a'}$ remain unchanged in this update.

\paragraph{Step 4: First–order estimation $\Omega(s)$.}
Plugging \eqref{eq:entropy_grad_logit} and
\eqref{eq:delta_z_sa_ratio} into the inner product, we have
\begin{align*}
\Omega(s)
&=
\left\langle
  \frac{\partial \mathcal{H}(\pi_{\theta}^{k} \mid s)}{\partial z},
  \, z^{k+1}(s)-z^{k}(s)
\right\rangle                                                  \\
&=
\sum_{a\in \mathcal{V}}
\frac{\partial \mathcal{H}(\pi_{\theta}^{k} \mid s)}{\partial z_{s,a}}\,
\bigl(z^{k+1}_{s,a}-z^{k}_{s,a}\bigr)                          \\
&=
\sum_{a\in \mathcal{V}}
\Bigl[
 -\,\pi_\theta^k(a\mid s)\,
    \bigl(\log \pi_\theta^k(a\mid s)
          + \mathcal{H}(\pi_\theta^k\mid s)\bigr)
\Bigr]                                                         
\Bigl[
 \frac{\eta}{L}\,
 \frac{\mathbb{I}_{\text{clip}}(s,a)\,A(s,a)}{\pi_{\text{old}}(a\mid s)}
 \,
 \pi_\theta^k(a\mid s)\,
 \bigl(1-\pi_\theta^k(a\mid s)\bigr)
\Bigr]                                                         \\
&=
-\,\frac{\eta}{L}\;
\sum_{a\in \mathcal{V}}
\frac{\mathbb{I}_{\text{clip}}(s,a)\,A(s,a)}{\pi_{\text{old}}(a\mid s)}
\,
\bigl[\pi_\theta^k(a\mid s)\bigr]^{2}
\bigl(1-\pi_\theta^k(a\mid s)\bigr)
\bigl(
\log \pi_\theta^k(a\mid s)
+
\mathcal{H}(\pi_\theta^k\mid s)
\bigr)                                                        \\
&=
-\,\frac{\eta}{L}\;
\mathbb{E}_{a \sim \pi_\theta^k(\cdot \mid s)}
\Big[
\frac{\mathbb{I}_{\text{clip}}(s,a)\,A(s,a)}{\pi_{\text{old}}(a\mid s)}
\,
\pi_\theta^k(a\mid s)\,
\bigl(1-\pi_\theta^k(a\mid s)\bigr)\,
\bigl(
\log \pi_\theta^k(a\mid s)
+
\mathcal{H}(\pi_\theta^k\mid s)
\bigr)
\Big],
\end{align*}
where $L$ denotes $\sum_{i'=1}^G  |o_{i'}|$ for short.
In the last line, we rewrote the sum as an expectation under
$a\sim\pi_\theta^k(\cdot\mid s)$ by absorbing one factor
$\pi_\theta^k(a\mid s)$ into the measure.
This is exactly Eq.~\eqref{eq:Omega_expanded}.

\paragraph{Step 5: Remainder term $\Phi(s)$.}
By multivariate Taylor's theorem, the higher–order remainder can be
written as a quadratic form in the logit increment, so there exists a
constant $C>0$ such that for any state $s$,
\begin{equation}
\label{eq:phi_bound_pre}
|\Phi(s)|
\;\le\;
C \,\Bigl\|z^{k+1}(s)-z^{k}(s)\Bigr\|_2^{2}.
\end{equation}
For a given state $s$, only the sampled action $a$ has a nonzero logit
update, hence
\[
\Bigl\|z^{k+1}(s)-z^{k}(s)\Bigr\|_2^{2}
=
\bigl(z^{k+1}_{s,a}-z^{k}_{s,a}\bigr)^2.
\]
Denote $L\triangleq\sum_{i'=1}^G |o_{i'}|, r(s,a) \triangleq \frac{\pi_\theta^k(a\mid s)}{\pi_{\text{old}}(a\mid s)}$,
and assume the importance ratios and the token–level advantage are uniformly bounded:
\[
|r(s,a)| \le r_{\max},
\qquad
|A(s,a)| \le A_{\max}
\quad \text{for all } (s,a).
\]
Using the GRPO update in Eq.~\eqref{eq:delta_z_sa_ratio}, we obtain
\begin{align*}
\Bigl\|z^{k+1}(s)-z^{k}(s)\Bigr\|_2^{2}
&=
\bigl(z^{k+1}_{s,a}-z^{k}_{s,a}\bigr)^2 \\[2pt]
&=
\eta^2
\left[
   \frac{\mathbb{I}_{\text{clip}}(s,a)\,A(s,a)}{L}\,
   r(s,a)\,\bigl(1-\pi_\theta^k(a\mid s)\bigr)
\right]^{\!2} \\[2pt]
&\le
\eta^2
\left[
   \frac{\mathbb{I}_{\text{clip}}(s,a)\,A(s,a)}{L}\,
   r(s,a)
\right]^{\!2} \\[2pt]
&\le
\eta^2
\left[
   \frac{A(s,a) \, r(s,a)}{L}\,
\right]^{\!2} \\[2pt]
&\le
\eta^2
\left[
   \frac{A_{\max}\,r_{\max}}{L}
\right]^{\!2},
\end{align*}
Combining this bound with \eqref{eq:phi_bound_pre} and absorbing all
fixed constants into $C$ yields
\begin{equation}
\label{eq:remainder_bound_final}
|\Phi(s)|
\;\le\;
C \,\eta^{2}
\left[
   \frac{A_{\max}\,r_{\max}}{L}
\right]^{\!2}.
\end{equation}
Thus the remainder term is of order $\mathcal{O}(\eta^2)$ and is further
suppressed by the per-update normalization factor $L^{-2}$ in GRPO.

Combining the above steps completes the proof.
\end{proof}

\clearpage

\end{document}